\documentclass[lettersize,journal]{IEEEtran}
\usepackage{amsmath,amsfonts}
\usepackage{algorithmic}
\usepackage{algorithm}
\usepackage{setspace}

\usepackage{array}
\usepackage{textcomp}
\usepackage{stfloats}
\usepackage{url}
\usepackage{verbatim}
\usepackage{graphicx}
\usepackage{cite}
\usepackage{color}
\usepackage{bm}
\usepackage{subfloat}
\usepackage{bbding}
\usepackage{makecell}                 
\usepackage{booktabs}                 
\usepackage{tabu}                     
\usepackage{multirow}                 
\usepackage{multicol}                 
\usepackage{multirow}                 
\usepackage{xcolor}
\usepackage[export]{adjustbox}
\usepackage{arydshln}
\usepackage[inline]{enumitem}
\usepackage{overpic}
\usepackage{pifont}
\usepackage{hyperref}
\usepackage[caption=false,font=small]{subfig}

\begin{document}
\title{Active Adversarial Noise Suppression for Image Forgery Localization}

\author{
	Rongxuan~Peng,~\IEEEmembership{Student Member,~IEEE,}
	Shunquan~Tan,~\IEEEmembership{Senior Member,~IEEE,}
	Xianbo~Mo,
	Alex C. Kot,~\IEEEmembership{Life Fellow,~IEEE, }%
	and~Jiwu~Huang,~\IEEEmembership{Fellow,~IEEE}%
    \thanks{Rongxuan Peng is with the College of Electronic and Information Engineering, Shenzhen University. Rongxuan Peng and Shunquan Tan are with Shenzhen Key Laboratory of Media Security.}%
    
    \thanks{Shunquan Tan, Xianbo Mo, Alex C. Kot, and Jiwu Huang are with the Guangdong Laboratory of Machine Perception and Intelligent Computing, Faculty of Engineering, Shenzhen MSU-BIT University.}%
    
    \thanks{Corresponding author: Shunquan Tan. (e-mail: tansq@smbu.edu.cn)}%
}

\maketitle
\begin{abstract}
Recent advances in deep learning have significantly propelled the development of image forgery localization. However, existing models remain highly vulnerable to adversarial attacks: imperceptible noise added to forged images can severely mislead these models. In this paper, we address this challenge with an Adversarial Noise Suppression Module (ANSM) that generate a defensive perturbation to suppress the attack effect of adversarial noise. We observe that forgery-relevant features extracted from adversarial and original forged images exhibit distinct distributions. To bridge this gap, we introduce Forgery-relevant Features Alignment (FFA) as a first-stage training strategy, which reduces distributional discrepancies by minimizing the channel-wise Kullback–Leibler divergence between these features. To further refine the defensive perturbation, we design a second-stage training strategy, termed Mask-guided Refinement (MgR), which incorporates a dual-mask constraint. MgR ensures that the perturbation remains effective for both adversarial and original forged images, recovering forgery localization accuracy to their original level. Extensive experiments across various attack algorithms demonstrate that our method significantly restores the forgery localization model's performance on adversarial images. Notably, when ANSM is applied to original forged images, the performance remains nearly unaffected.
To our best knowledge, this is the first report of adversarial defense in image forgery localization tasks. We have released the source code and anti-forensics dataset.\footnote{\url{https://github.com/SZAISEC/ANSM}}
\end{abstract}
	
\begin{IEEEkeywords}
	Image forgery localization, adversarial noise, adversarial noise suppression module, forgery-relevant features alignment, mask-guided refinement.
\end{IEEEkeywords}

\section{Introduction}
\label{section:introduction}
\IEEEPARstart{W}{ith} the image editing technology and generative AI tools, the forged images without perceptual traces can be easily generated. The widespread distribution of such forged images lead to serious security risks in community.
\begin{figure}[t]
	\centering 
	\begin{overpic}[width=0.46\textwidth]{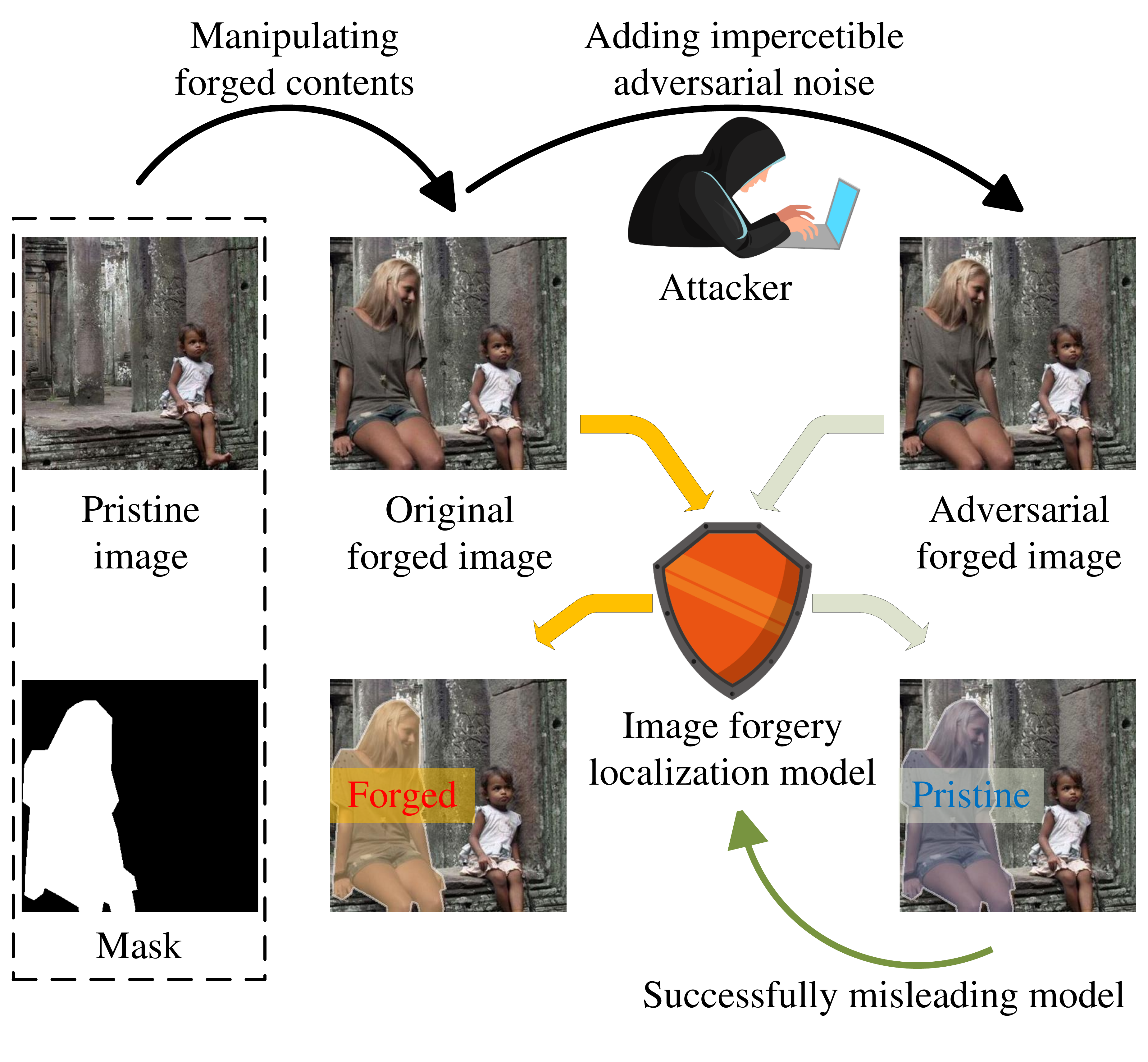} 
		
	\end{overpic}
	\caption{The diagram of forensics and anti-forensics. Once an attacker uses image editing tools to manipulate the forged contents, the forgery localization model can identify and localize the forgery region. However, by adding imperceptible adversarial noise to the forged image, the attacker can successfully mislead the model to make a wrong prediction.}
	\label{fig:intro}
\end{figure}

Image forgery localization, which aims to identify and localize tampered regions within an image, is a fundamental task in image forensics. In recent years, deep learning has become the dominant approach for this task, showing remarkable progress in addressing specific types of forgeries such as splicing~\cite{wu2017deep, destruel2018color, zhang2021multi}, copy-move~\cite{wu2018busternet, islam2020doa}, and inpainting~\cite{zhu2018deep, yang2020spatiotemporal}. More advanced models have further demonstrated the ability to localize multiple types of forgeries simultaneously~\cite{liu2022pscc, kwon2022learning, dong2022mvss, wu2022robust, peng2024employing, han2024hdf}.	

As forgery localization models continue to advance, attackers have likewise developed anti-forensics attack to mislead these models. Conventional anti-forensics attacks—such as JPEG compression, Gaussian noise addition, and blurring—are used to obscure forgery traces and reduce the effectiveness of forgery localization models. Although state-of-the-art methods~\cite{kwon2022learning, dong2022mvss, wu2022robust, peng2024employing} have shown improved robustness against such conventional attacks, a more subtle and powerful threat—adversarial attack—has received limited attention in this task. Recent studies have revealed that adversarial noise~\cite{mo2023poster, mo2025query} can effectively mislead forgery localization models, yet none of the existing models have been explicitly designed to resist such attacks.

It is well known that adversarial attacks pose a critical vulnerability for deep learning-based  models. As illustrated in Fig.~\ref{fig:intro}, the attacker can add imperceptible adversarial noise to a forged image, producing an adversarial forged image that appears visually identical to the original but causes the model to fail in correctly identifying the forged region.
Adversarial attacks~\cite{costa2024deep} commonly have white-box and black-box settings. In the white-box scenario, the attacker has complete knowledge of the victim model, including its architecture, parameters, and gradients. With this detailed information, the attacker can meticulously tailor highly precise noise to sharply degrade the model's performance. Consequently, such attacks are particularly potent and pose a formidable challenge to defense mechanisms.
Conversely, black-box setting assumes limited knowledge of the victim model. The attacker must either issue numerous queries to probe the victim model’s responses or rely on the transferability of adversarial images crafted on a similar surrogate model. As a result, black-box attack is generally less effective than white-box attack.
Our research focuses on a highly challenging scenario where the attacker operates under a white-box setting, with  free to employ any attack algorithm, while, the defender lacks prior knowledge of the specific attack algorithm being used. Our objective is to develop a defense strategy achieving generalization across multiple attack algorithms, thereby enhancing the resilience of the forgery localization model against such potential threats.

In this paper, we propose an Adversarial Noise Suppression Module (ANSM) to effectively suppress the attack effect of adversarial noise. In specific, ANSM generates a defensive perturbation, which is added to the adversarial forged image, obtaining defensive perturbed image before conducting forgery localization. Through empirical analysis, we observe that adversarial noise causes significantly larger distribution shift in the forgery-relevant feature space than in the raw pixel space, highlighting its disruptive effect on internal forgery representations. This insight motivates us to shift the defense paradigm from aimless pixel-wise reconstruction to the proposed Forgery-relevant Features Alignment (FFA). FFA aims to reducing the distribution discrepancy between forgery-relevant features extracted from adversarial and original forged image. While FFA ensures global consistency in the forgery-relevant features space, it does not always guarantee the expected pixel-level precise of the predicted forgery mask. To address this, we design the second-stage training strategy, called Mask-guided Refinement (MgR). MgR introduces a dual-mask constraint, encouraging the defensive perturbation to remain effective across both adversarial and original forged images. This constraint not only helps restore the forgery localization accuracy on adversarial forged image but also mitigates unintended performance degradation on original forged images. With the two-stage training strategy, ANSM effectively suppresses potential adversarial noise, thereby enhancing the robustness of forgery localization models against adversarial attacks. We constructed an anti-forensics dataset using six widely-used attack algorithms for comprehensive evaluation. And the experimental results validate the superior defense effect of our proposed method against multiple attack algorithms under white-box setting.

Our primary contributions can be summarized as follows:
\begin{itemize}
	\item Our experiments and analysis reveals that adversarial noise induces significantly distributional shifts in the forgery-relevant feature space compared to the pixel space. This finding underscores the necessity to focus on aligning forgery-relevant features.
	
	\item We design an adversarial noise suppression module that generates defensive perturbation to the input image, obtaining defensive perturbed image before performing forgery localization. This suppresses potential adversarial noise. ANSM acts as a pre-processing module during inference, suppressing adversarial noise without modifying the forgery localization model.
	
	\item Motivated by the finding, we propose two-stage training strategy: forgery-relevant features alignment and mask-guide refinement, to optimize the defensive perturbation. FFA is to reduce the distributional discrepancy between the forgery-relevant features extracted from adversarial and original forged images. MgR is to further enhance pixel-level precision of forgery localization results, which employs a dual-mask constraint that encourage the defensive perturbation remains effective on adversarial forged images while minimizing performance degradation on original forged images.
	
	\item Experimental results demonstrate that the proposed method can assist the forgery localization model to significantly recover its performance on adversarial forged images, while keeping its performance on original forged images largely unaffected. Our method also shows superior generalization ability across multiple attack algorithms and datasets under white-box setting.	
\end{itemize}

The remainder of this paper is structured as follows. Section~\ref{sect:Related Work} presents the related works about image forgery localization, adversarial attack and adversarial defense. Section~\ref{sect:Our Proposal} presents the problem statement and analysis of adversarial noise. Section~\ref{section:Defensive Framework} presents our proposed defensive framework. Section~\ref{sect:Experiment} presents the details of settings and comprehensive experimental results with various datasets and adversarial attack algorithms. Section~\ref{sect:Conclusion} concludes the paper and gives the prospect of our future work.

\section{Related Work}
\label{sect:Related Work}
\subsection{Image Forgery Localization}
Image forgery localization is a pixel-level binary classification task that distinguish the forgery region from the pristine region in an image. 
Compared to earlier works~\cite{wu2017deep, destruel2018color, zhang2021multi, wu2018busternet, islam2020doa, zhu2018deep, yang2020spatiotemporal} that focus on a single manipulation type, recent approaches are capable of handling multiple types of forgeries simultaneously, and attempts to capture more tampering traces from multiple perspectives. For instance, DFCN~\cite{zhuang2021image} used dense fully convolutional network to well capture tampering
traces and achieved good document forensics performance with the generated simulated data for training. MVSS-Net~\cite{dong2022mvss} employed ResNet-50~\cite{he2016deep} as the backbone to extract the boundary artifacts around the forgery region, and capture the noise inconsistency between the forged and pristine region. 
To capture the compression artifact from the frequency domain, CAT-Net v2~\cite{kwon2022learning} proposed a two-stage network to capture image acquisition artifacts in the RGB domain and learn the distribution of discrete cosine transform (DCT) coefficients. 
Later, since the transmission of online social networks will introduce unknown noise, IF-OSN~\cite{wu2022robust} directly adopted the SCSE U-Net~\cite{roy2018recalibrating} as the forgery localization model, paired with an U-Net~\cite{ronneberger2015u} based noise generator that mimics post-processing operations of online social networks to enhance the model's robustness against complicated noises. CoDE~\cite{peng2024employing} employed reinforcement learning to design an actor-critic model based on A3C (Asynchronous Advantage Actor-Critic) algorithm~\cite{mnih2016asynchronous}. In this model, each pixel is equipped with an agent that performs Gaussian distribution-based continuous action to iteratively update the respective forgery probability, enhancing the accuracy of forgery localization. HDF-Net~\cite{han2024hdf} transferred the prior knowledge of steganalysis rich model (SRM)~\cite{fridrich2012rich} and forms a dual-stream network with RGB.

Although the aforementioned state-of-the-art image forgery localization models have demonstrated strong robustness against conventional anti-forensics attacks—such as JPEG compression, resizing, cropping, blurring, and distortions introduced by online social network transmission—they still lack dedicated mechanisms to resist adversarial noise, which is specifically crafted to mislead the model in a targeted manner.
To address this limitation, we propose to suppress the potential adversarial noise in the input image actively before performing forgery localization. Notably, our approach does not require retraining the forgery localization models or access their whole training set, making it a flexible and low-cost solution.

\subsection{Adversarial Attack and Defense}
Adversarial attacks have been extensively investigated across various vision tasks, including image classification~\cite{goodfellow2014explaining, carlini2017towards, kurakin2018adversarial, madry2018towards}, instance segmentation~\cite{kang2020adversarial, zhang2022adversarial, rony2023proximal}, and object detection~\cite{li2021universal, yin2022adc, tang2023adversarial}, revealing the inherent vulnerabilities of deep neural networks. Though adversarial attacks on image forgery localization, a security-critical task, remain largely unexplored, our works demonstrates  that state-of-the-art forgery localization models are highly susceptible to the existing adversarial attacks. This underscores the pressing need for specialized adversarial defenses tailored to forgery localization. 

According to the access level to the victim model, adversarial attack has white-box and black-box settings. the attacker is assumed to have full knowledge of the victim model, including its architecture, parameters, and gradients. White-box attacks can be further divided into gradient-based methods (e.g., FGSM~\cite{goodfellow2014explaining}, BIM~\cite{kurakin2018adversarial}, PGD~\cite{madry2018towards}) and optimization-based methods (e.g., C\&W~\cite{carlini2017towards}). In contrast, the black-box setting assumes no access to the victim model’s internal details. Black-box attacks are commonly classified into two types: query-based attacks (e.g., ZOO~\cite{uesatos2018adversarial}, Square Attack~\cite{andriushchenko2020square}) that estimate gradients or search for adversarial noise via numerous model queries; and transfer-based attacks (e.g., MI-FGSM~\cite{dong2018boosting}, PGN~\cite{ge2023boosting}) that generate adversarial images on similar surrogate models.

To safeguard deployed models against adversarial attacks, one common strategy is adversarial training, which is to retrain models using adversarial images~\cite{goodfellow2014explaining, madry2018towards}. While this approach enhances robustness against specific attack algorithms, it often exhibits limited generalization capability and incurs high computational costs. Tsipras et al.~\cite{tsipras2019robustness} demonstrated that adversarial training can degrade performance on images without adversarial noise. Another defense strategy is to perform input transformation to suppress the attack effect of adversarial noise before inference. For instance, Dziugaite et al.~\cite{dziugaite2016study} and Das et al.~\cite{das2017keeping} employed JPEG compression as a defense mechanism, while Xie et al.~\cite{xie2017adversarial} proposed random resizing to mitigate the attack effect. Recently, generative models have gained traction for adversarial defense. For example, Samangouei et al.~\cite{samangouei2018defense} and Laykaviriyakul et al.~\cite{laykaviriyakul2023collaborative} trained a Generative Adversarial Model to reconstruct an approximate clean version from the adversarial image. GDMP~\cite{wang2022guided} utilized a Denoised Diffusion Probabilistic Model~\cite{ho2020denoising} to gradually submerge adversarial noise through the introduction of Gaussian noise during the diffusion process.

However, these defense strategies are primarily tailored for classification tasks and fail to adequately address the unique challenges posed by image forgery localization, such as preserving subtle tampering traces and ensuring pixel-level localization accuracy. Our proposed method effectively bridges this gap by explicitly addressing these limitations, offering enhanced robustness against adversarial attacks while maintaining precise localization performance.

\section{adversarial noise}
\label{sect:Our Proposal}
In this section, we first formulate the problem of adversarial attacks targeting image forgery localization. We then illustrate the generation process of adversarial noise with practical examples, analyze its impact on forgery localization performance, and discuss limitations of conventional defense strategies. This analysis reveals the necessity for improved defense strategy and provides essential support for the motivation and design of our proposed method.
\begin{table}[t]\small
	\centering
	\renewcommand{\arraystretch}{1.}
	\newcolumntype{P}[1]{>{\raggedright\arraybackslash}p{#1}}  
	\caption{high-frequency glossary.}
	\label{table:high-frequency glossary}
	\begin{tabular}{|l|P{3.3cm}|} 
		\hline
		\multicolumn{1}{|c|}{\textbf{Symbol}} & \multicolumn{1}{c|}{\textbf{Description}} \\
		\hline
		$\bm{X_{o}} \in \mathbb{R}^{3\times H\times W}$ & original forged image \\
		\hline
		$\bm{X_{a}} \in \mathbb{R}^{3\times H\times W}$ & Adversarial forged image \\
		\hline
		\multirow{2}*{$\bm{X_{po}},\bm{X_{pa}} \in \mathbb{R}^{3\times H\times W}$} & Defensive perturbed image of $\bm{X_{o}},\bm{X_{a}}$  \\
		\hline
		\multirow{2}*{$\bm{P_{o}},\bm{P_{a}},\bm{P_{po}},\bm{P_{pa}} \in {[0, 1]}^{H\times W}$} & Forgery probability map of $\bm{X_{o}},\bm{X_{a}},\bm{X_{po}},\bm{X_{pa}}$\\
		\hline
		\multirow{2}*{$\bm{M_{o}},\bm{M_{a}},\bm{M_{po}},\bm{M_{pa}} \in {\{0, 1\}}^{H\times W}$} & Binary forgery mask of $\bm{P_{o}},\bm{P_{a}},\bm{P_{po}},\bm{P_{pa}}$ \\
		\hline
		\multirow{2}*{$\bm{M_{gt}} \in {\{0, 1\}}^{H\times W}$} & Ground-truth forgery mask \\
		\hline
		$\delta \in {[0, 1]}$ & Forgery threshold\\
		\hline
		$\xi \in \mathbb{R}^{3\times H\times W}$ & adversarial noise\\
		\hline
	\end{tabular}
\end{table}

\subsection{Problem Statement}
For the sake of convenience, we list some used symbols
and corresponding description employed in this context in Table~\ref{table:high-frequency glossary}. The forgery localization model, denoted as $V_{\theta}$, provides forensic analysis by generating a forgery probability map $\bm{P_{o}}$ for the original forged image $\bm{X_{o}}$. Each value within $\bm{P_{o}}$ represents the probability that a corresponding pixel belongs to a forgery region. Subsequently, a predefined forgery threshold $\delta$ is applied to transform $\bm{P_{o}}$ into a binary mask $\bm{M_{o}}$, where 0 denotes the pristine region and 1 indicates the forgery region. The process is formalized in  Eq.~\eqref{eq:1} and Eq.~\eqref{eq:2}.
\begin{equation}
	\bm{P_{o}} = V_{\theta}(\bm{X_{o}})
	\label{eq:1}
\end{equation}
\begin{equation}
	\begin{gathered}
		\bm{M_{o}}^{(i, j)} = \begin{cases} 
			1 & \textbf{if } \bm{P_{o}}^{(i, j)} > \delta, \\
			0 & \textbf{otherwise.}
		\end{cases}\\
		(1\le i \le H, 1\le j \le W)
	\end{gathered}
	\label{eq:2}
\end{equation}
where $(i, j)$ is the position index, $H$ and $W$ denote the height and width. However, in adversarial attack scenarios, the attackers often design the adversarial noise $\xi$ to generate the adversarial forged image $\bm{X_{a}}=\bm{X_{o}}+\xi$ to mislead the forgery localization models. $\bm{X_{a}}$ is visually quite similar to $\bm{X_{o}}$, but such imperceptible adversarial noise can greatly decrease the accuracy of the predicted forgery mask $\bm{M_{a}}$ compared to the original mask $\bm{M_{o}}$, undermining the reliability of the forgery localization models. Moreover, in an ideal situation for an attacker, sophisticated adversarial noise could be added to the image to greatly increase the miss rate. This would cause the forgery localization  models to predict a mask $\bm{M_{a}} = \{0\}^{H\times W}$, meaning that the adversarial forged image is completely mistaken for the pristine image.
\begin{figure}[t]
	\centering
	\def\fig_scale{0.085}
	\setlength\tabcolsep{0.5mm}
	\def\model{CoDE}
	\begin{tabular}{c c c c c c c c c c c c}
		& & & & & & & & & & \\
		\begin{overpic}[scale=\fig_scale]{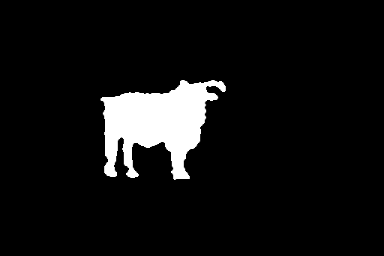}
			\put(22,80){\small $\bm{M_{gt}}$}
		\end{overpic}&
		\begin{overpic}[scale=\fig_scale]{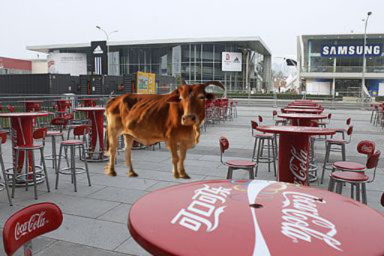}
			\put(25,80){\small $\bm{X_{o}}$}
		\end{overpic}&
		\begin{overpic}[scale=\fig_scale]{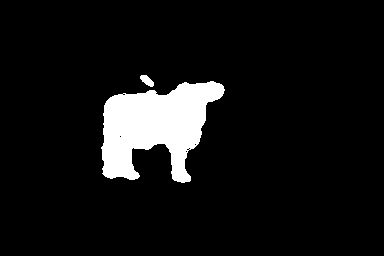}
			\put(25,80){\small $\bm{M_{o}}$}
		\end{overpic}&
		\begin{overpic}[scale=\fig_scale]{./pic/attack_effect/\model/casia1/image/Sp_D_NND_A_sec0067_ani0096_0620.png}
			\put(10,80){\small $\bm{X_{a}^{\text{FGSM}}}$}
		\end{overpic}&
		\begin{overpic}[scale=\fig_scale]{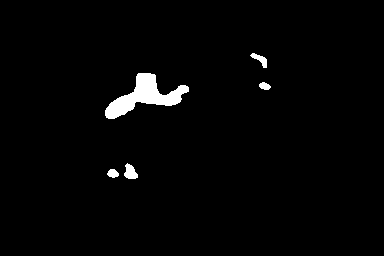}
			\put(3,80){\small $\bm{M_{a}^{\text{FGSM}}}$}
		\end{overpic}&
		\begin{overpic}[scale=\fig_scale]{./pic/attack_effect/\model/casia1/image/Sp_D_NND_A_sec0067_ani0096_0620.png}
			\put(10,80){\small $\bm{X_{a}^{\text{C\&W}}}$}
		\end{overpic}&
		\begin{overpic}[scale=\fig_scale]{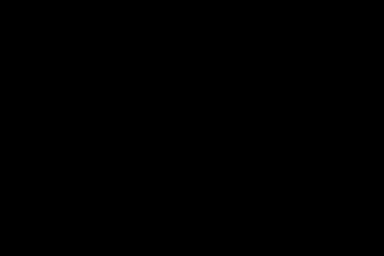}
			\put(6,80){\small $\bm{M_{a}^{\text{C\&W}}}$}
		\end{overpic}\\
		
		\begin{overpic}[scale=\fig_scale]{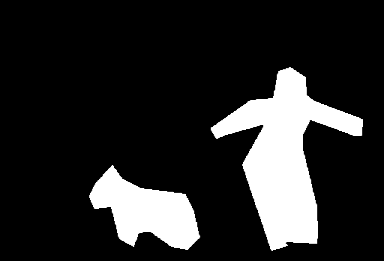}
			
		\end{overpic}&
		\begin{overpic}[scale=\fig_scale]{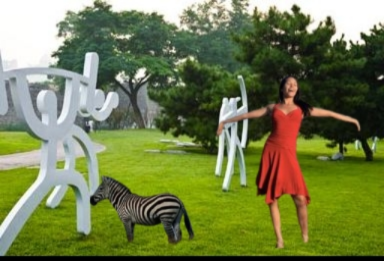}
			
		\end{overpic}&
		\begin{overpic}[scale=\fig_scale]{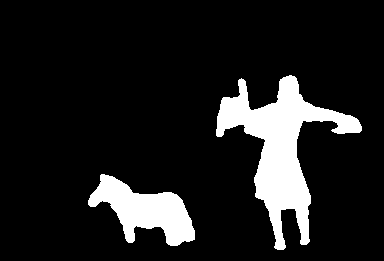}
			
		\end{overpic}&
		\begin{overpic}[scale=\fig_scale]{./pic/attack_effect/\model/misd/image/Sp_D_ani_0001_cha_00063_sec_00081_202.png}
			
		\end{overpic}&
		\begin{overpic}[scale=\fig_scale]{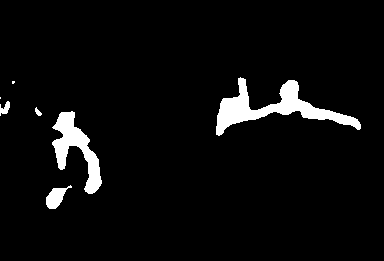}
			
		\end{overpic}&
		\begin{overpic}[scale=\fig_scale]{./pic/attack_effect/\model/misd/image/Sp_D_ani_0001_cha_00063_sec_00081_202.png}
			
		\end{overpic}&
		\begin{overpic}[scale=\fig_scale]{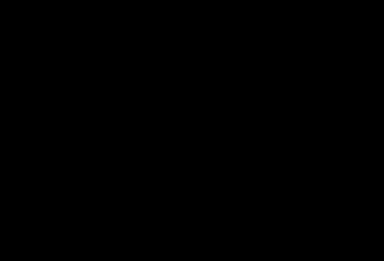}
			
		\end{overpic}\\
	\end{tabular}
	\caption{Visual comparisons before and after adversarial attack.}
	\label{fig:attack_results} 
\end{figure}

\subsection{Analysis of Adversarial Noise}
\label{section:Prior Analysis of adversarial noise}
We first outline the process of generating adversarial forged images using two typical approaches: gradient-based and optimization-based attack algorithms. Subsequently, we provide the quantitative analysis of such adversarial attack and common defensive strategies.

\subsubsection{Gradient-based Attack}
\begin{figure}[t]
	\centering
	\subfloat[FGSM Attack]{
		\includegraphics[scale=0.3]{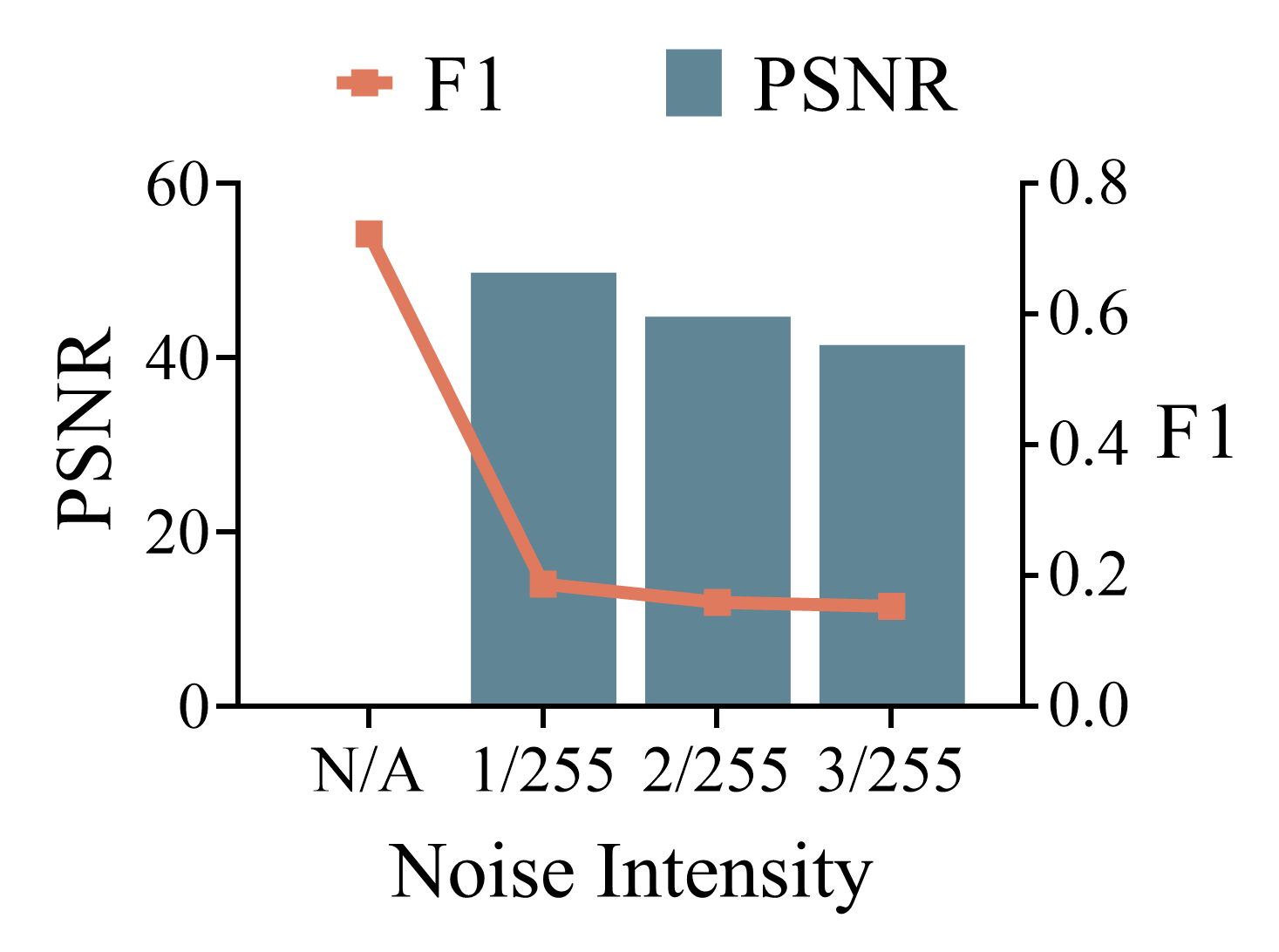} 
		\label{fig:fgsm_attack_effect}
	}
	\subfloat[C$\&$W Attack]{
		\includegraphics[scale=0.3]{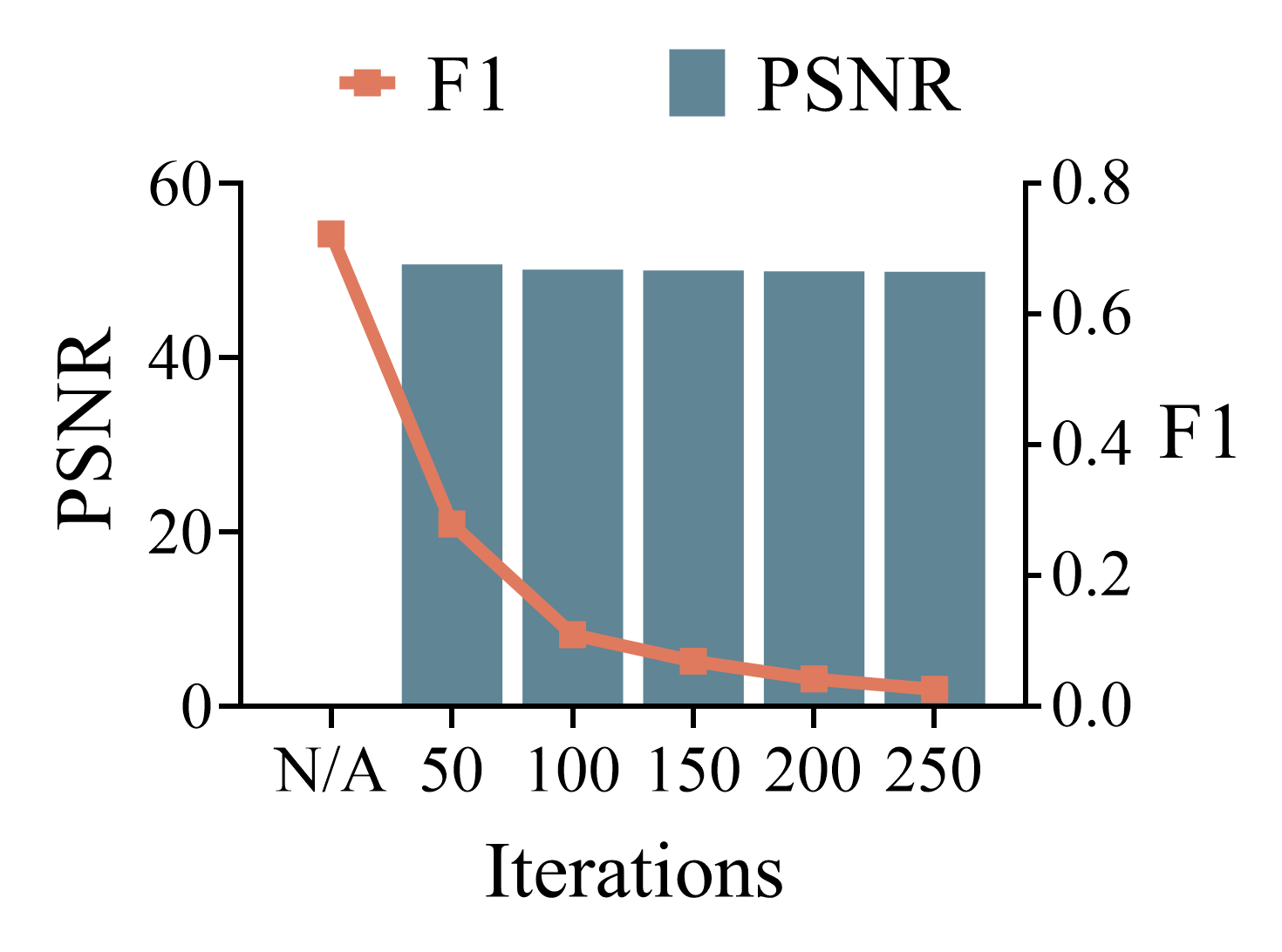}
		\label{fig:cw_attack_effect}
	}	
	\caption{Attack effect of FGSM and C$\&$W with different parameter settings.}
\end{figure}

Here we show example with the most classic gradient-based attack algorithm, FGSM (Fast Gradient Sign Method)~\cite{goodfellow2014explaining}, which performs a one-shot and non-targeted attack. The core idea behind FGSM is to apply a tiny adversarial noise to $\bm{X_{o}}$ in the direction of the gradient of the loss with respect to $\bm{X_{o}}$. The formula for generating $\bm{X_{a}}$ is as:
\begin{equation}
	\begin{gathered}
		\xi = \varphi \Delta_{\bm{X_{o}}} \text{sgn}(\nabla_{\bm{X_{o}}} \mathcal{J}(\theta, \bm{X_{o}}, \bm{M_{gt}}))\\
		\bm{X_{a}} = \text{TRUNC}\left(\bm{X_{o}} + \xi\right)
	\end{gathered}
	\label{eq:3}
\end{equation}
where $\varphi$ is the noise intensity. $\Delta_{\bm{X_{o}}}$ is the difference between the maximum and minimum values in the range (for example, when $\bm{X_{o}} \in \left[-1, 1\right]^{3\times H\times W}$, $\Delta_{\bm{X_{o}}} = 2$). $\mathcal{J}(\cdot)$ is the loss function. $\text{TRUNC}$ refers to truncating the values while keeping them within original range. As shown in Fig.~\ref{fig:attack_results}, FGSM attack cause the predicted forgery mask $\bm{M_{a}^{\text{FGSM}}}$ look more chaotic, which is caused by the simultaneous increase of false alarm rate and miss rate to some extent. And such attack can maintain relatively high image quality, which is shown in Fig.~\ref{fig:fgsm_attack_effect}.
\begin{figure}[t]
	\centering
	\subfloat[JPEG Compression]{
		\includegraphics[scale=0.3]{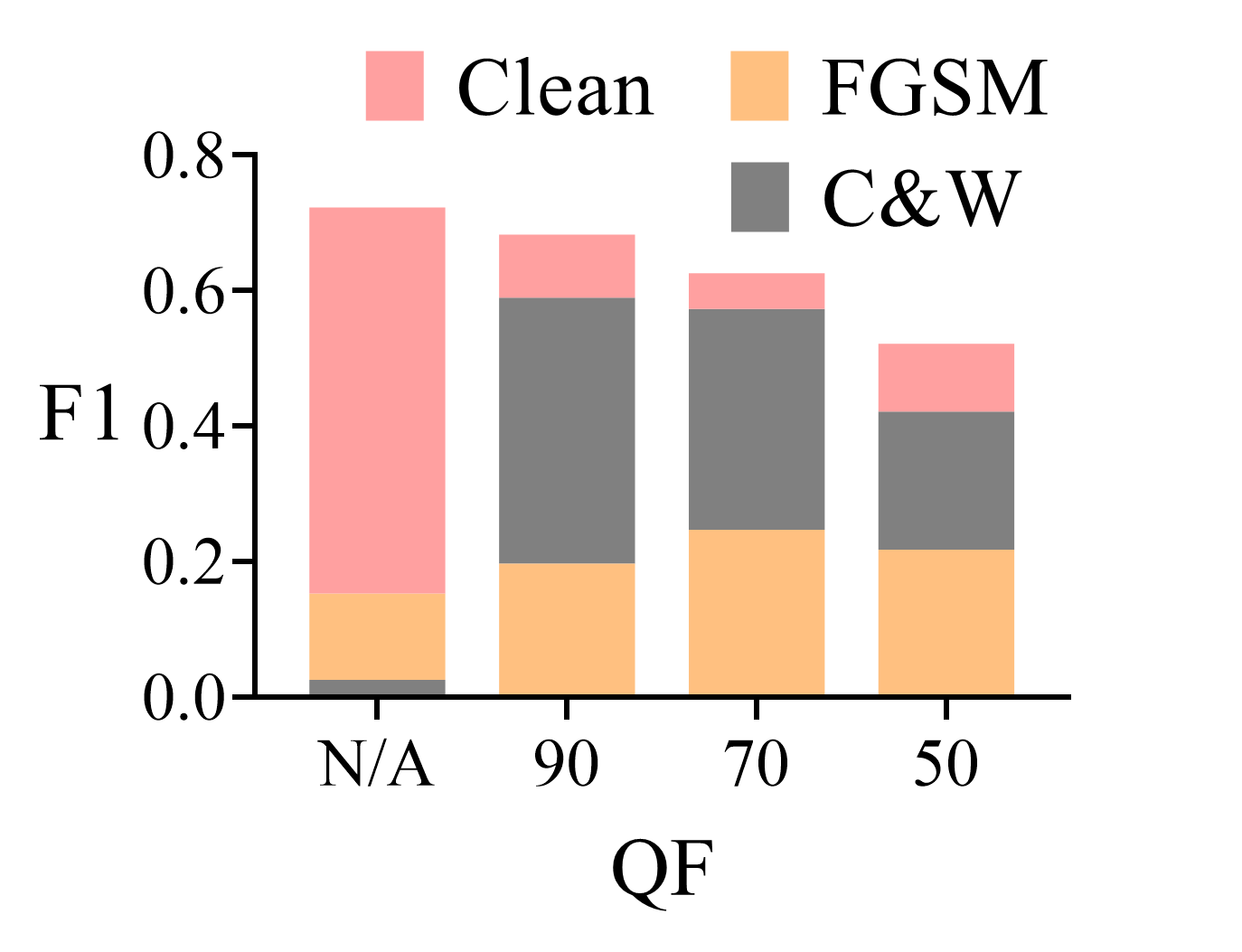} 
		\label{fig:osn_jpegcompress_vs_attack}
	}
	\subfloat[Resizing]{
		\includegraphics[scale=0.3]{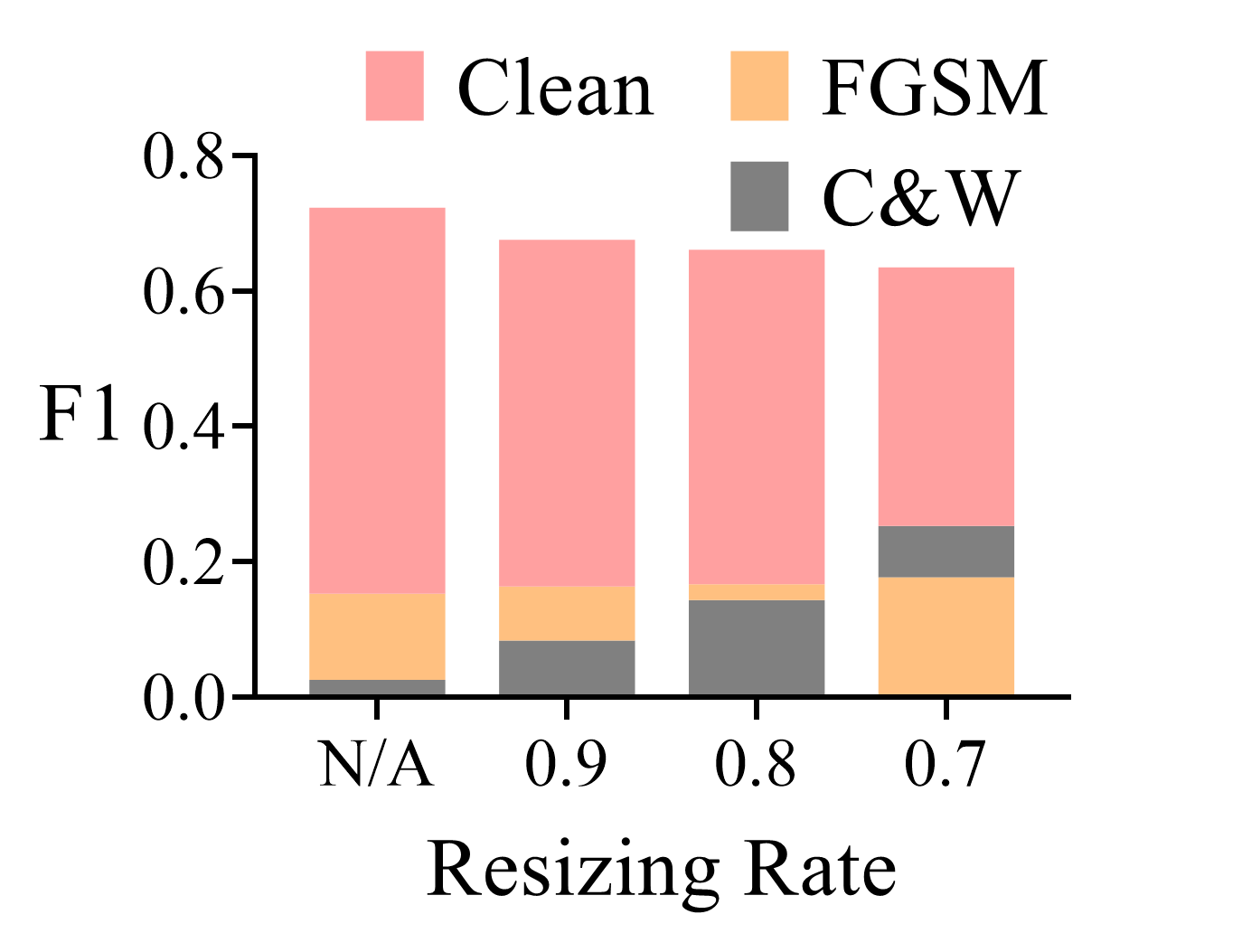}
		\label{fig:osn_resize_vs_attack}
	}
	
	\subfloat[Gaussian Noise]{
		\includegraphics[scale=0.3]{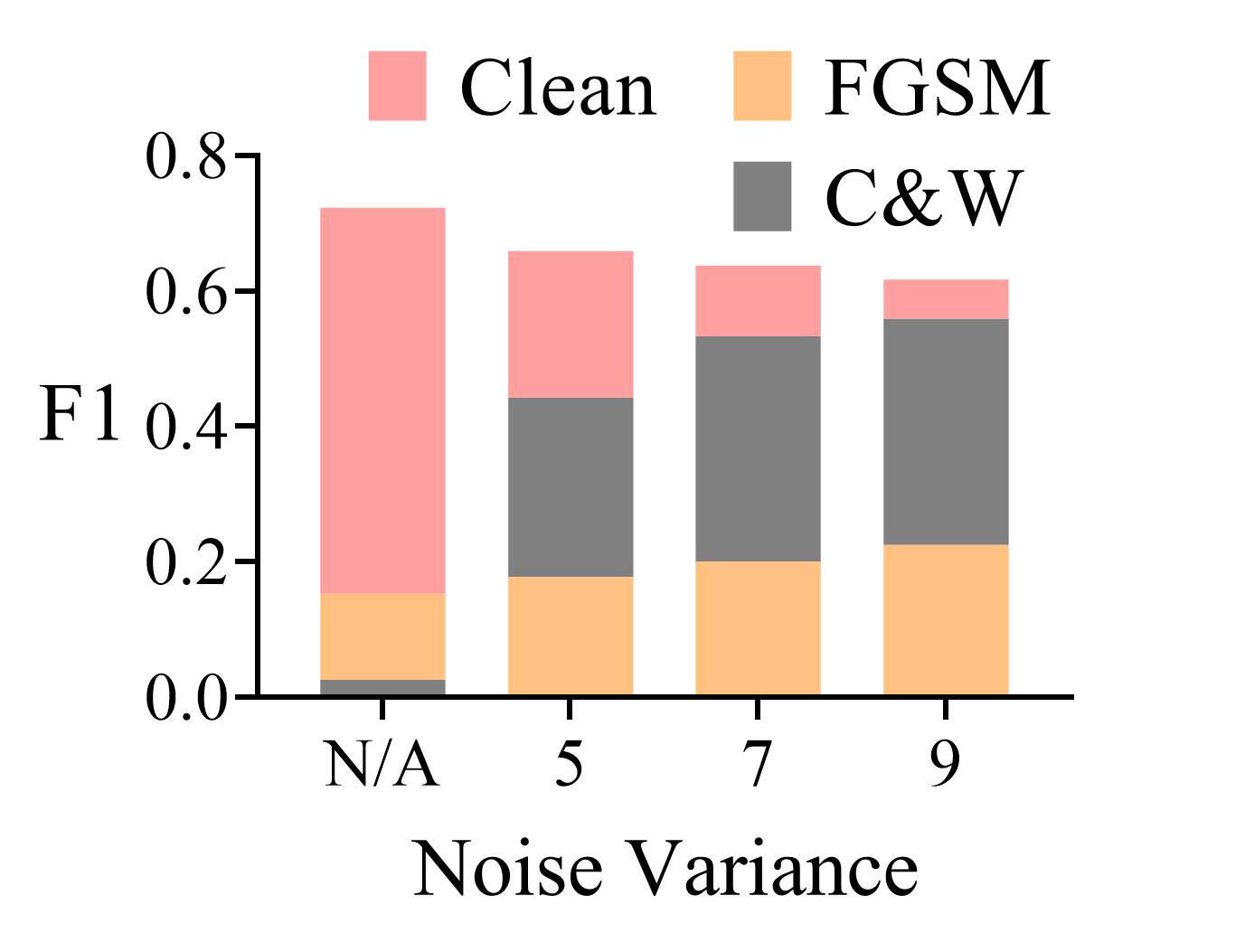} 
		\label{fig:osn_gaussiannoise_vs_attack}
	}
	\subfloat[Median Filter]{
		\includegraphics[scale=0.3]{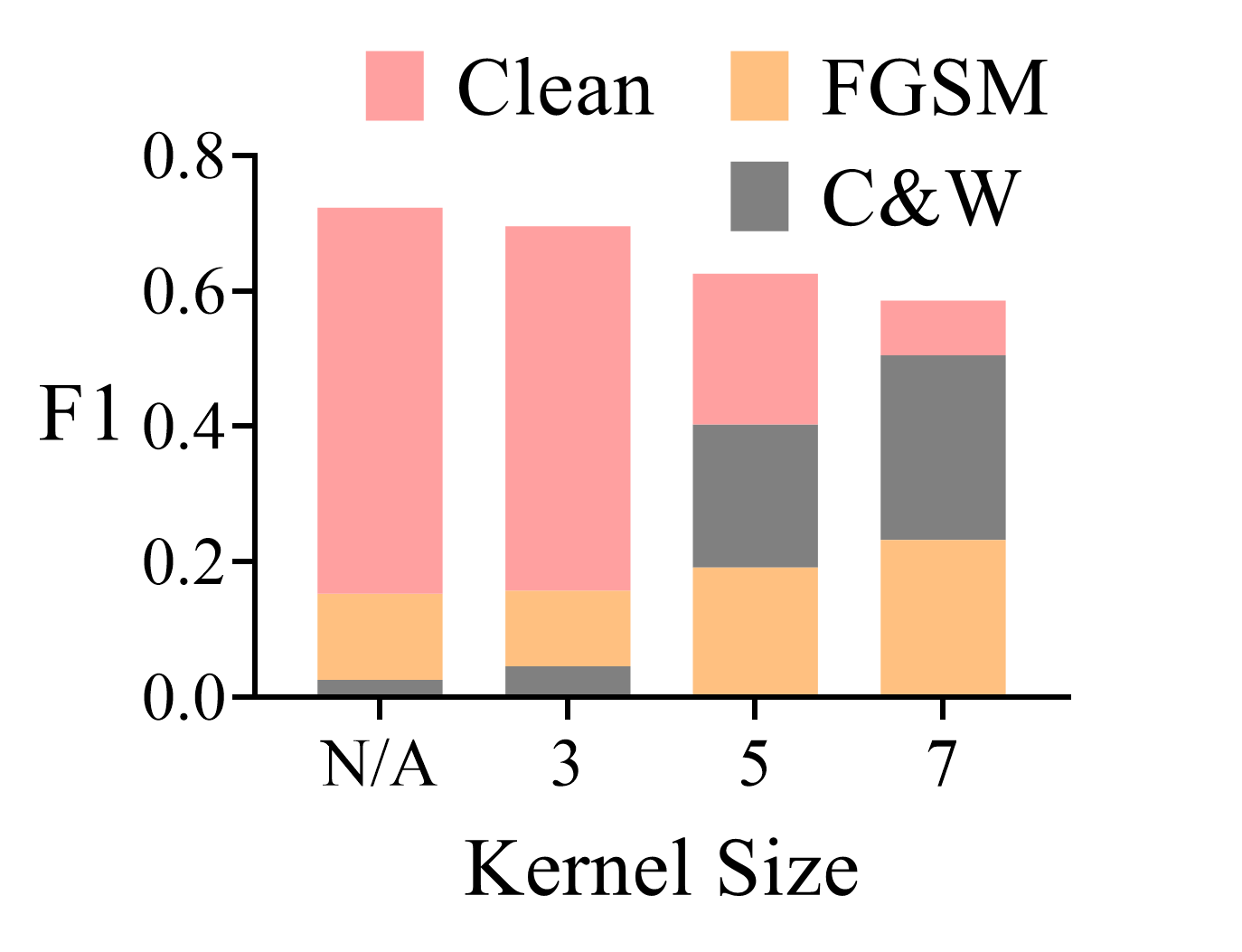}
		\label{fig:osn_medianblur_vs_attack}
	}	
	\caption{Defense effect of common defensive strategies.}
	\label{fig:postprocessing_vs_attack}
\end{figure}

\subsubsection{Optimization-based Attack} 
Unlike gradient-based attack, optimization-based attack first fixes a pre-defined target $\bm{M_{a}^{\text{C\&W}}}={\{0\}}^{H\times W}$ and then iteratively optimize the adversarial noise $\xi$ so that the forgery localization model’s predicted forgery mask converges to this target while simultaneously keeping the noise intensity as small as possible. Here we use the most classic optimization-based algorithm, C$\&$W (Carlini$\&$Wagner)~\cite{carlini2017towards}. At first, we present its loss function targeting the forgery localization task:
\begin{equation}
	\label{eq:4pre}
	\begin{gathered}
		\mathcal{L}(\xi) = ||\xi||^{2}_{2} + (\bm{P_{a}}-(1-\bm{P_{a}})-k) \\
		\bm{P_{a}} = V_{\theta}(\bm{X_{o}}+\xi) \\
	\end{gathered}
\end{equation}
where $k=\delta-(1-\delta)$, and $\delta$ is the forgery threshold. Next, the $\xi$ is optimized by $m$ times, and the last time step of $\xi$ is added to the $\bm{X_{o}}$ to generate $\bm{X_{a}}$. The optimization process is formulated as:
\begin{equation}
	\label{eq:4}
	\begin{gathered}
	\xi^{(t)} = \begin{cases} 
		0, & t=0, \\
		\xi^{(t-1)}-\nabla_{\xi}\eta\mathcal{L}(\xi^{(t-1)}), & t=1,...,m,\\
	\end{cases}\\
	\bm{X_{a}} = \text{TRUNC}\left(\bm{X_{o}} + \xi^{(m)}\right) \\
	\end{gathered}
\end{equation}
where $\eta$ is the learning rate. As shown in Fig.~\ref{fig:cw_attack_effect}, C$\&$W attack can maintain much higher image quality compared to FGSM, meaning that such adversarial noise are more imperceptible. The F1 gradually decreases to almost zero along the iterations.

To investigate the defense effect of conventional strategies, we introduce JPEG compression, resizing, Gaussian noise, and median filtering with varying intensities to pre-process image before performing forgery localization. As illustrated in Fig.~\ref{fig:postprocessing_vs_attack}, these strategies yield negligible improvements against FGSM, regardless of intensity. Although moderate gains are observed against C\&W, the performance remains markedly inferior to that on original forged images. These results underscore the limited efficacy of conventional lossy transformations or stochastic noise in mitigating adversarial noise.

\begin{figure}[t]
	\label{fig:distribution_image}
	\centering
	\def\fig_height{2.2cm}
	\setlength\tabcolsep{6mm}
	\subfloat[Pixel space - FGSM]{
		\label{fig:pixel_fgsm}
		\begin{tabular}{c}
			\begin{overpic}[height=\fig_height]{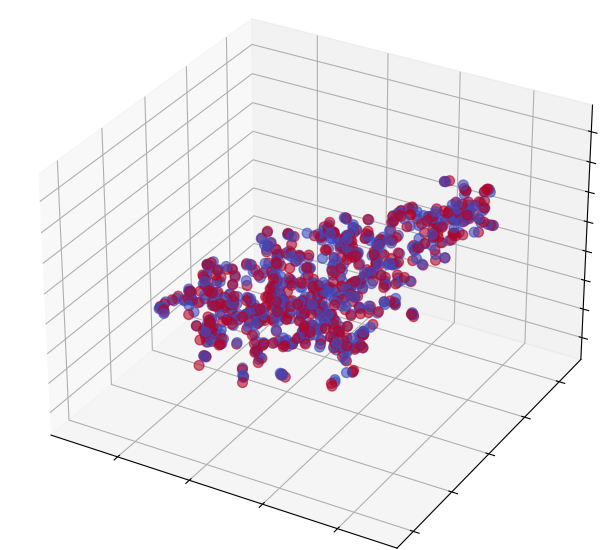}
			\end{overpic}
		\end{tabular}
	}	
	\subfloat[Pixel space - C$\&$W]{
		\label{fig:pixel_cw}
		\begin{tabular}{c}
			\begin{overpic}[height=\fig_height]{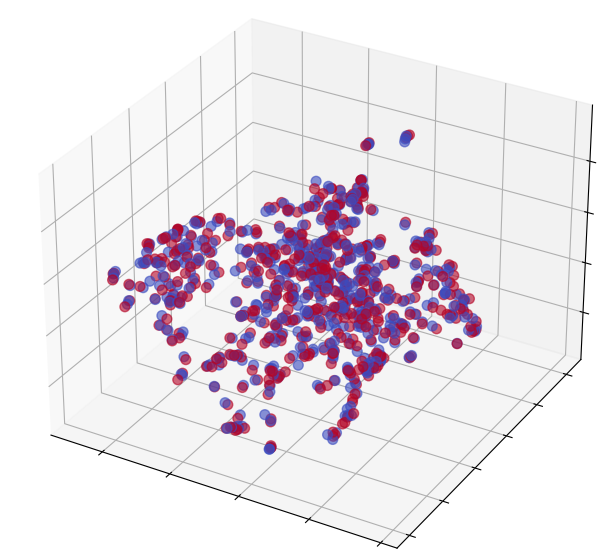}
			\end{overpic}
		\end{tabular}
	}	

	\subfloat[Feature space - FGSM]{
		\label{fig:feature_fgsm}
		\hspace{-0.1cm}
		\begin{tabular}{c}
			\begin{overpic}[height=\fig_height]{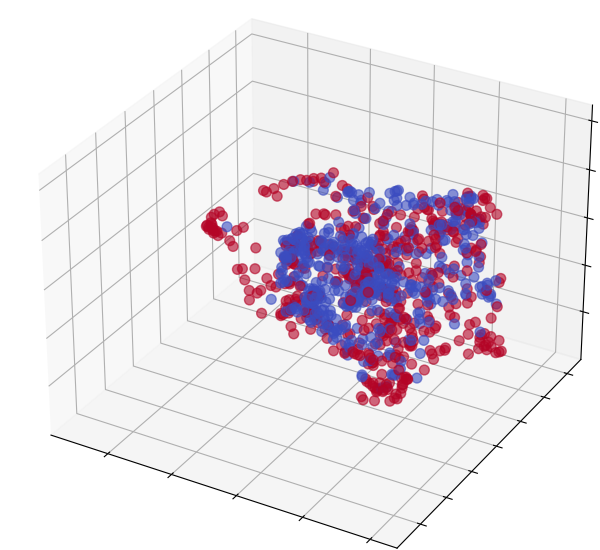}
			\end{overpic}
		\end{tabular}
	}	
	\subfloat[Feature space - C$\&$W]{
		\label{fig:feature_cw}
		\hspace{-0.3cm}
		\begin{tabular}{c}
			\begin{overpic}[height=\fig_height]{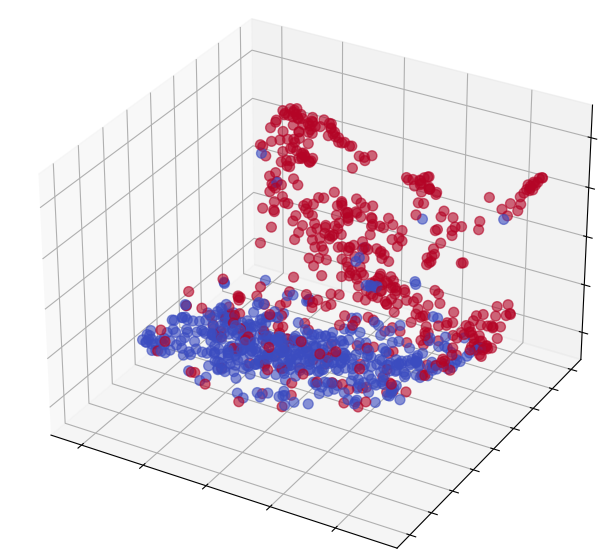}
			\end{overpic}
		\end{tabular}
	}
	\caption{Visualized distribution among 500 pairs of original forged images (red points) and adversarial forged images (blue points) using the dimensionality reduction algorithm UMAP. (a) and (b) are the projection of raw pixel values. (c) and (d) are the projection of forgery-relevant features.}
\end{figure}

This raises an important question: why do above diverse input transformations fail to suppress adversarial noise? To answer this, we examine whether the discrepancy between original and adversarial forged images can be observed in the pixel domain. Using UMAP ~\cite{mcinnes2018umap} for dimensionality reduction, we project 500 pairs of clean and adversarial forged images into a three-dimensional space based on their raw pixel values. As illustrated in Fig.~\ref{fig:pixel_fgsm} and Fig.~\ref{fig:pixel_cw}, the distributions of original and adversarial forged images appear nearly identical, suggesting that the adversarial noise is not well reflected in the pixel domain. In contrast, when we apply the same projection to their forgery-relevant features extracted by the forgery localization model, a clear distributional divergence emerges, as illustrated in Fig.~\ref{fig:feature_fgsm} and Fig.~\ref{fig:feature_cw}. This contrast result reveals that adversarial noise primarily disrupts the forgery-relevant features, so that mislead the prediction of forgery localization model. To effectively suppress adversarial noise, it is essential to design defense mechanisms that target the preservation and alignment of forgery-relevant features.

\section{Adversarial Defensive Framework}
\begin{figure*}[t]
	\centering 
	\subfloat[Training phase]{
	\label{fig:training_phase}
		\begin{overpic}[width=0.95\textwidth]{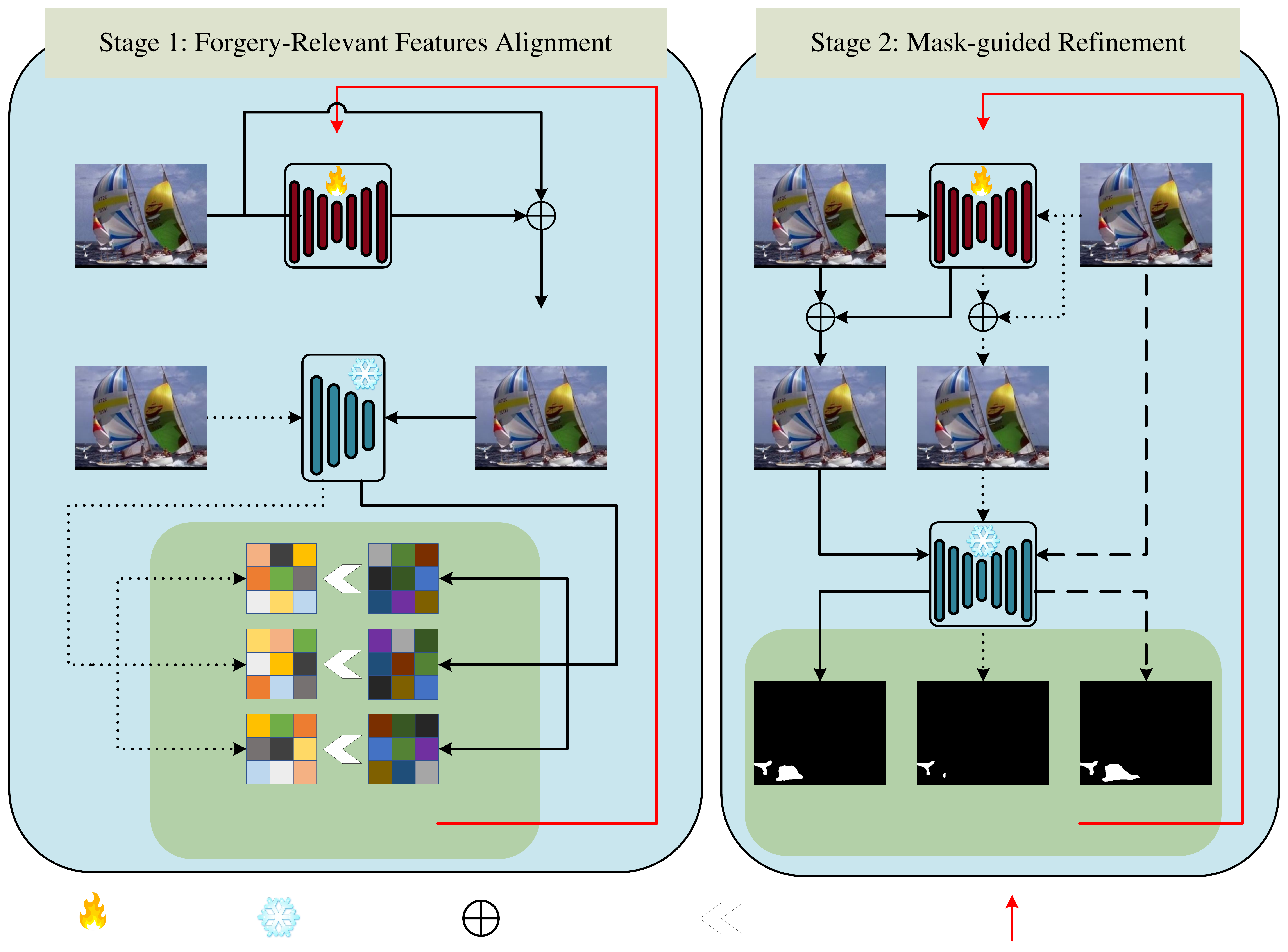} 
			\put(3, 64){Adversarial forged}
			\put(3, 62){image $\bm{X_{a}}$}
			
			\put(22., 62){ANSM $S_{\phi}$}
			
			\put(4.5, 48.5){Original forged}
			\put(4.5, 46.5){image $\bm{X_{o}}$}
			
			\put(23.5, 49){Feature}
			\put(21.5, 47){Extractor $E_{\theta}$}
			
			\put(34, 48.5){Defensive perturbed}
			\put(34, 46.5){image $\bm{X_{pa}}$}
			
			\put(14, 30){$\bm{D_{o}}^{(1)}$}
			\put(14, 23.5){$\bm{D_{o}}^{(2)}$}
			\put(14, 17){$\bm{D_{o}}^{(n)}$}
			
			\put(36, 30){$\bm{D_{pa}}^{(1)}$}
			\put(36, 23.5){$\bm{D_{pa}}^{(2)}$}
			\put(36, 17){$\bm{D_{pa}}^{(n)}$}
			
			\put(22.5, 9.5){Loss $\mathcal{L}_{FFA}$}
			
			\put(62.5, 62){$\bm{X_{a}}$}
			\put(72, 62){ANSM $S_{\phi^{'}}$}
			\put(88, 62){$\bm{X_{o}}$}
			
			\put(58.5, 46.5){$\bm{X_{pa}}$}
			\put(71.5, 46.5){$\bm{X_{po}}$}
			
			\put(66, 34.5){Loc. Model}
			\put(77.5, 34.5){$V_{\theta}$}
			
			\put(58.5, 22){$\bm{P_{pa}}$}
			\put(71.5, 22){$\bm{P_{po}}$}
			\put(84, 22){$\bm{M_{o}}$}
			
			\put(71.5, 9.5){Loss $\mathcal{L}_{MgR}$}
			
			\put(10, 2){Trainable}
			\put(25, 2){Frozen}
			\put(40, 3){Element-wise}
			\put(40, 1){addition}
			\put(60, 3){Channel-wise}
			\put(60, 1){KL divergence}
			\put(80, 2){Back propagation}
		\end{overpic}
		}	
	
	\subfloat[Test phase]{
	\label{fig:test_phase}
		\begin{overpic}[width=0.85\textwidth]{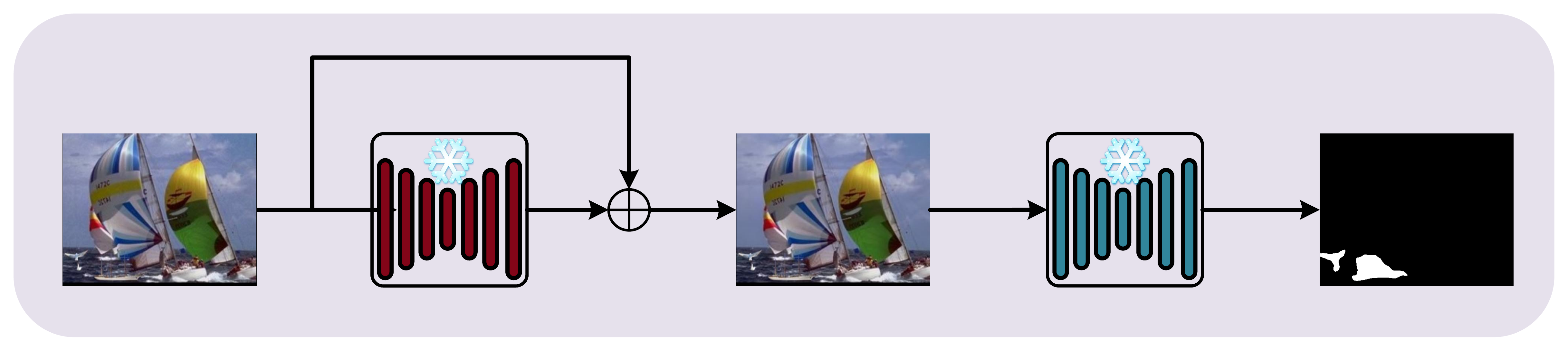}
			\put(5, 15){Input image}
			\put(24, 15){ANSM $S_{\phi^{\star}}$}
			
			\put(48.5, 17){Defensive}
			\put(46, 15){perturbed image}
			
			\put(66.5, 17){Loc. Model}
			\put(71, 15){$V_{\theta}$}
			
			\put(86, 17){Predicted}
			\put(84, 15){forgery mask}
		\end{overpic}
	}	
	\caption{The proposed defensive framework for image forgery localization task.}
	\label{fig:model}
\end{figure*}
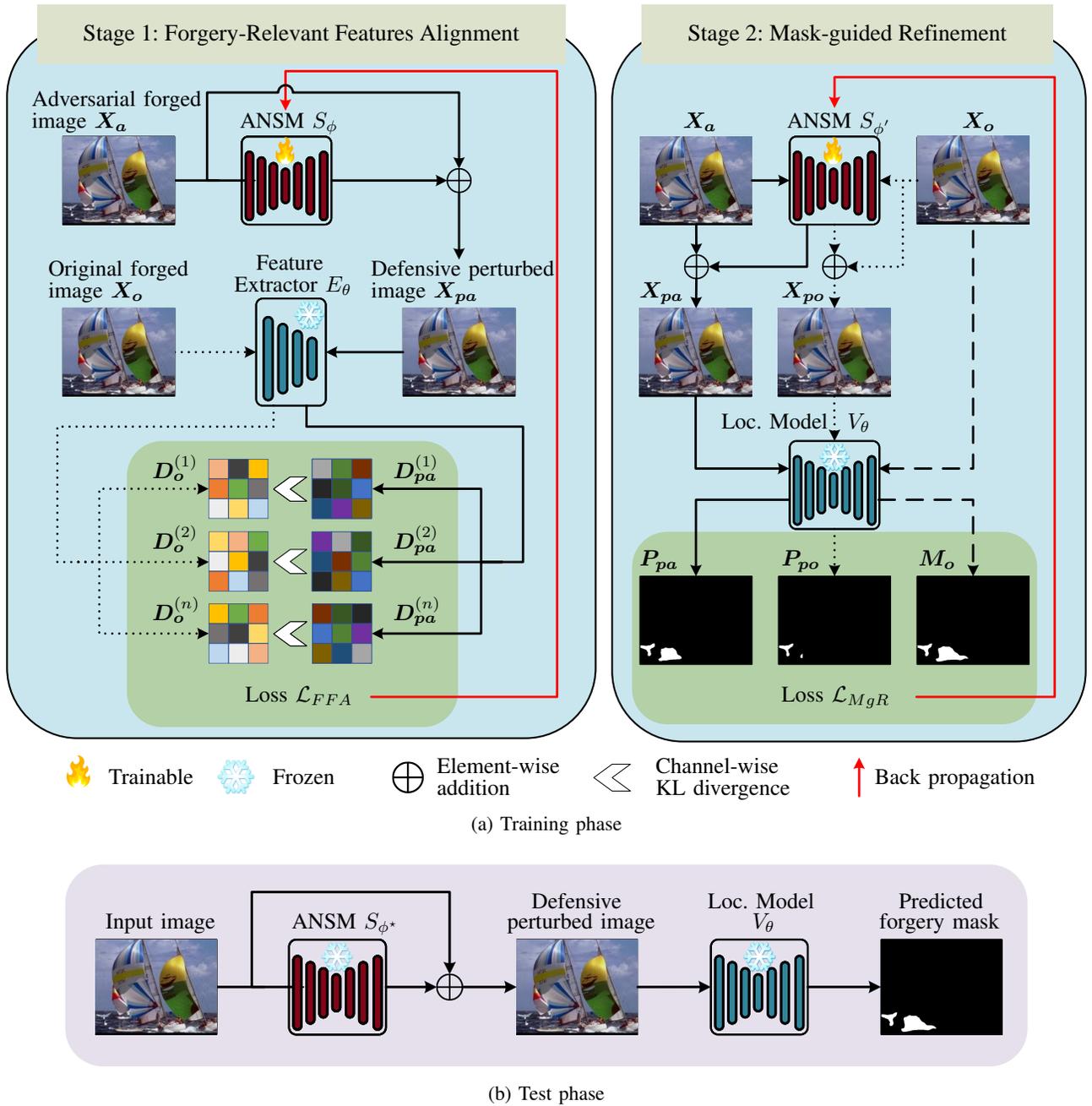
\label{section:Defensive Framework}
In this section, we present both the training and test phase of proposed defensive framework, as illustrated in Fig.~\ref{fig:model}. The framework comprises an ANSM, along with a two-stage training strategy: FFA and MgR.
\subsection{Adversarial noise Suppression Module}
\label{sec:Training and Testing Strategy}
Different white-box attack algorithms employ distinct strategies for generating adversarial noise, such as the gradient-based FGSM algorithm (Eq.~\eqref{eq:3}) and the optimization-based C\&W algorithm (Eq.~\eqref{eq:4}). Despite their methodological differences, these algorithms share a common underlying principle: they generate adversarial noise and directly add it to the original forged image, i.e., $\bm{X_{a}} = \text{TRUNC}(\bm{X_{o}} + \xi)$. Thus, adversarial noise can fundamentally be regarded as an additive noise.
Recognizing this, a natural defense strategy arises: generating a defensive perturbation that effectively cancels out the adversarial noise, thereby suppressing its attack effect. This motivates the design of the ANSM, which learns to estimate the defensive perturbation added to the adversarial forged image. Formally, the defensive perturbed image $\bm{X_{pa}}$ can be expressed as:
\begin{equation}
	\label{eq:5}
	\bm{X_{pa}} = \text{TRUNC}\left(\bm{X_{a}} + S_\phi(\bm{X_{a}})\right) \\
\end{equation}
where $S_\phi$ denotes the ANSM parameterized by $\phi$. 
The ANSM is designed as a light-weight encoder-decoder architecture.  
Specifically, the encoder employs EfficientNet-B4~\cite{tan2019efficientnet} backbone pretrained on the ImageNet~\cite{ILSVRC15}, leveraging rich feature representations to capture intricate adversarial noise patterns.
The decoder is inspired by U-Net~\cite{ronneberger2015u}, incorporating upsampling and skip connections between the encoder and decoder to integrate multi-scale features, thereby enhancing defense performance. The decoder comprises five upsampling blocks interleaved with four fusion blocks. Each upsampling block includes an upsampling operation with a scaling factor of 2, a $3\times3$ convolutional layer, a batch normalization layer, and a ReLU activation function. Each fusion block contains two sequential $3\times3$ convolutional layers, separated by a batch normalization layer and followed by a ReLU activation function.
Following the final upsampling block, the decoder branches into two parallel output heads. Each head consists of a $3\times3$ convolutional layer, a batch normalization layer, a ReLU activation function, and another $3\times3$ convolutional layer. One head applies the SoftSign function, scaling outputs to $(-1, 1)$ and denoted as $\gamma_{1} \in (-1,1)^{3\times H\times W}$. The other head employs the Sigmoid function, scaling outputs to $(0,1)$ and denoted as $\gamma_{2} \in (0,1)^{3\times H\times W}$. Finally, $S_\phi(\bm{X_{a}})$ is the element-wise manipulation of $\gamma_{1}$ and $\gamma_{2}$.

\subsection{Forgery-relevant Features Alignment}
\label{sect:Forgery-relevant Feature Alignment}
As analyzed in Section~\ref{section:Prior Analysis of adversarial noise}, the forgery-relevant features $\bm{F_{o}}$ and $\bm{F_{a}}$ extracted by the forgery localization model $V_{\theta}$ exhibit significant distributional discrepancies. After obtaining the defensive perturbed image $\bm{X_{pa}}$ with ANSM, we propose the FFA strategy to align the $\bm{F_{pa}}$ with $\bm{F_{o}}$. This alignment restores feature-level consistency and enhances the robustness of $V_{\theta}$ against adversarial noise.

In specific, we first froze the $V_{\theta}$ and only employ selected feature layers as the feature extractor $E_{\theta}$ to extract the multi-layer forgery-relevant features from $\bm{X_{o}}$ and $\bm{X_{pa}}$, which is formulated as:
\begin{equation}
	\left\{\bm{F_{o}}^{(1)},\bm{F_{o}}^{(2)},...,\bm{F_{o}}^{(n)}\right\} = E_{\theta}(\bm{X_{o}}) \\ 
	\label{eq:7}
\end{equation}
\begin{equation}
	\left\{\bm{F_{pa}}^{(1)},\bm{F_{pa}}^{(2)},...,\bm{F_{pa}}^{(n)}\right\} = E_{\theta}(\bm{X_{pa}}) \\ 
	\label{eq:8}
\end{equation}
Here $n$ denotes the number of feature layers. The shape of $i$-th forgery-relevant features $\bm{F_{o}}^{(i)}$ and $\bm{F_{pa}}^{(i)}$ is $(c_{i}, h_{i}, w_{i})$, where $c_{i}$ is the number of channels, $h_{i}$ and $w_{i}$ are the height and width, respectively.

Having extracted $\bm{F_{o}}^{(i)}$ and $\bm{F_{pa}}^{(i)}$, a critical question emerges: which feature layers of the forgery localization model are most effective for alignment? Although aligning all layers might initially appear to be an intuitive choice, this approach is computationally expensive and introduces considerable training overhead. To identify an optimal alignment strategy, we simply classify the layers into three categories based on their depth in the network: shallow layers (the first one-third of the layers of the neural network), middle layers (the intermediate one-third of the layers), and topmost layers (the last one-third of the layers). As detailed in Section~\ref{section:aligned_feature_layers}, experimental results demonstrate that aligning features from the middle layers achieves the best balance between preserving performance on original forged images and restoring performance on adversarial forged images.
This can be attributed to the characteristics of feature layers of different depths. It is well known that shallow layers primarily capture low-level features such as textures and edges~\cite{zeiler2014visualizing}. They may have limited sensitivity to forgery-relevant features and exhibit minimal differences between adversarial and original forged image. Conversely, the topmost layers extract highly forgery-relevant features that are strongly influenced by adversarial noise, but lack sufficient spatial detail to optimize the defensive perturbation more precisely. Middle layers balance forgery-relevant features and spatial context, providing more stable and informative representation for alignment.

Next, we aim to narrow the distribution discrepancy between $\bm{F_{o}}^{(i)}$ and $\bm{F_{pa}}^{(i)}$ using KL (Kullback–Leibler) divergence. The KL divergence is chosen as the optimization metric because it offers a principled measure of the discrepancy between two distributions, making it particularly suitable for FFA. Specifically, we first apply the channel-wise SoftMax transformation, termed $\text{{SoftMax}}_\text{C}$, on $\bm{F_{o}}^{(i)}$ and $\bm{F_{pa}}^{(i)}$ to normalize the values within [0, 1] along each channel, getting $\bm{D_{o}}^{(i)}$ and $\bm{D_{pa}}^{(i)}$, respectively:
\begin{equation}
	\bm{D_{o}}^{(i)} = \text{{SoftMax}}_\text{C}(\bm{F_{o}}^{(i)}), \ (1\leq i \leq n)
	\label{eq:9}
\end{equation}
\begin{equation}
	\bm{D_{pa}}^{(i)} = \text{{SoftMax}}_\text{C}(\bm{F_{pa}}^{(i)}), \ (1\leq i \leq n)
	\label{eq:10}
\end{equation}
Then we treat $\bm{D_{o}}^{(i)}$ as the target distribution and $\bm{D_{pa}}^{(i)}$ as the distribution to be optimized. The channel-wise KL divergence $\text{KL}_\text{C}$ is expressed as:
\begin{equation}
	\text{KL}(m \parallel q) = \frac{1}{N} \sum_{i=1}^{N} m_i \log \frac{m_i+\alpha}{q_i+\alpha}\\
	\label{eq:11}
\end{equation}
\begin{equation}
	\text{KL}_\text{C}(\bm{D_{pa}}^{(i)} \parallel \bm{D_{o}}^{(i)}) = \frac{1}{c_i} \sum_{j=1}^{c_i} \text{KL}((\bm{D_{pa}}^{(i)})_{j} \parallel (\bm{D_{o}}^{(i)})_{j})\\
	\label{eq:12}
\end{equation}
where $N$ denotes the size $h_i \times w_i$. The $\alpha = 1.0\times 10^{-8}$ is for preventing division by zero errors. $\text{KL}_\text{C}$ denotes the channel-wise KL, and $j$ denotes the $j$-th channel.
Then, the loss function $\mathcal{L}_{FFA}$ is designed to be the average of the multi-layer channel-wise KL divergence, which is expressed as:
\begin{equation}
	\mathcal{L}_{FFA} = \frac{1}{n} \sum_{i=1}^{n} \text{KL}_\text{C}\left(\bm{D_{pa}}^{(i)} \parallel \bm{D_{o}}^{(i)}\right)
	\label{eq:13}
\end{equation}
\subsection{Mask-guided Refinement}
\label{sect:Mask-guided Refinement}
After FFA optimization, the ANSM has a certain defensive effect against adversarial noise via achieving global consistency of forgery-relevant features. However, FFA does not fully guarantee spatial precision in mask prediction. To address this limitation, we design the second-stage training strategy, MgR. This stage explicitly guides the predicted forgery mask of $\bm{X_{pa}}$ is as precise as that of $\bm{X_{o}}$. Moreover, considering that the input image may be $\bm{X_{o}}$, we also conduct the same guidance for the predicted forgery mask of $\bm{X_{po}}$.

In this stage, we first obtain $\bm{X_{po}}$ and $\bm{X_{pa}}$ with the ANSM. Then $\bm{X_{po}}$ and $\bm{X_{pa}}$ are fed into the forgery localization model $V_{\theta}$ to generate two corresponding forgery probability maps: $\bm{P_{po}}$ and $\bm{P_{pa}}$, which serve as the optimization objectives. Next, we use a loss function composed of binary cross-entropy (bce) loss and dice loss~\cite{milletari2016v}:
\begin{equation}
	\mathcal{L}_{bce}(\bm{P}, \bm{M}) = -\frac{1}{N} \sum_{i=1}^{N} (\bm{M}_{i}\log(\bm{P}_{i})+(1-\bm{M}_{i})\log(1-\bm{P}_{i}))\
	\label{eq:bce_loss}
\end{equation}
\begin{equation}
	\mathcal{L}_{dice}(\bm{P}, \bm{M}) =  \frac{2\sum_{i=1}^{N}\bm{P}_{i}\bm{M}_{i}}{\sum_{i=1}^{N}\bm{P}^{2}_{i}+\sum_{i=1}^{N}\bm{M}^{2}_{i}}
	\label{eq:dice_loss}
\end{equation}
\begin{equation}
	\mathcal{L}_{m}(\bm{P}, \bm{M}) = \lambda_{bce}{\mathcal{L}_{bce}(\bm{P}, \bm{M})} + (1-\lambda_{bce}){\mathcal{L}_{dice}(\bm{P}, \bm{M})}
	\label{eq:combination_loss}
\end{equation}
where $\bm{P}_{i}$ and $\bm{M}_{i}$ denote the $i$-th value of predicted forgery probability map and the supervised forgery mask, respectively. $N$ is the number of values. The $\lambda_{bce}$ is set to 0.3 experimentally. Then we simultaneously optimize $\bm{P_{po}}$ and $\bm{P_{pa}}$, which is expressed as:
\begin{equation}
	\mathcal{L}_{MgR} = {\mathcal{L}_{m}(\bm{P_{po}}, \bm{M})} + {\mathcal{L}_{m}(\bm{P_{pa}}, \bm{M})}
	\label{eq:16}
\end{equation}

Notably, the supervised forgery mask $\bm{M}$ has two possible options: the ground-truth forgery mask $\bm{M_{gt}}$ and the predicted forgery mask of original forged image $\bm{M_{o}}$. In general, $\bm{M_{o}}$ is less accurate than $\bm{M_{gt}}$, and the degree of deviation depends on the generalization ability of forgery localization model. We use $\bm{M_{o}}$ as the supervised forgery mask to retain the forgery localization model’s original generalization ability. As comparison, once using $\bm{M_{gt}}$, ANSM not only learns to suppress adversarial noise, but also tends to learn redundant patterns irrelevant to adversarial noise to force $\bm{P_{pa}}$ and $\bm{P_{po}}$ nearly $\bm{M_{gt}}$. This may cause the overfitting problem that degrade the ANSM's performance on unseen datasets. In short, Eq.~\eqref{eq:16} is ultimately expressed as:
\begin{equation}
	\mathcal{L}_{MgR} = {\mathcal{L}_{m}(\bm{P_{po}}, \bm{M_{o}})} + {\mathcal{L}_{m}(\bm{P_{pa}}, \bm{M_{o}})}
	\label{eq:17}
\end{equation}

MgR complements FFA by explicitly guiding the predicted forgery masks of ANSM-processed images to align with those of the original forged images, thereby enhancing the effectiveness of the defensive perturbations in guiding more accurate forgery localization results, without compromising the original generalization ability of the localization model.
\begin{algorithm}[t]
	\small
	\setstretch{0.9}
	\caption{The Training Algorithm} 		
	\begin{algorithmic}[1]
		\label{al:training_algorithm}
		\REQUIRE Clean dataset $\mathcal{D}_{o}$ and FGSM dataset $\mathcal{D}_{a}$; frozen forgery-relevant features extractor $E_{\theta}$; initial ANSM $S_\phi$; training epoches $e_1,e_2$; learning rate $\eta_1,\eta_2$.
		\ENSURE	Well-trained ANSM $S_{\phi^{\star}}$ for $V_{\theta}$.\\
		$\ $\\
		\textit{\textbf{Stage 1: Forgery-relevant Features Alignment}} \\
		\FOR{epoch in $1$ to $e_1$}
		\FOR{($\bm{X_{o}}$, $\bm{X_a}$) $\subset$ ($\mathcal{D}_o$, $\mathcal{D}_a$)}
		\STATE Get $\bm{X_{pa}}$ with $S_\phi$; $\rhd$Eq.~\eqref{eq:5}\\
		\STATE Get $n$ layers of $\{\bm{F_{o}}^{(i)}\}^{n}_{i=1}$ and $\{\bm{F_{pa}}^{(i)}\}^{n}_{i=1}$ from $\bm{X_{o}}$ and $\bm{X_{pa}}$ using $E_{\theta}$; $\rhd$Eq.~\eqref{eq:7}~\eqref{eq:8}\\
		\STATE Get $\{\bm{D_{o}}^{(i)}\}^{n}_{i=1}$ and $\{\bm{D_{pa}}^{(i)}\}^{n}_{i=1}$ using channel-wise SoftMax transformation on $\{\bm{F_{o}}^{i}\}^{n}_{i=0}$ and $\{\bm{F_{pa}}^{i}\}^{n}_{i=0}$; $\rhd$Eq.~\eqref{eq:9}~\eqref{eq:10}\\
		\STATE Calculate $\mathcal{L}_{FFA}$; $\rhd$Eq.~\eqref{eq:13}\\
		\STATE Update the parameters $\phi \gets \phi - \eta_1 \nabla_{\phi}\mathcal{L}_{FFA}$;\\
		\ENDFOR
		\ENDFOR
		\STATE $S_{\phi^{\prime}} = S_\phi$;\\
		$\ $\\
		\textit{\textbf{Stage 2: Mask-guided Refinement}} \\
		\FOR{epoch in $1$ to $e_2$}
		\FOR{($\bm{X_{o}}$, $\bm{X_a}$) $\subset$ ($\mathcal{D}_o$, $\mathcal{D}_a$)}
		\STATE Get $\bm{X_{po}}$ and $\bm{X_{pa}}$ with $S_{\phi^{\prime}}$; $\rhd$Eq.~\eqref{eq:5}\\
		\STATE Get $\bm{P_{po}}$ and $\bm{P_{pa}}$; $\rhd$Eq.~\eqref{eq:1}\\
		\STATE Get $\bm{M_{o}}$; $\rhd$Eq.~\eqref{eq:1}~\eqref{eq:2}\\
		\STATE Calculate $\mathcal{L}_{MgR}$ using $\bm{P_{po}}$, $\bm{P_{pa}}$ and $\bm{M_{o}}$; $\rhd$Eq.~\eqref{eq:17}\\
		\STATE Update the parameters $\phi^{\prime} \gets \phi^{\prime} - \eta_2 \nabla_{\phi^{\prime}} \mathcal{L}_{MgR}$;\\
		\ENDFOR
		\ENDFOR\\
		\RETURN $S_{\phi^{\star}} = S_{\phi^{\prime}}$;
	\end{algorithmic} 
\end{algorithm}

\subsection{Training Phase}
In the preceding sections, we have provided a detailed exposition of the ANSM and the two-stage optimization strategy. The complete training process is presented in Algorithm~\ref{al:training_algorithm}.

Initially, we prepare a dataset of forged images without adversarial noise (clean dataset $\mathcal{D}_{o}$), along with a corresponding adversarial dataset $\mathcal{D}_{a}$ generated using FGSM (Eq.~\eqref{eq:3}) with fixed intensity. We then initialize the ANSM $S_\phi$ and perform the two-stage optimization strategy: FFA and MgR. 

During the FFA stage, we first obtain the defensive perturbed image $\bm{X_{pa}}$ with $S_\phi$. Subsequently, we extract $n$ layers of forgery-relevant features $\{\bm{F_{o}}^{i}\}^{n}_{i=1}$ and $\{\bm{F_{pa}}^{i}\}^{n}_{i=1}$ using the frozen feature extractor $E_{\theta}$. These features undergo a channel-wise SoftMax transformation, obtaining $\{\bm{D_{o}}^{(i)}\}^{n}_{i=1}$ and $\{\bm{D_{pa}}^{(i)}\}^{n}_{i=1}$. Finally, we compute the $n$ layers of channel-wise KL divergence between $\{\bm{D_{o}}^{(i)}\}^{n}_{i=1}$ and $\{\bm{D_{pa}}^{(i)}\}^{n}_{i=1}$ to derive the loss $\mathcal{L}_{FFA}$. After $e_1$ iterations, we obtain the updated ANSM $S_{\phi^{\prime}}$. 

In the MgR stage, we first obtain the defensive perturbed images $\bm{X_{po}}$ and $\bm{X_{pa}}$ with $S_{\phi^{\prime}}$. We then acquire the corresponding forgery probability maps $\bm{P_{o}}$, $\bm{P_{po}}$, $\bm{P_{pa}}$ via the frozen forgery localization model $V_{\theta}$, with $\bm{P_{o}}$ being converted into 
$\bm{M_{o}}$ to serve as the supervised forgery mask. In the end, we calculate the loss $\mathcal{L}_{MgR}$ and update $S_{\phi^{\prime}}$. After $e_2$ iterations, we obtain the final ANSM $S_{\phi^{\star}}$.

\section{Experiments}
\label{sect:Experiment}
\subsection{Experimental Setup}
\subsubsection{Forgery Localization Models}
In order to verify the universality of our proposed method for forgery localization models with different architectures, we adopt the open-sourced state-of-the-art models, including three deep learning-based methods, MVSS-Net~\cite{dong2022mvss}, IF-OSN~\cite{wu2022robust}, HDF-Net~\cite{han2024hdf}, and a reinforcement learning-based method, CoDE~\cite{peng2024employing}.

\subsubsection{Adversarial Attack Methods for Testing}
We evaluate our proposed ANSM against six widely–used adversarial attack algorithms with white-box setting:
\begin{itemize}
	\item \textbf{FGSM}~\cite{goodfellow2014explaining}\
	A single-step attack that perturbs the input image along the sign of the gradient. It is widely used for testing adversarial robustness due to its simplicity and low computational cost.
	
	\item \textbf{C\&W}~\cite{carlini2017towards}\
	A targeted attack with the objective of misleading the predicted forgery mask to all zero, subject to an $L_2$ norm constraint on the perturbation magnitude.
	
	\item \textbf{BIM}~\cite{kurakin2018adversarial}\
	An iterative variant of FGSM that applies multiple small-step updates, offering stronger attack effect.
	
	\item \textbf{PGD}~\cite{madry2018towards}\
	A widely adopted first-order attack that starts from a randomly perturbed input image within a bounded region and applies iterative updates similar to BIM, serving as a benchmark for testing adversarial robustness.
	
	\item \textbf{MI-FGSM}~\cite{dong2018boosting}  
	An extension of BIM that incorporates a momentum term into the gradient calculation, stabilizing update directions and improving attack effect.
	
	\item \textbf{PGN}~\cite{ge2023boosting}\
	A gradient-regularized attack that penalizes the gradient norm during optimization, encouraging updates toward flatter region in the loss landscape.
\end{itemize}
Given the white-box setting, for the optimization-based attack C\&W, we vary the number of optimization steps $m$ across five settings, i.e., $m \in \{50, 100, 150, 200, 250\}$. For the rest, we set three levels of noise intensity $\varphi \in \{1/255, 2/255, 3/255\}$, corresponding to progressively stronger but still imperceptible input modifications.

\subsubsection{Datasets}
We adopt four widely-used benchmark datasets in image forgery localization task, including Columbia~\cite{ng2004data}, CASIAv2 and CASIAv1~\cite{dong2013casia}, IMD20~\cite{novozamsky2020imd2020}, and MISD~\cite{kadam2021multiple}. For each dataset, we construct its adversarial versions by applying aforementioned six adversarial attack algorithms.
\begin{itemize}
\item[$\bullet$] \textbf{COLUMBIA} contains 160 digitally manipulated images, specifically crafted via image splicing operations.

\item[$\bullet$] \textbf{CASIA} dataset includes two versions: v1 with 920 manipulated images and v2 with 5,123 images. The primary manipulation methods involve splicing and copy-move forgery, with additional processing such as filtering and blurring applied in some cases to obscure artifacts.

\item[$\bullet$] \textbf{IMD20} consists of 2,010 forged images derived from authentic images collected from online sources. The tampering strategies involves splicing, copy-move, and object removal.

\item[$\bullet$] \textbf{MISD} comprises 227 digitally spliced images generated using Figma software. Each manipulated image involves multi-source forgeries, with objects from multiple external images integrated onto a single base image.
\end{itemize}
During training ANSM, the datasets we used consist solely of the CASIAv2 dataset, including both its original version and adversarial version using FGSM attack ($\varphi=3/255$). The rest are used for testing, allowing for a comprehensive evaluation of our proposed ANSM's generalization ability.

\subsubsection{Metrics}
As with most prior works~\cite{dong2022mvss, wu2022robust, peng2024employing, han2024hdf}, we use pixel-level F1 to evaluate the forgery localization performance. Higher F1 indicates superior forgery localization performance. And to more clearly show the performance changes under adversarial attack and our defense, we propose another metric, RP (Residual Performance), which is the F1 ratio of input image to its original version (F1 on original forged image is denoted as 100\%). Higher RP indicates performance closer to the forgery localization model's original capability.
\begin{table}[t]\footnotesize
	\centering
	\renewcommand{\arraystretch}{0.9}
	\tabcolsep=0.08cm
	\caption{The settings to train ANSM for different forgery localization models.}
	\begin{tabular}{c c c c}
		\Xhline{2pt}
		Model & Input size & Learning rate & Maximum epoch \\
		\cmidrule(lr){1-1}\cmidrule(lr){2-2}\cmidrule(lr){3-3}\cmidrule(lr){4-4}
		\multirow{2}*{IF-OSN} 
		& Random cropping & $\gamma_{1}=5e-4$ & $e_1=1000$ \\
		& 256$\times$256 & $\gamma_{2}=1e-5$ & $e_2=20$\\
		\cmidrule(lr){1-1}\cmidrule(lr){2-2}\cmidrule(lr){3-3}\cmidrule(lr){4-4}
		\multirow{2}*{MVSS-Net} 
		& Resizing & $\gamma_{1}=5e-4$ & $e_1=200$ \\
		& 512$\times$512 & $\gamma_{2}=1e-5$ & $e_2=30$\\
		\cmidrule(lr){1-1}\cmidrule(lr){2-2}\cmidrule(lr){3-3}\cmidrule(lr){4-4}
		
		\multirow{2}*{HDF-Net} 
		& Resizing & $\gamma_{1}=5e-4$ & $e_1=200$\\
		& 256$\times$256 & $\gamma_{2}=1e-5$ & $e_2=10$\\
		\cmidrule(lr){1-1}\cmidrule(lr){2-2}\cmidrule(lr){3-3}\cmidrule(lr){4-4}
		
		\multirow{2}*{CoDE} 
		& Resizing & $\gamma_{1}=5e-4$ & $e_1=200$ \\
		& 512$\times$512 & $\gamma_{2}=1e-5$ & $e_2=30$\\
		\Xhline{2pt}
	\end{tabular}
	\label{table:training_settings}
\end{table}

\subsubsection{Implementation Details}
The proposed framework is implemented using PyTorch and trained on two Tesla A100 GPUs. The input size, learning rate, maximum epochs, and batch size settings used to train the ANSM for different forgery localization models are detailed in Table~\ref{table:training_settings}. The AdamW optimizer~\cite{kingma2014adam} is employed, along with the ReduceLROnPlateau scheduler, which reduces the learning rate by a factor of 0.9 if the loss fails to improve for 10 consecutive epochs.

\subsection{Quantitative Analysis}
\label{section:defense_effect}
\begin{table*}[t!]\footnotesize
	\centering
	\renewcommand{\arraystretch}{1.15}
	\tabcolsep=0.13cm
	\caption{Comparison of forgery localization performance with and without ANSM. The bold part indicates the performance change after using ANSM.}
	\label{table:defense_effect}
	\begin{tabular}{c c | c c | c c | c c | c c | c c | c c | c c}	
		\hline\hline
		\multirow{3}*{Model} & \multirow{3}*{ANSM} & \multicolumn{14}{c}{Columbia}\\
		\cline{3-16}
		& & \multicolumn{2}{c|}{Original} & \multicolumn{2}{c|}{FGSM} & \multicolumn{2}{c|}{C$\&$W} & \multicolumn{2}{c|}{BIM} & \multicolumn{2}{c|}{PGD} & \multicolumn{2}{c|}{MI-FGSM} & \multicolumn{2}{c}{PGN} \\
		& & F1 & RP(\%) & F1 & RP(\%) & F1 & RP(\%)  & F1 & RP(\%) & F1 & RP(\%) & F1 & RP(\%) & F1 & RP(\%) \\
		\hline
		
		\multirow{3}*{\makecell[c]{IF-OSN}} 
		& \ding{55} 
		& 0.706 & 100.0 
		& 0.225 & 31.9 
		& 0.163 & 23.1 
		& 0.068 & 9.6  
		& 0.078 & 11.0  
		& 0.044 & 6.2 
		& 0.038 & 5.4 
		\\ 
		& \ding{52} 
		& 0.693 & 98.2 
		& 0.723 & 102.4 
		& 0.711 & 100.7 
		& 0.621 & 88.0 
		& 0.654 & 92.6 
		& 0.640 & 90.7 
		& 0.676 & 95.8 
		\\
		& 
		& \textbf{-0.013} & \textbf{-1.8}
		& \textbf{+0.498} & \textbf{+70.5}
		& \textbf{+0.548} & \textbf{+77.6}
		& \textbf{+0.553} & \textbf{+78.4}
		& \textbf{+0.576} & \textbf{+81.6}
		& \textbf{+0.596} & \textbf{+84.5}
		& \textbf{+0.638} & \textbf{+90.4}
		\\
		\hline
		
		\multirow{3}*{\makecell[c]{MVSS-Net}} 
		& \ding{55} 
		& 0.677 & 100.0 
		& 0.122 & 18.0 
		& 0.215 & 31.8 
		& 0.018 & 2.7 
		& 0.021 & 3.1 
		& 0.017 & 2.5 
		& 0.028 & 4.1 
		\\
		& \ding{52} 
		& 0.611	& 90.3 
		& 0.559	& 82.6 
		& 0.605	& 89.4 
		& 0.548	& 80.9 
		& 0.561	& 82.9 
		& 0.554	& 81.8 
		& 0.547	& 80.8 
		\\
		& 
		& \textbf{-0.066} & \textbf{-9.7}
		& \textbf{+0.437} & \textbf{+64.6}
		& \textbf{+0.390} & \textbf{+57.6}
		& \textbf{+0.530} & \textbf{+78.2}
		& \textbf{+0.540} & \textbf{+79.8}
		& \textbf{+0.537} & \textbf{+79.3}
		& \textbf{+0.519} & \textbf{+76.7}
		\\
		\hline
		
		\multirow{3}*{\makecell[c]{HDF-Net}} 
		& \ding{55} 
		& 0.527 & 100.0 
		& 0.288 & 54.6 
		& 0.091 & 17.3 
		& 0.118 & 22.4 
		& 0.132 & 25.0 
		& 0.107 & 20.3 
		& 0.124 & 23.5 
		\\
		& \ding{52} 
		& 0.506 & 96.0 
		& 0.512 & 97.2 
		& 0.488 & 92.6 
		& 0.489 & 92.8 
		& 0.494 & 93.7 
		& 0.497 & 94.3 
		& 0.496 & 94.1 
		\\
		& 
		& \textbf{-0.021} & \textbf{-4.0}
		& \textbf{+0.224} & \textbf{+42.6}
		& \textbf{+0.397} & \textbf{+75.3}
		& \textbf{+0.371} & \textbf{+70.4}
		& \textbf{+0.362} & \textbf{+68.7}
		& \textbf{+0.390} & \textbf{+74.0}
		& \textbf{+0.372} & \textbf{+70.6}
		\\
		\hline
		
		\multirow{3}*{\makecell[c]{CoDE}} 
		& \ding{55} 
		& 0.881 & 100.0 
		& 0.132 & 15.0 
		& 0.079 & 9.0 
		& 0.005 & 0.6 
		& 0.008 & 0.9 
		& 0.009 & 1.0 
		& 0.040 & 4.5 
		\\
		& \ding{52} 
		& 0.852 & 96.7  
		& 0.895 & 101.6 
		& 0.848 & 96.3  
		& 0.827 & 93.9 
		& 0.828 & 94.0 
		& 0.858 & 97.4 
		& 0.839 & 95.2 
		\\
		& 
		& \textbf{-0.029} & \textbf{-3.7}
		& \textbf{+0.763} & \textbf{+86.6}
		& \textbf{+0.769} & \textbf{+87.3}
		& \textbf{+0.822} & \textbf{+93.3}
		& \textbf{+0.820} & \textbf{+93.1}
		& \textbf{+0.849} & \textbf{+96.4}
		& \textbf{+0.799} & \textbf{+90.7}
		\\
		\hline\hline
		
		\hline
		\multirow{3}*{Model} & \multirow{3}*{ANSM} & \multicolumn{14}{c}{CASIAv1}\\
		\cline{3-16}
		& & \multicolumn{2}{c|}{Original} & \multicolumn{2}{c|}{FGSM} & \multicolumn{2}{c|}{C$\&$W} & \multicolumn{2}{c|}{BIM} & \multicolumn{2}{c|}{PGD}& \multicolumn{2}{c|}{MI-FGSM} & \multicolumn{2}{c}{PGN} \\
		& & F1 & RP(\%) & F1 & RP(\%) & F1 & RP(\%)  & F1 & RP(\%) & F1 & RP(\%) & F1 & RP(\%) & F1 & RP(\%) \\
		\hline
		
		\multirow{3}*{\makecell[c]{IF-OSN }} 
		& \ding{55} 
		& 0.509 & 100.0 
		& 0.200 & 39.3 
		& 0.103 & 20.2 
		& 0.131 & 25.7 
		& 0.133 & 26.1 
		& 0.104 & 20.4 
		& 0.113 & 22.2 
		\\ 
		& \ding{52} 
		& 0.475 & 93.3 
		& 0.540 & 106.1 
		& 0.499 & 98.0  
		& 0.506 & 99.4 
		& 0.492 & 96.7 
		& 0.514 & 101.0 
		& 0.490 & 96.3 
		\\
		& 
		& \textbf{-0.034} & \textbf{-6.7}
		& \textbf{+0.340} & \textbf{+66.8}
		& \textbf{+0.396} & \textbf{+77.8}
		& \textbf{+0.375} & \textbf{+73.7}
		& \textbf{+0.359} & \textbf{+70.6}
		& \textbf{+0.410} & \textbf{+80.6} 
		& \textbf{+0.377} & \textbf{+74.1} 
		\\
		\hline
		
		\multirow{3}*{\makecell[c]{MVSS-Net}} 
		& \ding{55} 
		& 0.431 & 100.0 
		& 0.170 & 39.4 
		& 0.176 & 40.8 
		& 0.071 & 16.5 
		& 0.080 & 18.6 
		& 0.060 & 13.9 
		& 0.067 & 15.5 
		\\
		& \ding{52} 
		& 0.446	& 103.5 
		& 0.502 & 116.5 
		& 0.453 & 105.1 
		& 0.464 & 107.7 
		& 0.453 & 105.1 
		& 0.465 & 107.9 
		& 0.446 & 103.4 
		
		\\
		& 
		& \textbf{+0.015} & \textbf{+3.5}
		& \textbf{+0.332} & \textbf{+77.1}
		& \textbf{+0.277} & \textbf{+64.3}
		& \textbf{+0.393} & \textbf{+91.2}
		& \textbf{+0.373} & \textbf{+86.5}
		& \textbf{+0.395} & \textbf{+94.0}
		& \textbf{+0.379} & \textbf{+87.9}
		\\
		\hline
		
		\multirow{3}*{\makecell[c]{HDF-Net}} 
		& \ding{55} 
		& 0.339 & 100.0 
		& 0.160 & 47.2 
		& 0.108 & 31.9 
		& 0.057 & 16.8 
		& 0.081 & 23.9 
		& 0.067 & 19.8 
		& 0.068 & 20.1 
		\\
		& \ding{52} 
		& 0.334 & 98.5 
		& 0.373 & 110.0 
		& 0.302 & 90.4 
		& 0.379 & 99.7 
		& 0.351 & 97.6 
		& 0.379 & 111.8 
		& 0.351 & 103.5 
		\\
		& 
		& \textbf{-0.005} & \textbf{-1.5}
		& \textbf{+0.213} & \textbf{+62.8}
		& \textbf{+0.194} & \textbf{+58.5}
		& \textbf{+0.276} & \textbf{+82.9}
		& \textbf{+0.245} & \textbf{+73.7}
		& \textbf{+0.312} & \textbf{+92.0}
		& \textbf{+0.283} & \textbf{+83.4}
		\\
		\hline
		
		\multirow{3}*{\makecell[c]{CoDE}} 
		& \ding{55} 
		& 0.723 & 100.0 
		& 0.166 & 23.0 
		& 0.105 & 14.5 
		& 0.024 & 3.3  
		& 0.026 & 3.6  
		& 0.015 & 2.1  
		& 0.026 & 3.6  
		\\
		& \ding{52} 
		& 0.681 & 94.2  
		& 0.783 & 108.3 
		& 0.680 & 94.1 
		& 0.724 & 100.1 
		& 0.695 & 96.1  
		& 0.745 & 103.3  
		& 0.710 & 98.2  
		\\
		& 
		& \textbf{-0.042} & \textbf{-5.8}
		& \textbf{+0.617} & \textbf{+84.7}
		& \textbf{+0.575} & \textbf{+79.6}
		& \textbf{+0.700} & \textbf{+96.8}
		& \textbf{+0.669} & \textbf{+92.5}
		& \textbf{+0.730} & \textbf{+101.2}
		& \textbf{+0.684} & \textbf{+94.6}
		\\
		\hline\hline
		
		\hline
		\multirow{3}*{Model} & \multirow{3}*{ANSM} & \multicolumn{14}{c}{IMD20}\\
		\cline{3-16}
		& & \multicolumn{2}{c|}{Original} & \multicolumn{2}{c|}{FGSM} & \multicolumn{2}{c|}{C$\&$W} & \multicolumn{2}{c|}{BIM} & \multicolumn{2}{c|}{PGD}& \multicolumn{2}{c|}{MI-FGSM} & \multicolumn{2}{c}{PGN} \\
		& & F1 & RP(\%) & F1 & RP(\%) & F1 & RP(\%)  & F1 & RP(\%) & F1 & RP(\%) & F1 & RP(\%) & F1 & RP(\%) \\
		\hline
		
		\multirow{3}*{\makecell[c]{IF-OSN}} 
		& \ding{55} 
		& 0.488 & 100.0 
		& 0.138 & 28.3 
		& 0.226 & 46.3 
		& 0.084 & 17.2 
		& 0.087 & 17.8 
		& 0.067 & 13.7 
		& 0.061 & 12.5 
		\\ 
		& \ding{52} 
		& 0.461 & 94.5 
		& 0.471 & 96.5 
		& 0.461 & 94.5 
		& 0.444 & 91.0 
		& 0.443 & 90.8 
		& 0.438 & 89.8 
		& 0.436 & 89.3 
		\\
		& 
		& \textbf{-0.027} & \textbf{-5.5}
		& \textbf{+0.333} & \textbf{+68.2}
		& \textbf{+0.235} & \textbf{+48.2}
		& \textbf{+0.360} & \textbf{+73.8}
		& \textbf{+0.356} & \textbf{+73.0}
		& \textbf{+0.371} & \textbf{+76.1} 
		& \textbf{+0.375} & \textbf{+76.8} 
		\\
		\hline
		
		\multirow{3}*{\makecell[c]{MVSS-Net}} 
		& \ding{55} 
		& 0.245 & 100.0 
		& 0.079 & 32.2 
		& 0.057 & 23.3 
		& 0.038 & 15.5 
		& 0.039 & 15.9 
		& 0.037 & 15.1 
		& 0.044 & 18.0 
		\\
		& \ding{52} 
		& 0.242 & 98.8 
		& 0.227 & 92.7 
		& 0.243 & 99.2 
		& 0.250 & 102.0 
		& 0.245 & 100.0 
		& 0.238 & 97.1 
		& 0.227 & 92.7 
		\\
		& 
		& \textbf{+0.010} & \textbf{+4.1}
		& \textbf{+0.148} & \textbf{+60.5}
		& \textbf{+0.186} & \textbf{+75.9}
		& \textbf{+0.212} & \textbf{+86.5}
		& \textbf{+0.206} & \textbf{+84.1}
		& \textbf{+0.201} & \textbf{+82.0}
		& \textbf{+0.183} & \textbf{+84.7}
		\\
		\hline
		
		\multirow{3}*{\makecell[c]{HDF-Net}} 
		& \ding{55} 
		& 0.341 & 100.0 
		& 0.193 & 56.6 
		& 0.167 & 49.0 
		& 0.099 & 29.0 
		& 0.104 & 30.5 
		& 0.090 & 26.4 
		& 0.090 & 26.4 
		\\
		& \ding{52} 
		& 0.333 & 97.7 
		& 0.322 & 94.4 
		& 0.329 & 96.5 
		& 0.298 & 87.4 
		& 0.308 & 90.3 
		& 0.320 & 93.8 
		& 0.316 & 92.7 
		\\
		& 
		& \textbf{-0.008} & \textbf{-2.3} 
		& \textbf{+0.129} & \textbf{+37.8}
		& \textbf{+0.162} & \textbf{+47.5}
		& \textbf{+0.199} & \textbf{+58.4}
		& \textbf{+0.204} & \textbf{+59.8}
		& \textbf{+0.230} & \textbf{+67.4}
		& \textbf{+0.226} & \textbf{+66.3}
		\\
		\hline
		
		\multirow{3}*{\makecell[c]{CoDE}} 
		& \ding{55} 
		& 0.742 & 100.0 
		& 0.175 & 23.6 
		& 0.045 & 6.1 
		& 0.024 & 3.2 
		& 0.025 & 3.4 
		& 0.020 & 2.7 
		& 0.031 & 4.2 
		\\
		& \ding{52} 
		& 0.707 & 95.3 
		& 0.707 & 95.3 
		& 0.708 & 95.4 
		& 0.702 & 94.6 
		& 0.701 & 94.5 
		& 0.694 & 93.5 
		& 0.694 & 93.5 
		\\
		& 
		& \textbf{-0.035} & \textbf{-4.7}
		& \textbf{+0.632} & \textbf{+71.7}
		& \textbf{+0.663} & \textbf{+89.3}
		& \textbf{+0.678} & \textbf{+91.4}
		& \textbf{+0.676} & \textbf{+91.1}
		& \textbf{+0.674} & \textbf{+90.8}
		& \textbf{+0.663} & \textbf{+89.3}
		\\
		\hline\hline	
		
		\hline
		\multirow{3}*{Model} & \multirow{3}*{ANSM} & \multicolumn{14}{c}{MISD}\\
		\cline{3-16}
		& & \multicolumn{2}{c|}{Original} & \multicolumn{2}{c|}{FGSM} & \multicolumn{2}{c|}{C$\&$W} & \multicolumn{2}{c|}{BIM} & \multicolumn{2}{c|}{PGD}& \multicolumn{2}{c|}{MI-FGSM} & \multicolumn{2}{c}{PGN} \\
		& & F1 & RP(\%) & F1 & RP(\%) & F1 & RP(\%)  & F1 & RP(\%) & F1 & RP(\%) & F1 & RP(\%) & F1 & RP(\%) \\
		\hline
		
		\multirow{3}*{\makecell[c]{IF-OSN }} 
		& \ding{55} 
		& 0.674 & 100.0 
		& 0.414 & 61.4 
		& 0.140 & 20.8  
		& 0.359 & 53.3 
		& 0.366 & 54.3 
		& 0.319 & 47.3 
		& 0.296 & 43.9 
		\\ 
		& \ding{52} 
		& 0.630 & 93.5 
		& 0.640 & 95.0 
		& 0.631 & 93.6 
		& 0.633 & 93.9 
		& 0.622 & 92.3 
		& 0.625 & 92.7 
		& 0.611 & 90.7 
		\\
		& 
		& \textbf{-0.044} & \textbf{-6.5}
		& \textbf{+0.226} & \textbf{+33.6}
		& \textbf{+0.491} & \textbf{+72.8}
		& \textbf{+0.274} & \textbf{+40.6}
		& \textbf{+0.256} & \textbf{+38.0}
		& \textbf{+0.306} & \textbf{+45.4}
		& \textbf{+0.315} & \textbf{+46.8}
		\\
		\hline
		
		\multirow{3}*{\makecell[c]{MVSS-Net}} 
		& \ding{55} 
		& 0.617 & 100.0 
		& 0.372 & 60.3 
		& 0.210 & 34.0  
		& 0.225 & 36.5 
		& 0.222 & 36.0 
		& 0.197	& 31.9  
		& 0.211	& 34.2  
		\\
		& \ding{52} 
		& 0.596	& 96.6 
		& 0.591	& 95.8  
		& 0.593	& 96.1  
		& 0.595	& 96.4  
		& 0.582	& 94.3  
		& 0.584	& 94.7  
		& 0.563	& 91.2  
		\\
		& 
		& \textbf{-0.021} & \textbf{-3.4}
		& \textbf{+0.219} & \textbf{+35.5}
		& \textbf{+0.383} & \textbf{+62.1}
		& \textbf{+0.370} & \textbf{+59.9}
		& \textbf{+0.360} & \textbf{+58.3}
		& \textbf{+0.387} & \textbf{+62.8}
		& \textbf{+0.352} & \textbf{+57.0}
		\\
		\hline
		
		\multirow{3}*{\makecell[c]{HDF-Net}} 
		& \ding{55} 
		& 0.585 & 100.0  
		& 0.322 & 55.0 
		& 0.170 & 29.1 
		& 0.138 & 23.6 
		& 0.160 & 27.4 
		& 0.149 & 25.5 
		& 0.175 & 29.9 
		\\
		& \ding{52} 
		& 0.559 & 95.6 
		& 0.613 & 104.8 
		& 0.534 & 91.3 
		& 0.574 & 98.1 
		& 0.566 & 96.8 
		& 0.627 & 107.2 
		& 0.583 & 99.7 
		\\
		& 
		& \textbf{-0.026} & \textbf{-4.4}
		& \textbf{+0.291} & \textbf{+49.8}
		& \textbf{+0.364} & \textbf{+62.2}
		& \textbf{+0.436} & \textbf{+74.5}
		& \textbf{+0.406} & \textbf{+69.4}
		& \textbf{+0.478} & \textbf{+81.7}
		& \textbf{+0.408} & \textbf{+69.8}
		\\
		\hline
		
		\multirow{3}*{\makecell[c]{CoDE}} 
		& \ding{55} 
		& 0.733 & 100.0 
		& 0.319 & 43.5 
		& 0.084 & 11.5 
		& 0.044 & 6.0 
		& 0.048 & 6.5 
		& 0.045 & 6.1 
		& 0.050 & 6.8 
		\\
		& \ding{52} 
		& 0.673 & 91.8 
		& 0.770 & 105.0 
		& 0.679 & 92.6 
		& 0.677 & 92.4 
		& 0.668 & 91.1  
		& 0.749 & 102.1 
		& 0.695 & 94.8  
		\\
		& 
		& \textbf{-0.060} & \textbf{-8.2}
		& \textbf{+0.451} & \textbf{+61.5}
		& \textbf{+0.595} & \textbf{+81.1}
		& \textbf{+0.633} & \textbf{+86.4}
		& \textbf{+0.620} & \textbf{+84.6}
		& \textbf{+0.704} & \textbf{+96.0}
		& \textbf{+0.645} & \textbf{+88.0}
		\\
		\hline\hline
	\end{tabular}
\end{table*}
\begin{figure*}[t!]\small
	\centering
	\renewcommand{\arraystretch}{0.85}
	\def\fig_height{0.8cm}
	\setlength\tabcolsep{0.7mm}
	\def\new_hspace{-0.6cm}
	\hspace{\new_hspace}
	\subfloat[IF-OSN]{
		\centering
		\def\model{IF-OSN}
		\def\figa{Sp_D_ani_00031_ani_00039_sec_00095_181.png}
		\def\figb{Sp_D_CNN_C_ani0068_ani0045_0099.png}
		\def\figc{Sp_D_CNN_R_art0044_art0018_0278.png}
		\def\leftmargin{-120}
		\def\topmargin{20}
		\begin{tabular}{l c c c c c c c c c c}
			$\quad $ &
			\begin{overpic}[height=\fig_height]{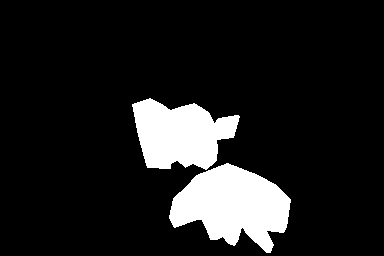}
				\put(\leftmargin,\topmargin){\small $\bm{M_{gt}}$}
			\end{overpic}&
			\begin{overpic}[height=\fig_height]{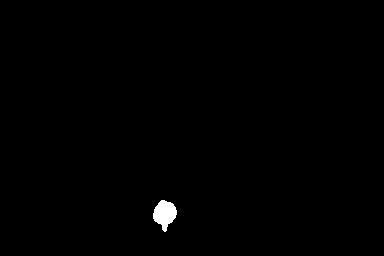}
			\end{overpic}&
			\begin{overpic}[height=\fig_height]{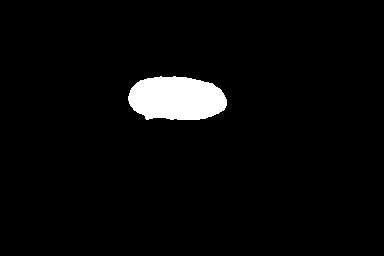}
			\end{overpic}&
			\vspace{0.05cm}
			\\
			$\quad $ &
			\begin{overpic}[height=\fig_height]{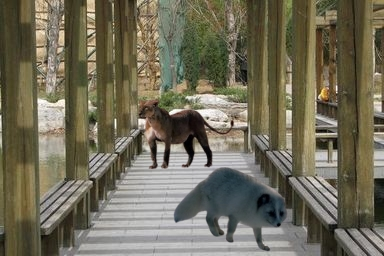}
				\put(\leftmargin,\topmargin){\small $\bm{X_{o}}$}
			\end{overpic}&
			\begin{overpic}[height=\fig_height]{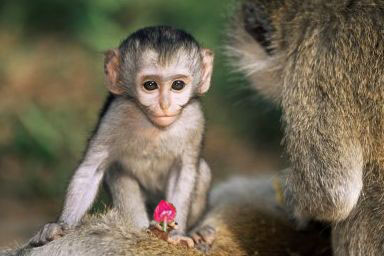}
			\end{overpic}&
			\begin{overpic}[height=\fig_height]{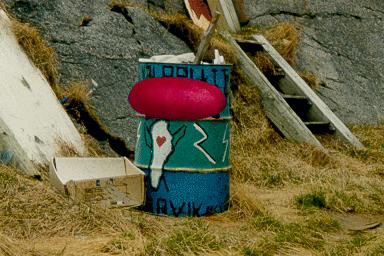}
			\end{overpic}&
			\\\\
			\cdashline{2-4}\\
			$\ $ &
			\begin{overpic}[height=\fig_height]{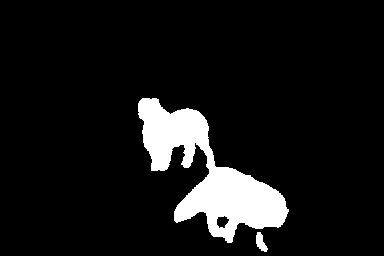}
				\put(\leftmargin,\topmargin){\small $\bm{M_{o}}$}
			\end{overpic}&
			\begin{overpic}[height=\fig_height]{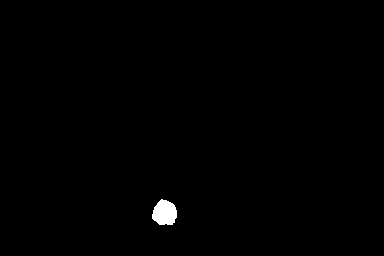}
			\end{overpic}&
			\begin{overpic}[height=\fig_height]{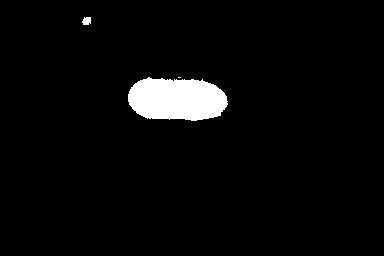}
			\end{overpic}&
			\vspace{0.05cm}
			\\
			$\ $ &
			\begin{overpic}[height=\fig_height]{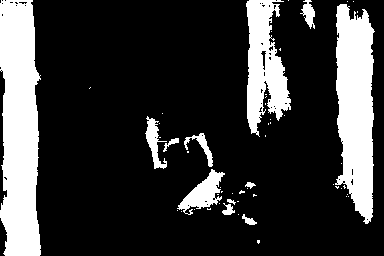}
				\put(\leftmargin,\topmargin){\small $\bm{M_{a}^{\text{FGSM}}}$}
			\end{overpic}&
			\begin{overpic}[height=\fig_height]{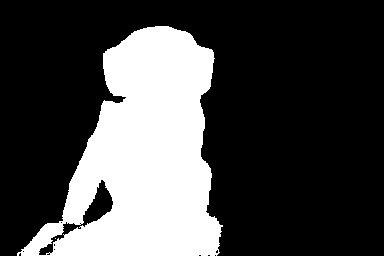}
			\end{overpic}&
			\begin{overpic}[height=\fig_height]{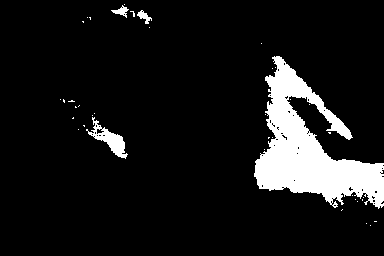}
			\end{overpic}&
			\vspace{0.05cm}
			\\
			$\ $ &
			\begin{overpic}[height=\fig_height]{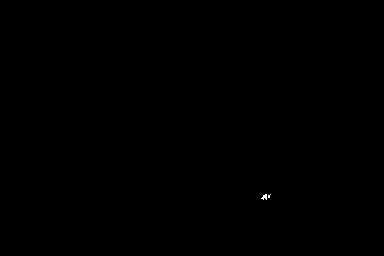}
				\put(\leftmargin,\topmargin){\small $\bm{M_{a}^{\text{C\&W}}}$}
			\end{overpic}&
			\begin{overpic}[height=\fig_height]{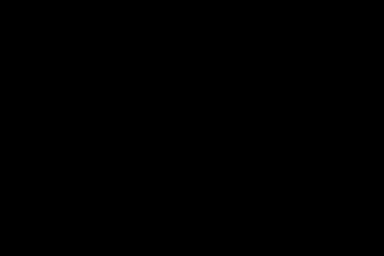}
			\end{overpic}&
			\begin{overpic}[height=\fig_height]{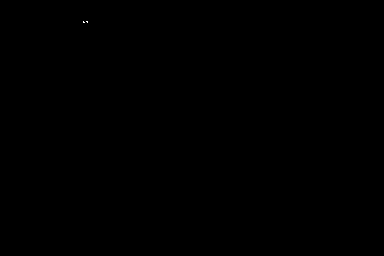}
			\end{overpic}&
			\vspace{0.05cm}
			\\
			$\ $ &
			\begin{overpic}[height=\fig_height]{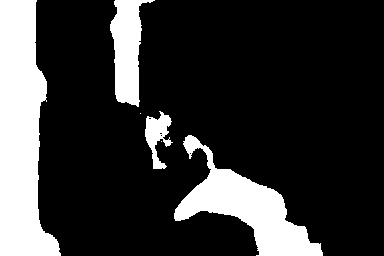}
				\put(\leftmargin,\topmargin){\small $\bm{M_{a}^{\text{BIM}}}$}
			\end{overpic}&
			\begin{overpic}[height=\fig_height]{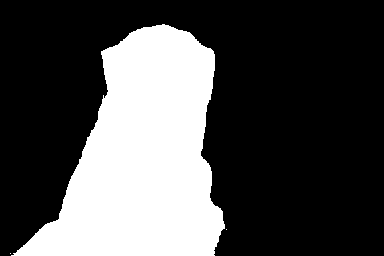}
			\end{overpic}&
			\begin{overpic}[height=\fig_height]{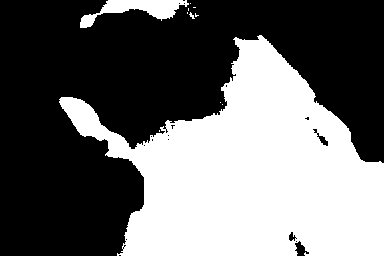}
			\end{overpic}&
			\vspace{0.05cm}
			\\
			$\ $ &
			\begin{overpic}[height=\fig_height]{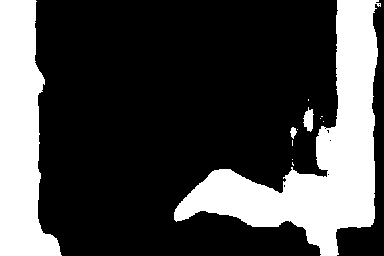}
				\put(\leftmargin,\topmargin){\small $\bm{M_{a}^{\text{PGD}}}$}
			\end{overpic}&
			\begin{overpic}[height=\fig_height]{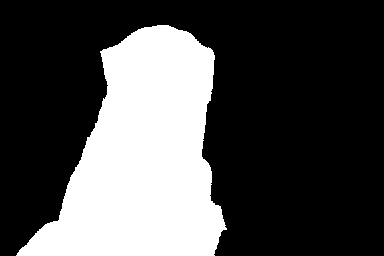}
			\end{overpic}&
			\begin{overpic}[height=\fig_height]{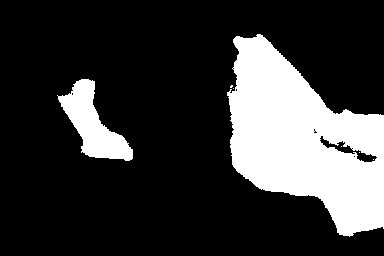}
			\end{overpic}&
			\vspace{0.05cm}
			\\
			$\ $ &
			\begin{overpic}[height=\fig_height]{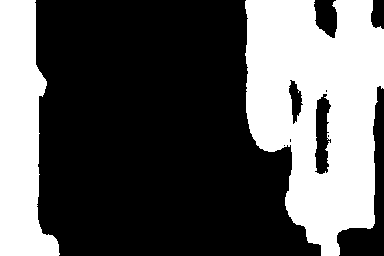}
				\put(\leftmargin,\topmargin){\small $\bm{M_{a}^{\text{MI-FGSM}}}$}
			\end{overpic}&
			\begin{overpic}[height=\fig_height]{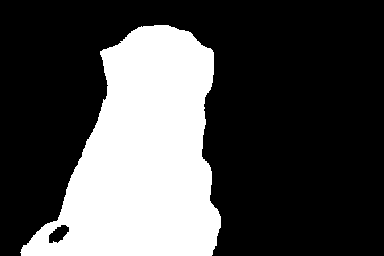}
			\end{overpic}&
			\begin{overpic}[height=\fig_height]{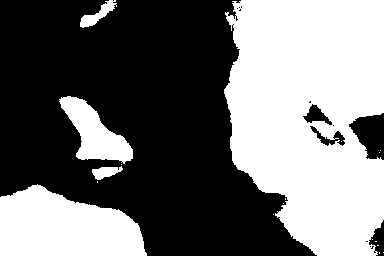}
			\end{overpic}&
			\vspace{0.05cm}
			\\
			$\ $ &
			\begin{overpic}[height=\fig_height]{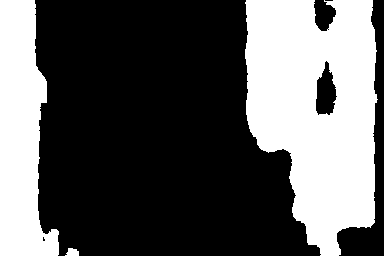}
				\put(\leftmargin,\topmargin){\small $\bm{M_{a}^{\text{PGN}}}$}
			\end{overpic}&
			\begin{overpic}[height=\fig_height]{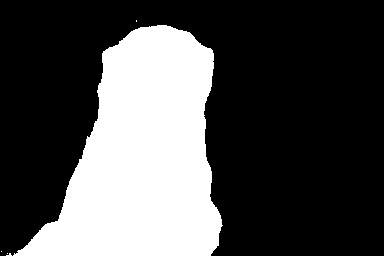}
			\end{overpic}&
			\begin{overpic}[height=\fig_height]{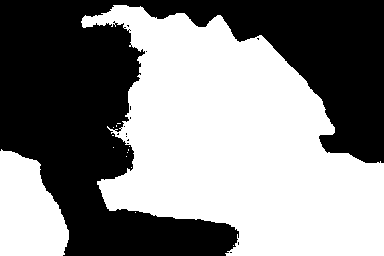}
			\end{overpic}&
			\\\\
			\cdashline{2-4}\\
			$\quad $ &
			\begin{overpic}[height=\fig_height]{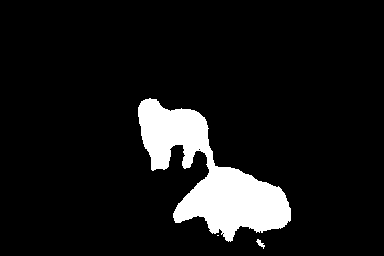}
				\put(\leftmargin,\topmargin){\small $\bm{M_{po}}$}
			\end{overpic}&
			\begin{overpic}[height=\fig_height]{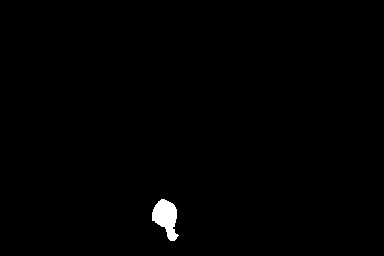}
			\end{overpic}&
			\begin{overpic}[height=\fig_height]{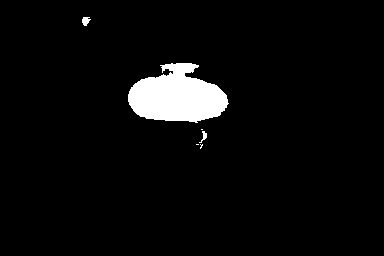}
			\end{overpic}&
			\vspace{0.05cm}
			\\
			$\ \ \ $ &
			\begin{overpic}[height=\fig_height]{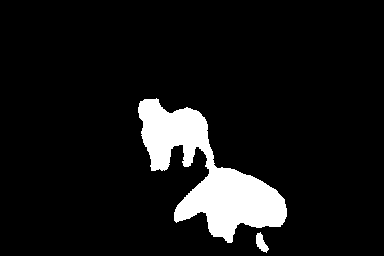}
				\put(\leftmargin,\topmargin){\small $\bm{M_{pa}^{\text{FGSM}}}$}
			\end{overpic}&
			\begin{overpic}[height=\fig_height]{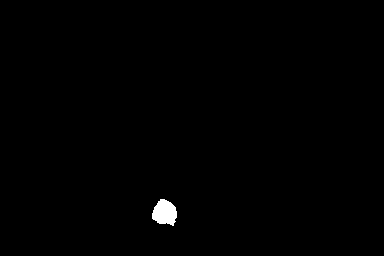}
			\end{overpic}&
			\begin{overpic}[height=\fig_height]{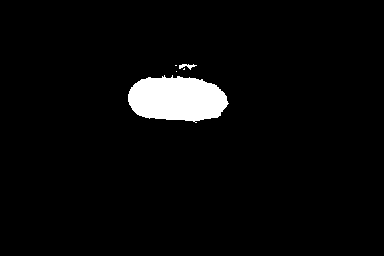}
			\end{overpic}&
			\vspace{0.05cm}
			\\
			$\ \ \ $ &
			\begin{overpic}[height=\fig_height]{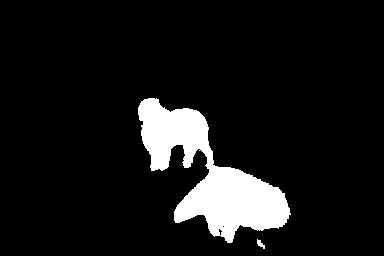}
				\put(\leftmargin,\topmargin){\small $\bm{M_{pa}^{\text{C\&W}}}$}
			\end{overpic}&
			\begin{overpic}[height=\fig_height]{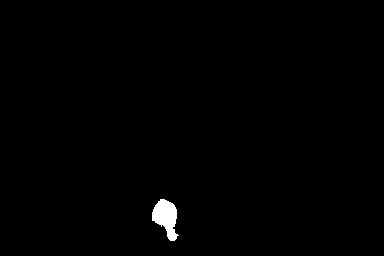}
			\end{overpic}&
			\begin{overpic}[height=\fig_height]{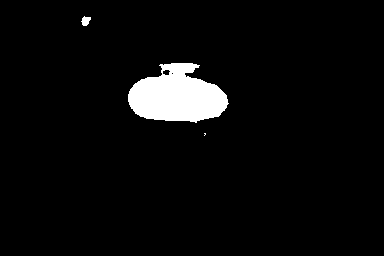}
			\end{overpic}&
			\vspace{0.05cm}
			\\
			$\ \ \ $ &
			\begin{overpic}[height=\fig_height]{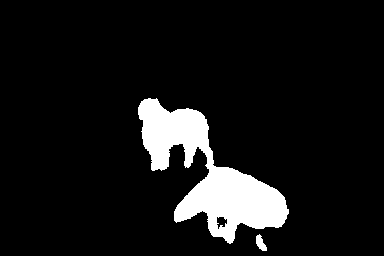}
				\put(\leftmargin,\topmargin){\small $\bm{M_{pa}^{\text{BIM}}}$}
			\end{overpic}&
			\begin{overpic}[height=\fig_height]{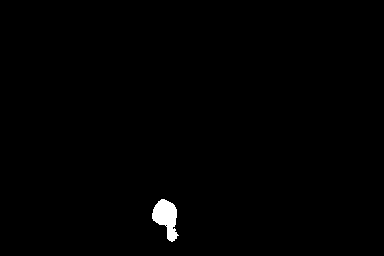}
			\end{overpic}&
			\begin{overpic}[height=\fig_height]{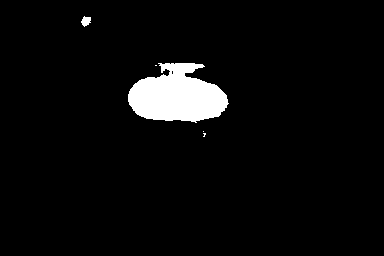}
			\end{overpic}&
			\vspace{0.05cm}
			\\
			$\ \ \ $ &
			\begin{overpic}[height=\fig_height]{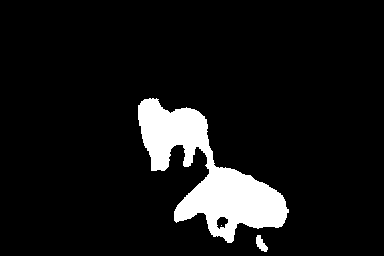}
				\put(\leftmargin,\topmargin){\small $\bm{M_{pa}^{\text{PGD}}}$}
			\end{overpic}&
			\begin{overpic}[height=\fig_height]{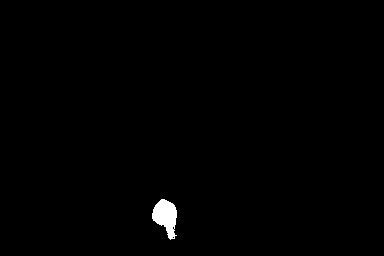}
			\end{overpic}&
			\begin{overpic}[height=\fig_height]{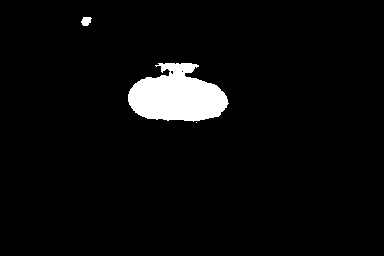}
			\end{overpic}&
			\vspace{0.05cm}
			\\
			$\ \ \ $ &
			\begin{overpic}[height=\fig_height]{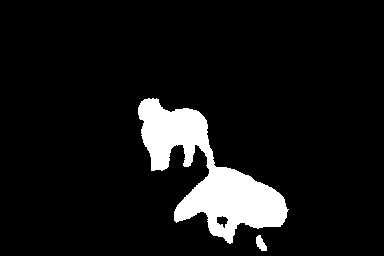}
				\put(\leftmargin,\topmargin){\small $\bm{M_{pa}^{\text{MI-FGSM}}}$}
			\end{overpic}&
			\begin{overpic}[height=\fig_height]{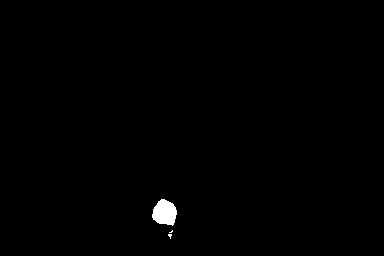}
			\end{overpic}&
			\begin{overpic}[height=\fig_height]{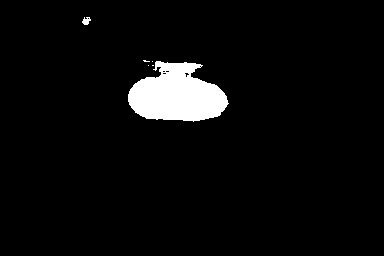}
			\end{overpic}&
			\vspace{0.05cm}
			\\
			$\ \ \ $ &
			\begin{overpic}[height=\fig_height]{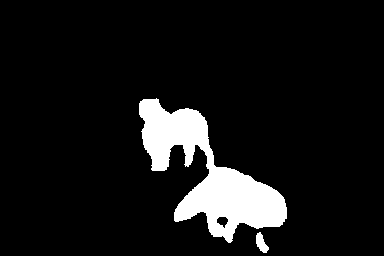}
				\put(\leftmargin,\topmargin){\small $\bm{M_{pa}^{\text{PGN}}}$}
			\end{overpic}&
			\begin{overpic}[height=\fig_height]{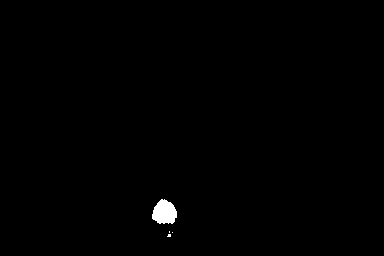}
			\end{overpic}&
			\begin{overpic}[height=\fig_height]{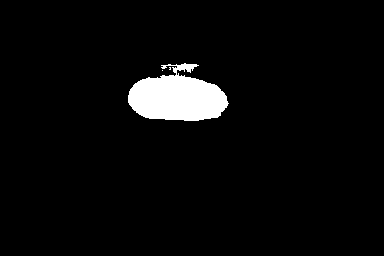}
			\end{overpic}
		\end{tabular}
	}
	\hspace{\new_hspace}
	\subfloat[MVSS-Net]{
		\def\model{MVSS-Net}
		\def\figa{Sp_D_ani_00027_ani_00067_ani_00052_117.png}
		\def\figb{Sp_D_CNN_A_ani0049_ani0084_0266.png}
		\def\figc{Sp_D_CNN_A_art0024_ani0032_0268.png}
		\begin{tabular}{c c c c c c c c c c}
			\begin{overpic}[height=\fig_height]{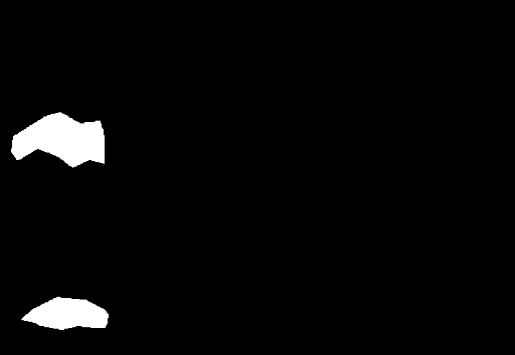}
			\end{overpic}&
			\begin{overpic}[height=\fig_height]{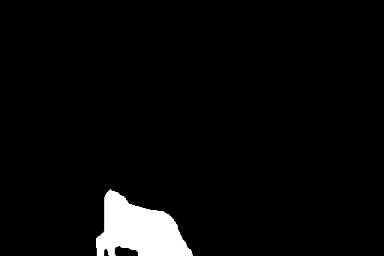}
			\end{overpic}&
			\begin{overpic}[height=\fig_height]{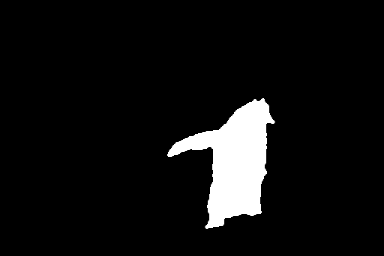}
			\end{overpic}&
			\vspace{0.05cm}
			\\
			\begin{overpic}[height=\fig_height]{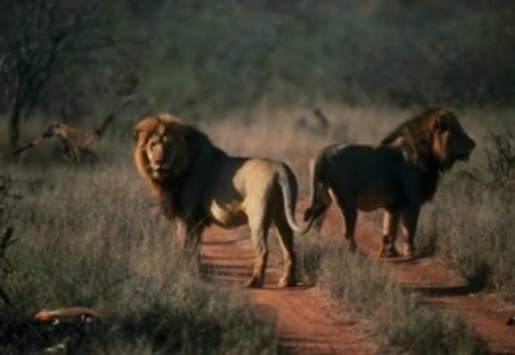}
			\end{overpic}&
			\begin{overpic}[height=\fig_height]{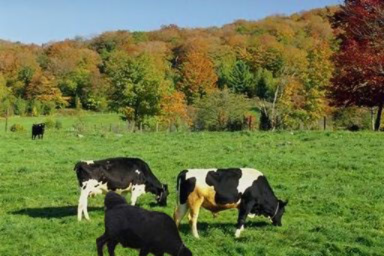}
			\end{overpic}&
			\begin{overpic}[height=\fig_height]{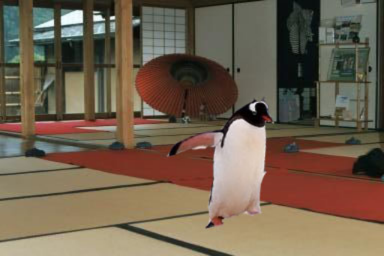}
			\end{overpic}&
			\\\\
			\cdashline{1-3}\\
			\begin{overpic}[height=\fig_height]{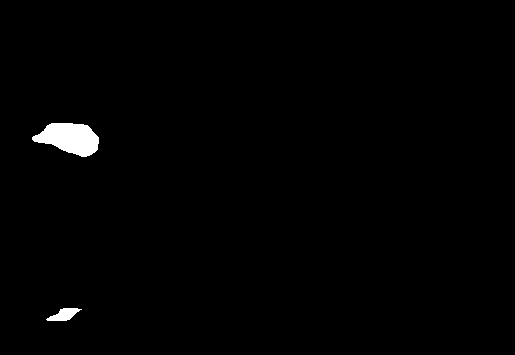}
			\end{overpic}&
			\begin{overpic}[height=\fig_height]{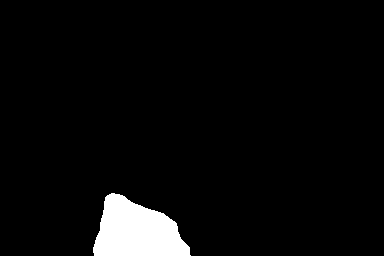}
			\end{overpic}&
			\begin{overpic}[height=\fig_height]{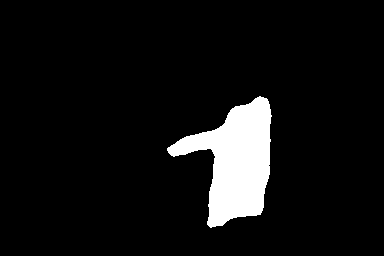}
			\end{overpic}&
			\vspace{0.05cm}
			\\
			\begin{overpic}[height=\fig_height]{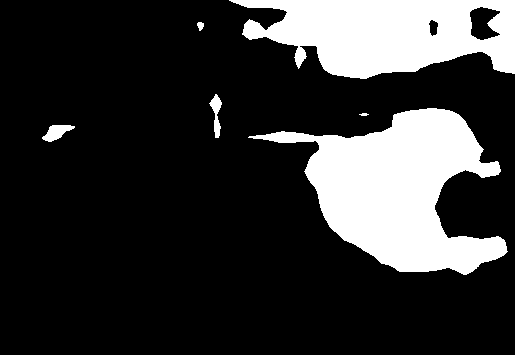}
			\end{overpic}&
			\begin{overpic}[height=\fig_height]{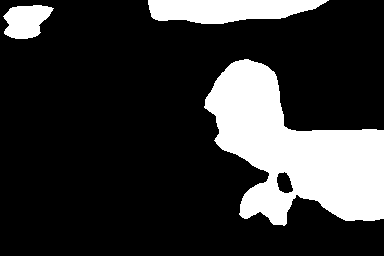}
			\end{overpic}&
			\begin{overpic}[height=\fig_height]{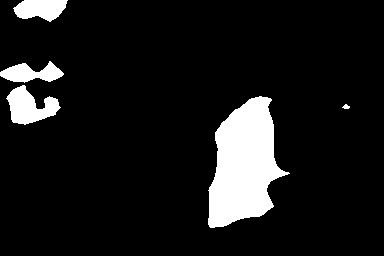}
			\end{overpic}&
			\vspace{0.05cm}
			\\
			\begin{overpic}[height=\fig_height]{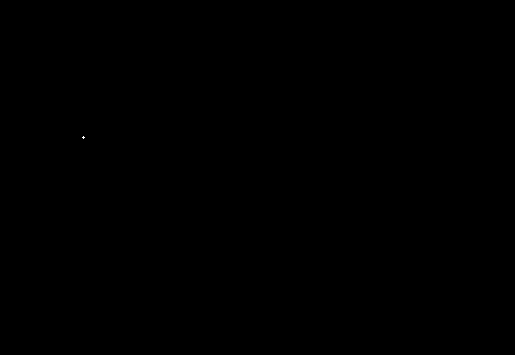}
			\end{overpic}&
			\begin{overpic}[height=\fig_height]{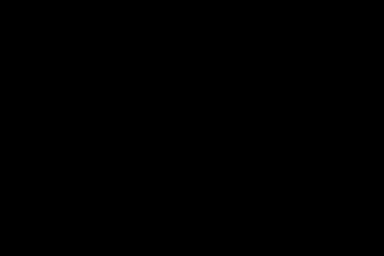}
			\end{overpic}&
			\begin{overpic}[height=\fig_height]{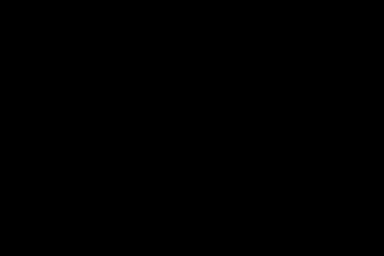}
			\end{overpic}&
			\vspace{0.05cm}
			\\
			\begin{overpic}[height=\fig_height]{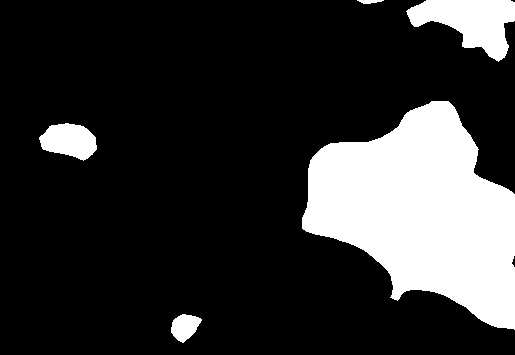}
			\end{overpic}&
			\begin{overpic}[height=\fig_height]{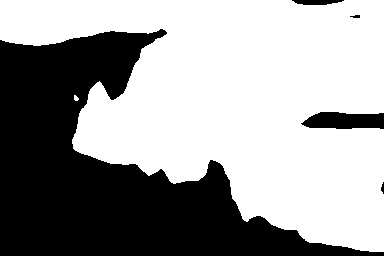}
			\end{overpic}&
			\begin{overpic}[height=\fig_height]{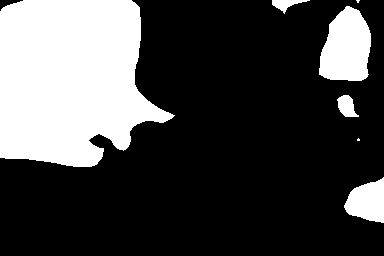}
			\end{overpic}&
			\vspace{0.05cm}
			\\
			\begin{overpic}[height=\fig_height]{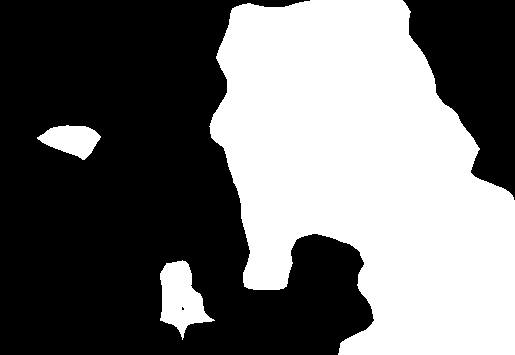}
			\end{overpic}&
			\begin{overpic}[height=\fig_height]{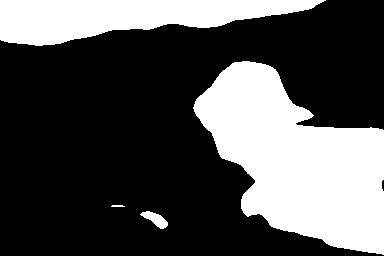}
			\end{overpic}&
			\begin{overpic}[height=\fig_height]{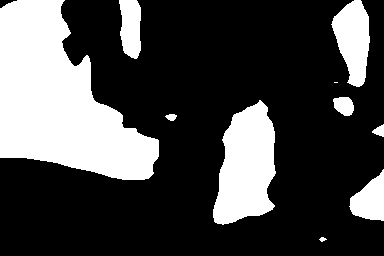}
			\end{overpic}&
			\vspace{0.05cm}
			\\
			\begin{overpic}[height=\fig_height]{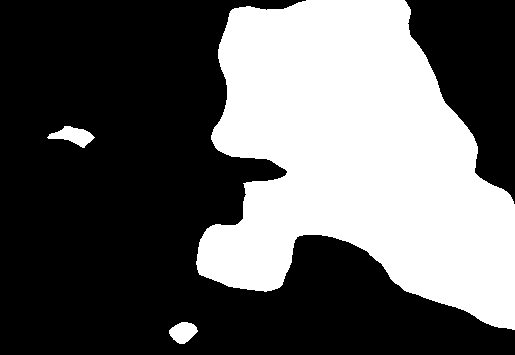}
			\end{overpic}&
			\begin{overpic}[height=\fig_height]{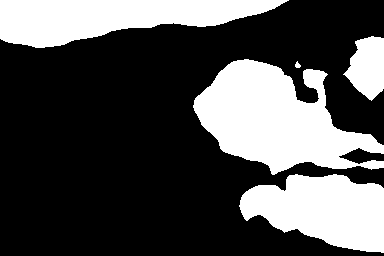}
			\end{overpic}&
			\begin{overpic}[height=\fig_height]{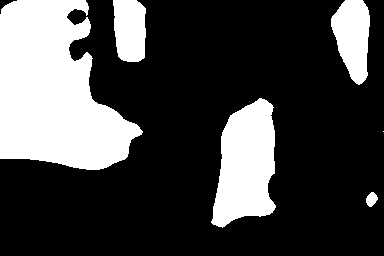}
			\end{overpic}&
			\vspace{0.05cm}
			\\
			\begin{overpic}[height=\fig_height]{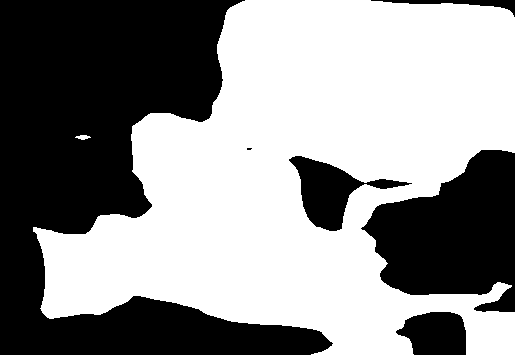}
			\end{overpic}&
			\begin{overpic}[height=\fig_height]{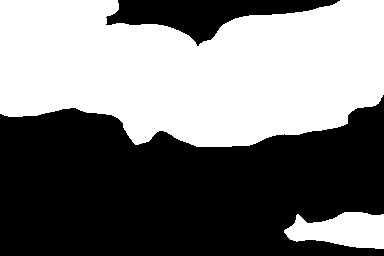}
			\end{overpic}&
			\begin{overpic}[height=\fig_height]{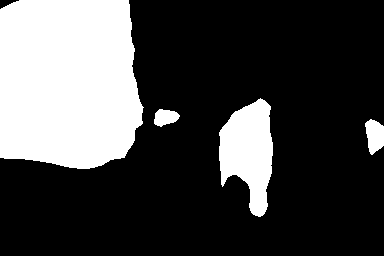}
			\end{overpic}&
			\\\\
			\cdashline{1-3}\\
			\begin{overpic}[height=\fig_height]{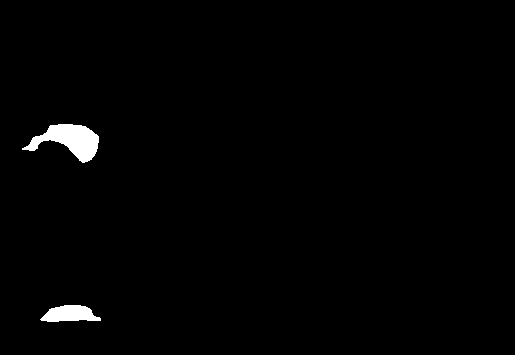}
			\end{overpic}&
			\begin{overpic}[height=\fig_height]{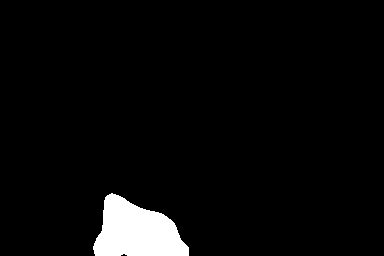}
			\end{overpic}&
			\begin{overpic}[height=\fig_height]{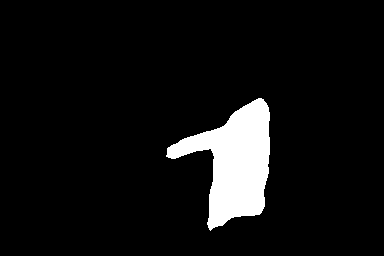}
			\end{overpic}&
			\vspace{0.05cm}
			\\
			\begin{overpic}[height=\fig_height]{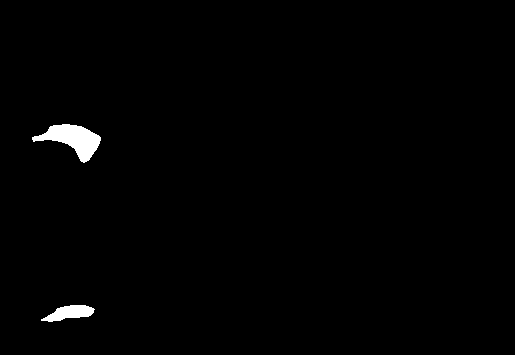}
			\end{overpic}&
			\begin{overpic}[height=\fig_height]{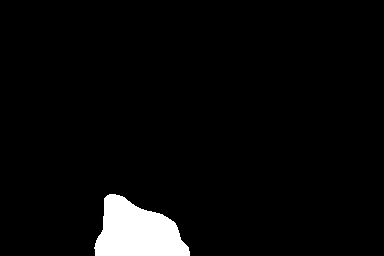}
			\end{overpic}&
			\begin{overpic}[height=\fig_height]{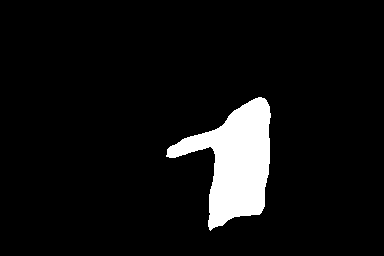}
			\end{overpic}&
			\vspace{0.05cm}
			\\
			\begin{overpic}[height=\fig_height]{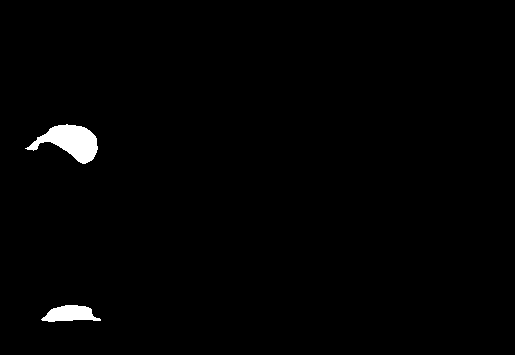}
			\end{overpic}&
			\begin{overpic}[height=\fig_height]{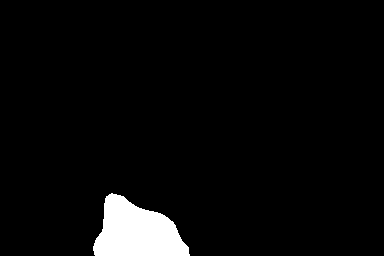}
			\end{overpic}&
			\begin{overpic}[height=\fig_height]{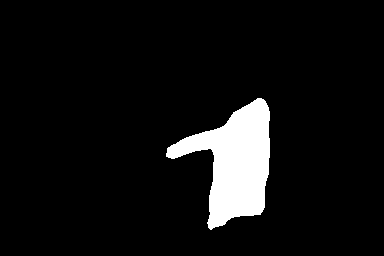}
			\end{overpic}&
			\vspace{0.05cm}
			\\
			\begin{overpic}[height=\fig_height]{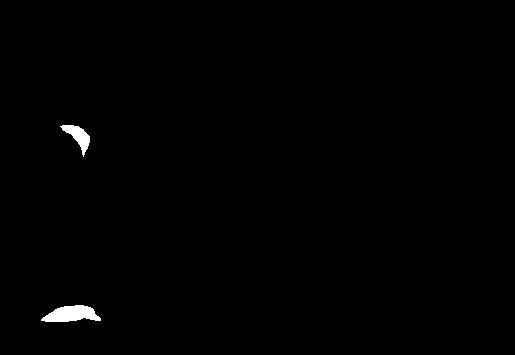}
			\end{overpic}&
			\begin{overpic}[height=\fig_height]{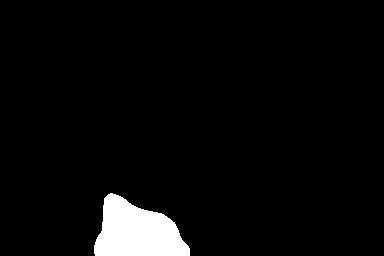}
			\end{overpic}&
			\begin{overpic}[height=\fig_height]{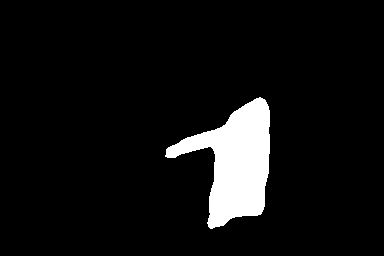}
			\end{overpic}&
			\vspace{0.05cm}
			\\
			\begin{overpic}[height=\fig_height]{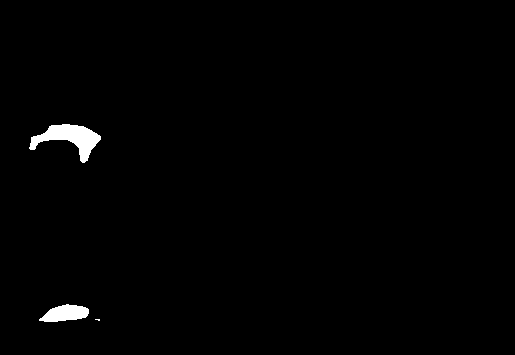}
			\end{overpic}&
			\begin{overpic}[height=\fig_height]{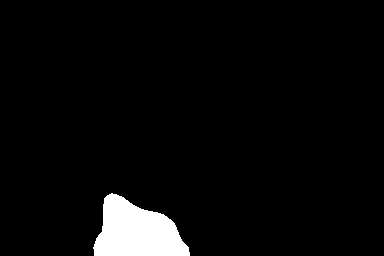}
			\end{overpic}&
			\begin{overpic}[height=\fig_height]{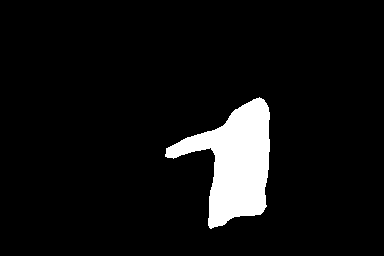}
			\end{overpic}&
			\vspace{0.05cm}
			\\
			\begin{overpic}[height=\fig_height]{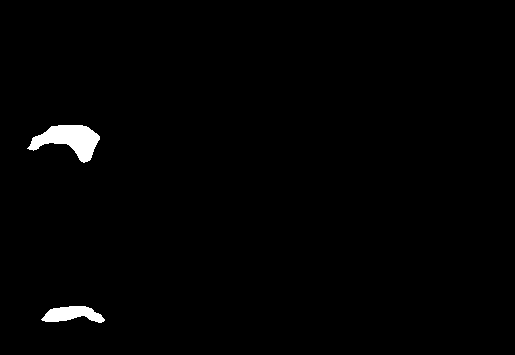}
			\end{overpic}&
			\begin{overpic}[height=\fig_height]{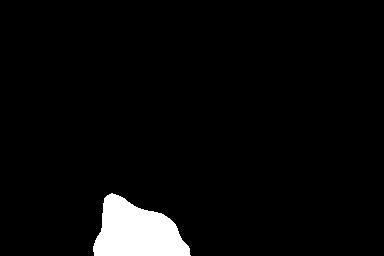}
			\end{overpic}&
			\begin{overpic}[height=\fig_height]{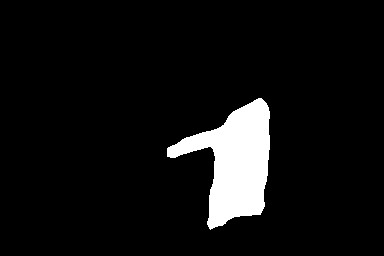}
			\end{overpic}&
			\vspace{0.05cm}
			\\
			\begin{overpic}[height=\fig_height]{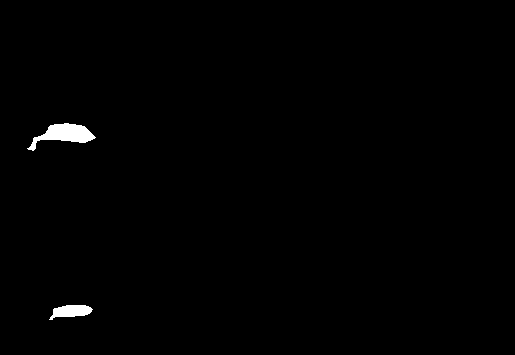}
			\end{overpic}&
			\begin{overpic}[height=\fig_height]{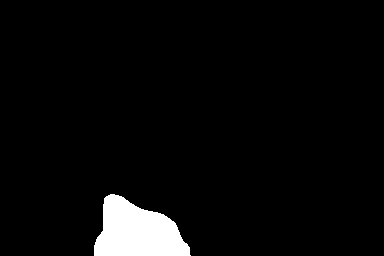}
			\end{overpic}&
			\begin{overpic}[height=\fig_height]{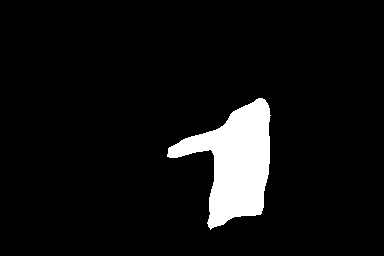}
			\end{overpic}&
		\end{tabular}
	}
	\hspace{\new_hspace}
	\subfloat[HDF-Net]{
		\def\model{HDF-Net}
		\def\figa{Sp_D_ani_00064_sec_00052_nat_00080_175.png}
		\def\figb{Sp_D_CRN_A_arc0064_cha0053_0377.png}
		\def\figc{canong3_canonxt_sub_13.png}
		\begin{tabular}{c c c c c c c c c c}
			\begin{overpic}[height=\fig_height]{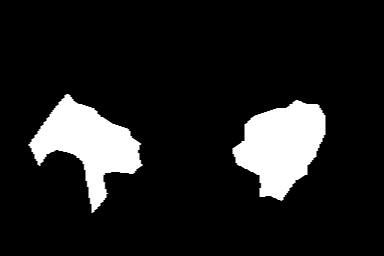}
			\end{overpic}&
			\begin{overpic}[height=\fig_height]{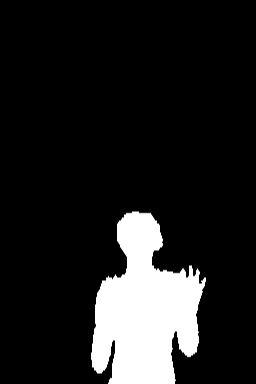}
			\end{overpic}&
			\begin{overpic}[height=\fig_height]{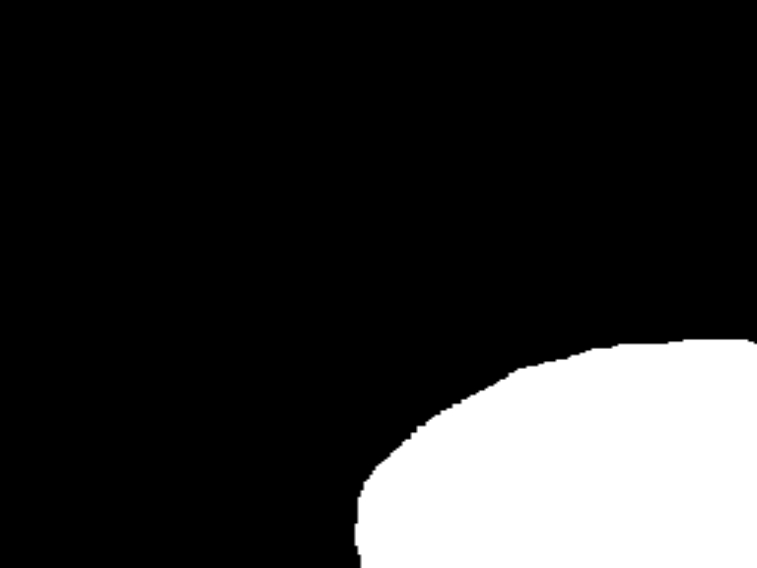}
			\end{overpic}&
			\vspace{0.05cm}
			\\
			\begin{overpic}[height=\fig_height]{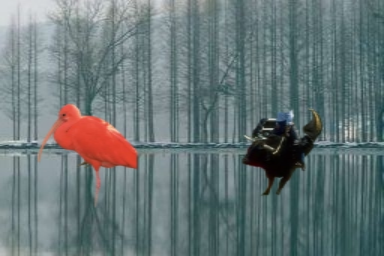}
			\end{overpic}&
			\begin{overpic}[height=\fig_height]{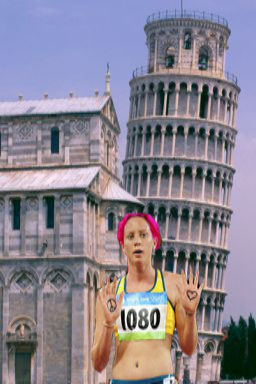}
			\end{overpic}&
			\begin{overpic}[height=\fig_height]{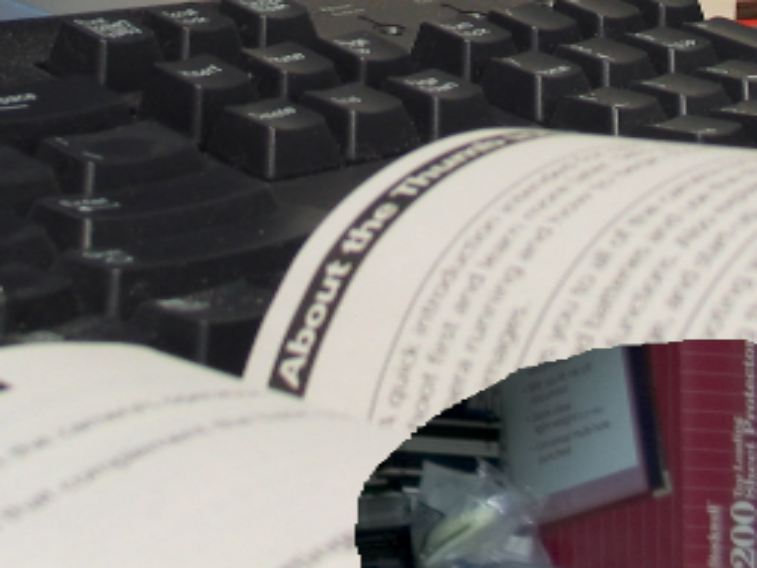}
			\end{overpic}&
			\\\\
			\cdashline{1-3}\\
			\begin{overpic}[height=\fig_height]{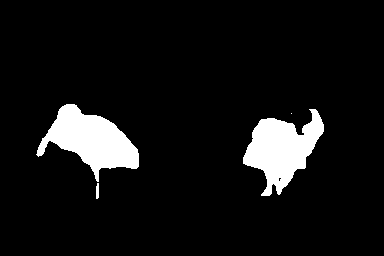}
			\end{overpic}&
			\begin{overpic}[height=\fig_height]{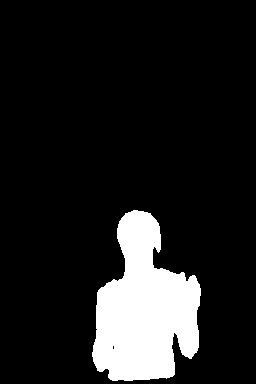}
			\end{overpic}&
			\begin{overpic}[height=\fig_height]{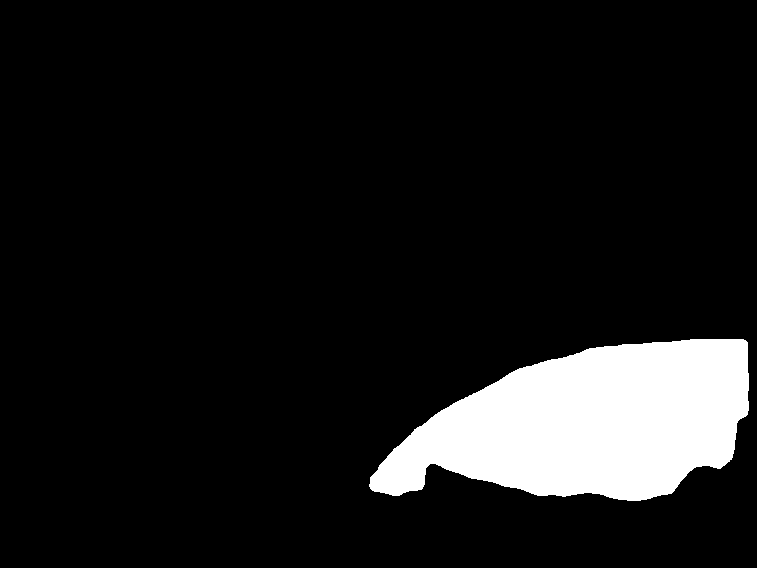}
			\end{overpic}&
			\vspace{0.05cm}
			\\
			\begin{overpic}[height=\fig_height]{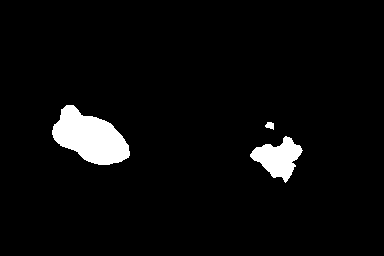}
			\end{overpic}&
			\begin{overpic}[height=\fig_height]{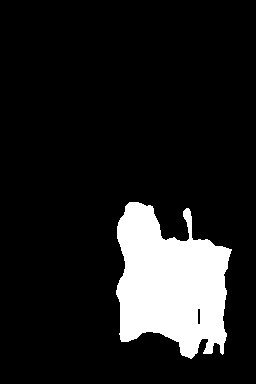}
			\end{overpic}&
			\begin{overpic}[height=\fig_height]{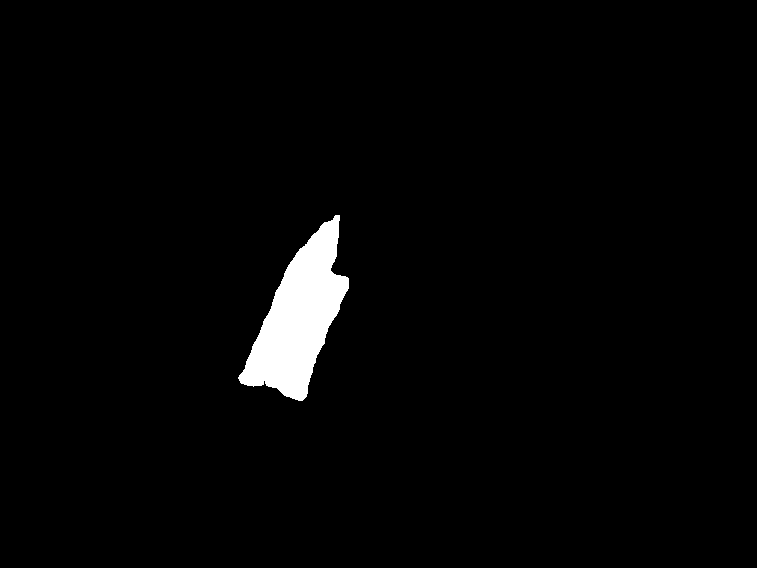}
			\end{overpic}&
			\vspace{0.05cm}
			\\
			\begin{overpic}[height=\fig_height]{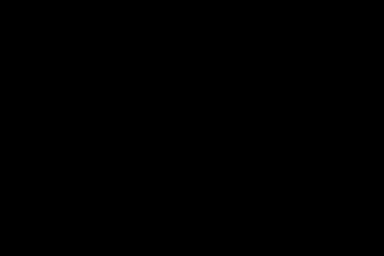}
			\end{overpic}&
			\begin{overpic}[height=\fig_height]{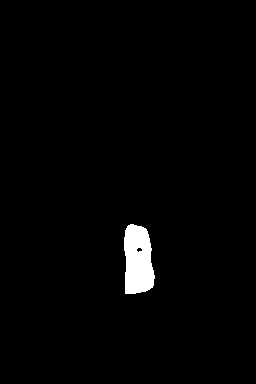}
			\end{overpic}&
			\begin{overpic}[height=\fig_height]{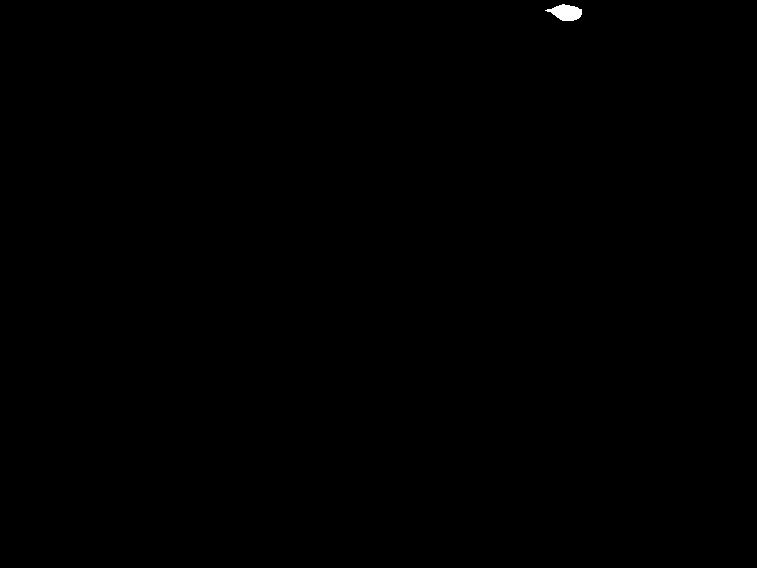}
			\end{overpic}&
			\vspace{0.05cm}
			\\
			\begin{overpic}[height=\fig_height]{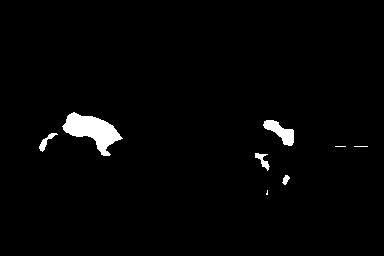}
			\end{overpic}&
			\begin{overpic}[height=\fig_height]{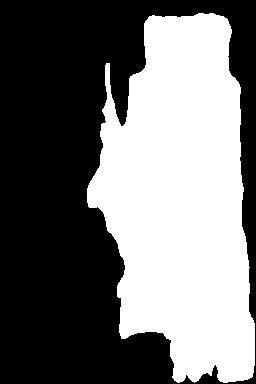}
			\end{overpic}&
			\begin{overpic}[height=\fig_height]{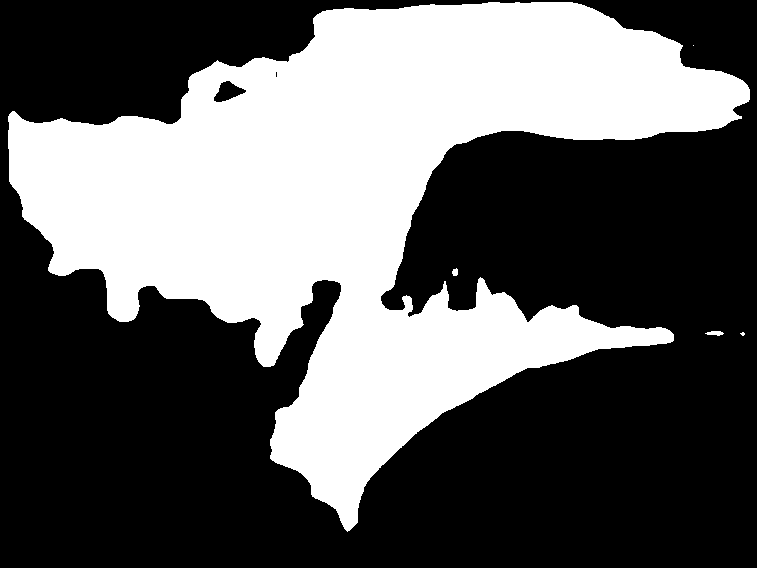}
			\end{overpic}&
			\vspace{0.05cm}
			\\
			\begin{overpic}[height=\fig_height]{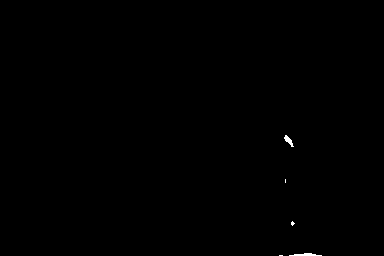}
			\end{overpic}&
			\begin{overpic}[height=\fig_height]{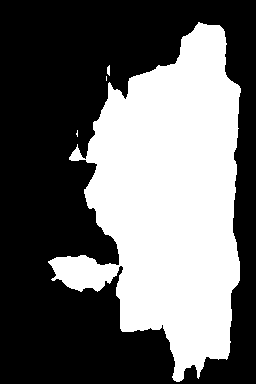}
			\end{overpic}&
			\begin{overpic}[height=\fig_height]{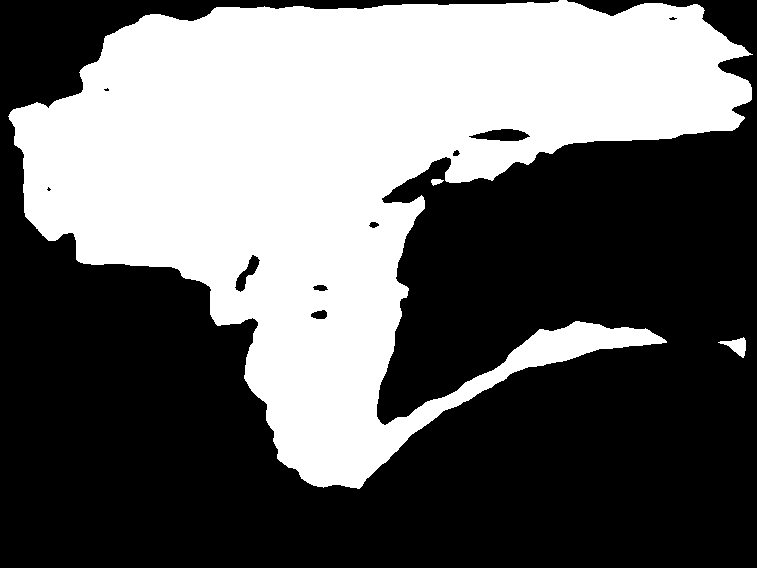}
			\end{overpic}&
			\vspace{0.05cm}
			\\
			\begin{overpic}[height=\fig_height]{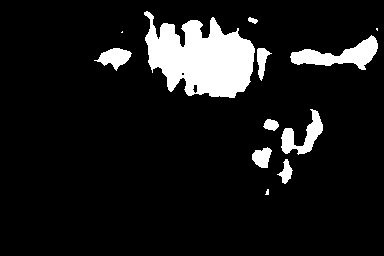}
			\end{overpic}&
			\begin{overpic}[height=\fig_height]{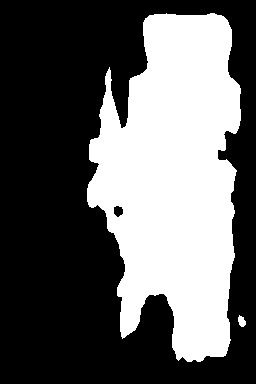}
			\end{overpic}&
			\begin{overpic}[height=\fig_height]{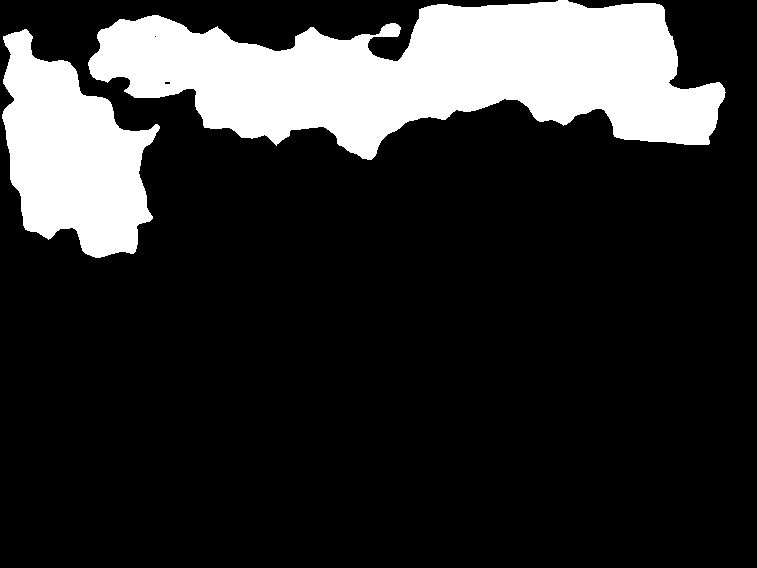}
			\end{overpic}&
			\vspace{0.05cm}
			\\
			\begin{overpic}[height=\fig_height]{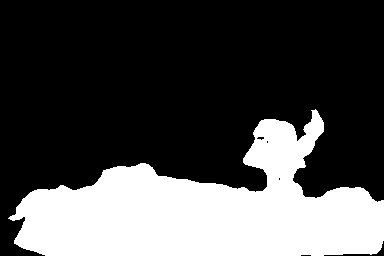}
			\end{overpic}&
			\begin{overpic}[height=\fig_height]{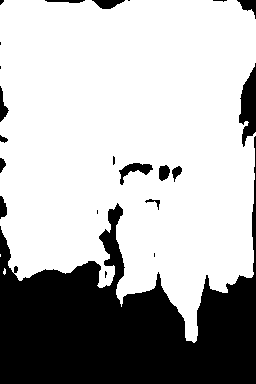}
			\end{overpic}&
			\begin{overpic}[height=\fig_height]{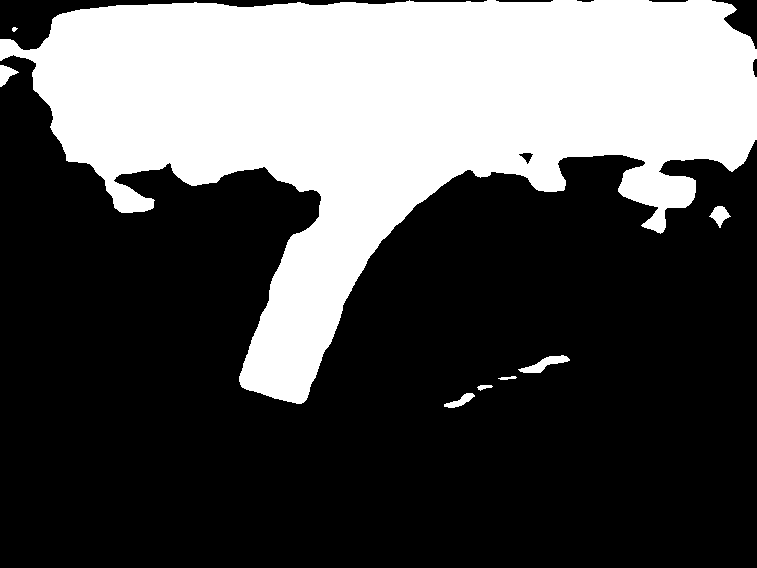}
			\end{overpic}&
			\\\\
			\cdashline{1-3}\\
			\begin{overpic}[height=\fig_height]{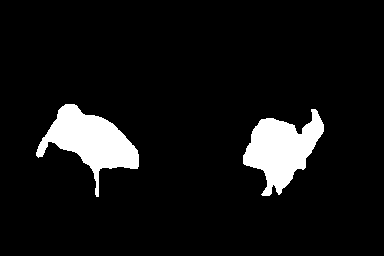}
			\end{overpic}&
			\begin{overpic}[height=\fig_height]{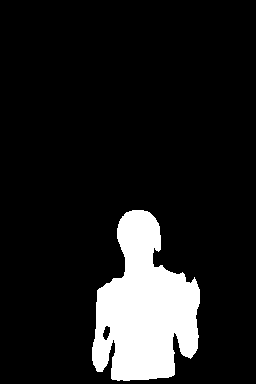}
			\end{overpic}&
			\begin{overpic}[height=\fig_height]{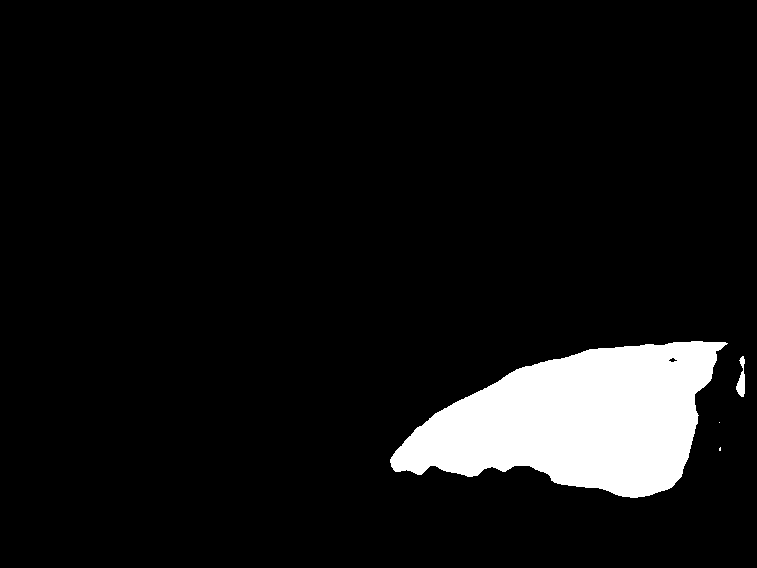}
			\end{overpic}&
			\vspace{0.05cm}
			\\
			\begin{overpic}[height=\fig_height]{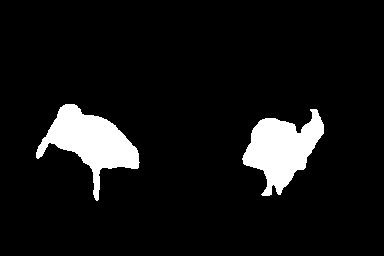}
			\end{overpic}&
			\begin{overpic}[height=\fig_height]{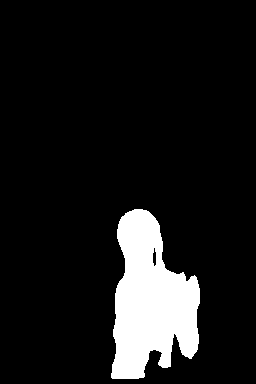}
			\end{overpic}&
			\begin{overpic}[height=\fig_height]{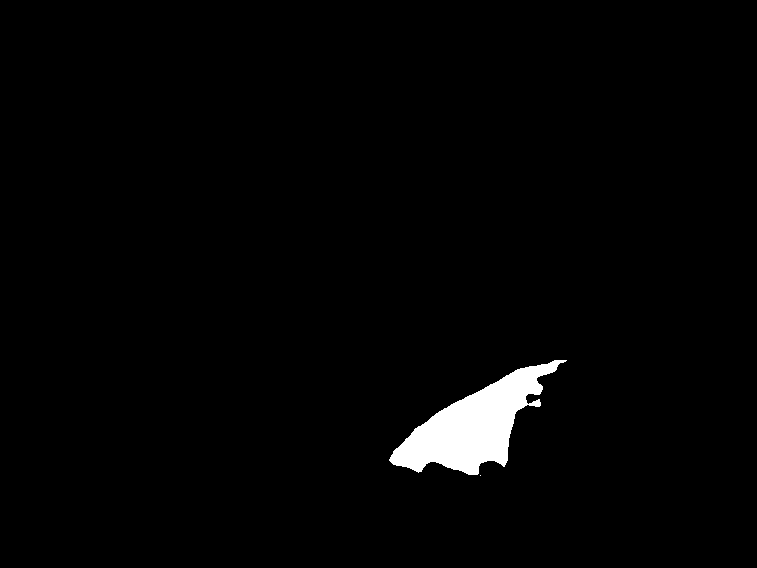}
			\end{overpic}&
			\vspace{0.05cm}
			\\
			\begin{overpic}[height=\fig_height]{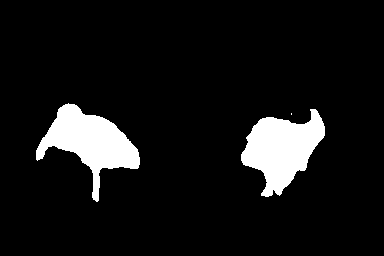}
			\end{overpic}&
			\begin{overpic}[height=\fig_height]{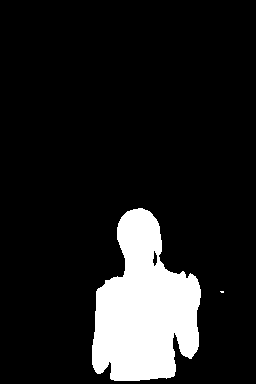}
			\end{overpic}&
			\begin{overpic}[height=\fig_height]{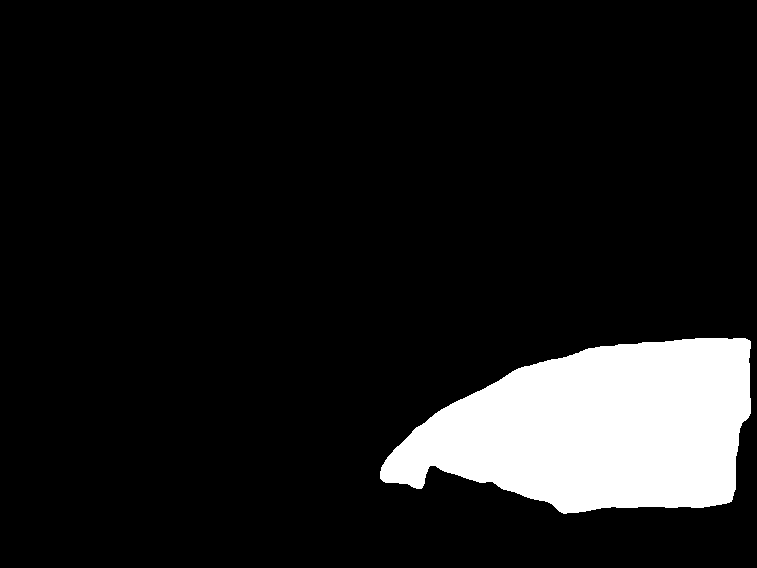}
			\end{overpic}&
			\vspace{0.05cm}
			\\
			\begin{overpic}[height=\fig_height]{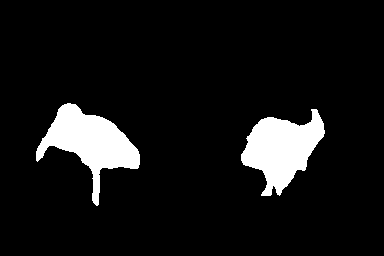}
			\end{overpic}&
			\begin{overpic}[height=\fig_height]{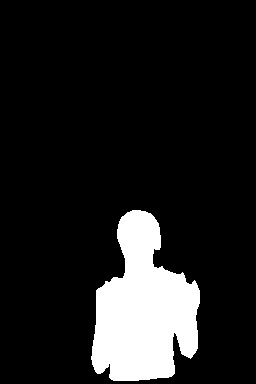}
			\end{overpic}&
			\begin{overpic}[height=\fig_height]{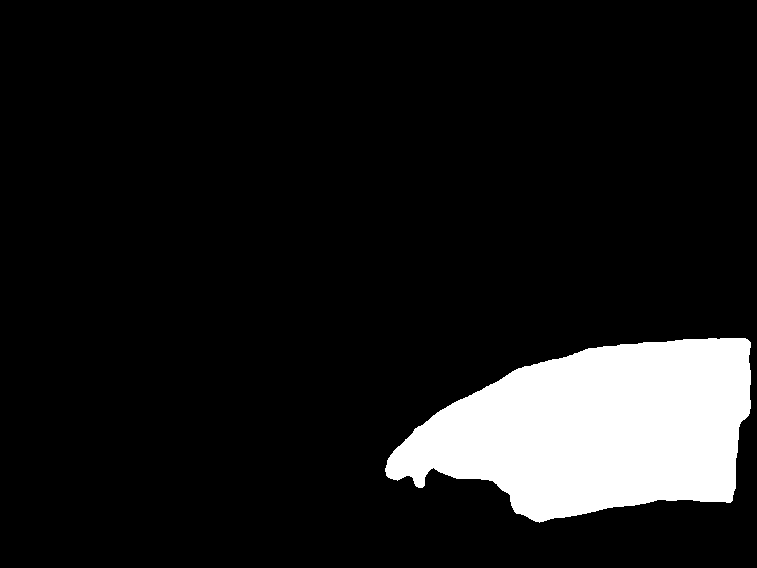}
			\end{overpic}&
			\vspace{0.05cm}
			\\
			\begin{overpic}[height=\fig_height]{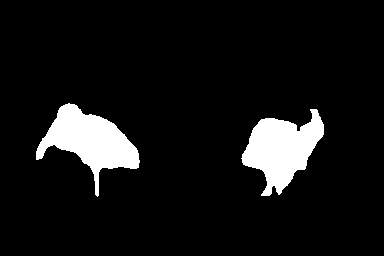}
			\end{overpic}&
			\begin{overpic}[height=\fig_height]{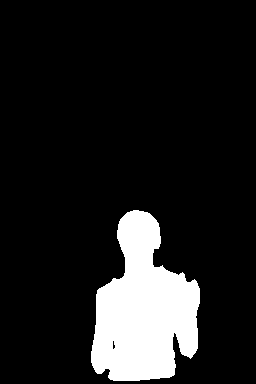}
			\end{overpic}&
			\begin{overpic}[height=\fig_height]{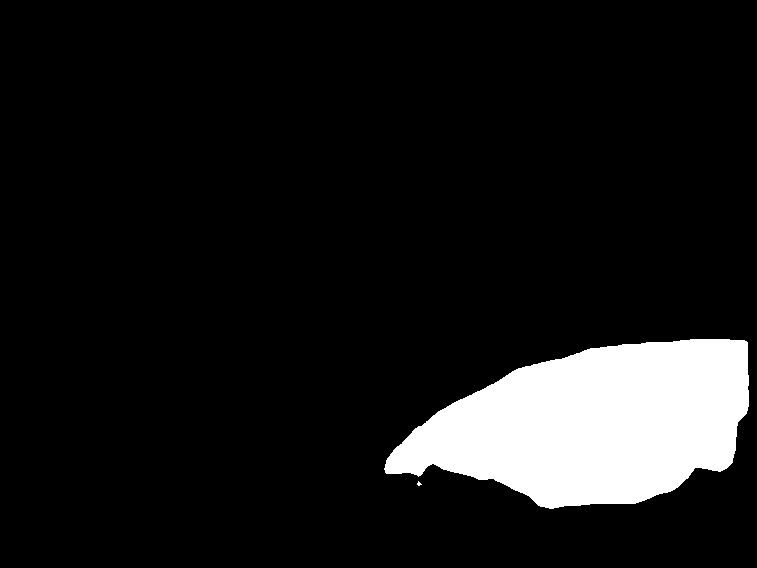}
			\end{overpic}&
			\vspace{0.05cm}
			\\
			\begin{overpic}[height=\fig_height]{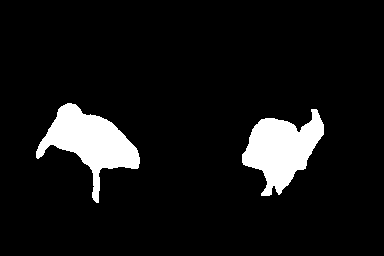}
			\end{overpic}&
			\begin{overpic}[height=\fig_height]{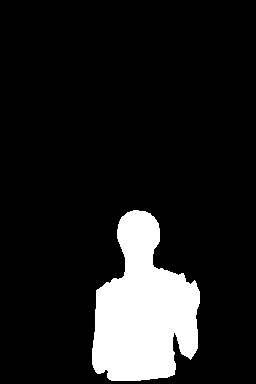}
			\end{overpic}&
			\begin{overpic}[height=\fig_height]{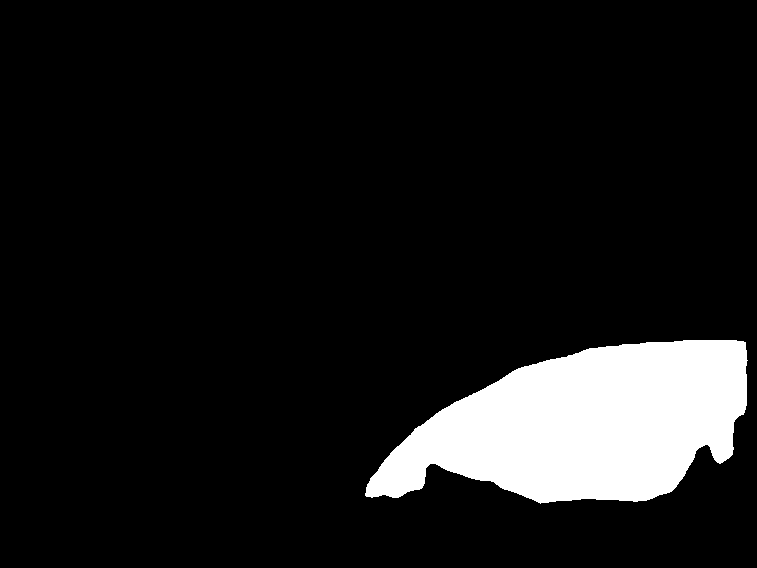}
			\end{overpic}&
			\vspace{0.05cm}
			\\
			\begin{overpic}[height=\fig_height]{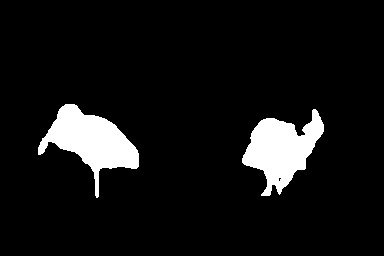}
			\end{overpic}&
			\begin{overpic}[height=\fig_height]{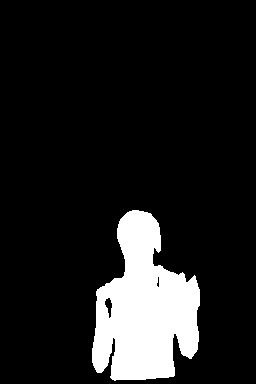}
			\end{overpic}&
			\begin{overpic}[height=\fig_height]{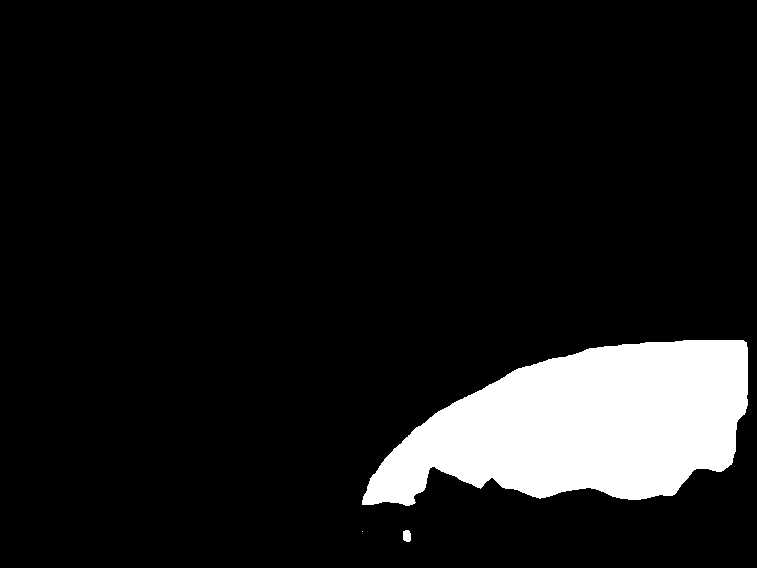}
			\end{overpic}&
		\end{tabular}
	}
	\hspace{\new_hspace}
	\subfloat[CoDE]{

		\def\model{CoDE}
		\def\figa{Sp_D_ani_0001_cha_00063_sec_00081_202.png}
		\def\figb{Sp_D_NND_A_sec0067_ani0096_0620.png}
		\def\figc{00044_fake.png}
		\begin{tabular}{c c c c c c c c c c}
			\begin{overpic}[height=\fig_height]{./pic/attack_effect/\model/misd/GT/\figa}
			\end{overpic}&
			\begin{overpic}[height=\fig_height]{./pic/attack_effect/\model/casia1/GT/\figb}
			\end{overpic}&
			\begin{overpic}[height=\fig_height]{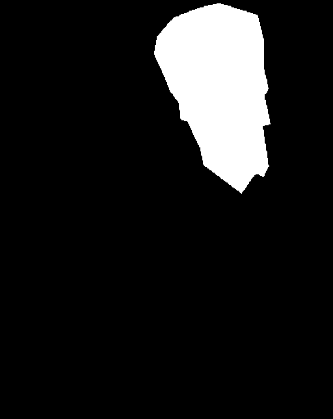}
			\end{overpic}&
			\vspace{0.05cm}
			\\
			\begin{overpic}[height=\fig_height]{./pic/attack_effect/\model/misd/image/\figa}
			\end{overpic}&
			\begin{overpic}[height=\fig_height]{./pic/attack_effect/\model/casia1/image/\figb}
			\end{overpic}&
			\begin{overpic}[height=\fig_height]{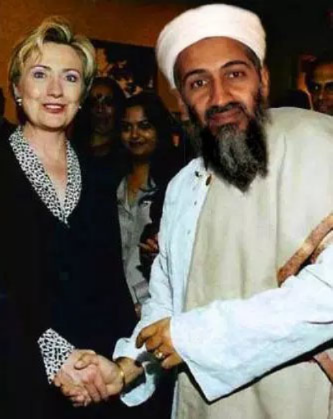}
			\end{overpic}&
			\\\\
			\cdashline{1-3}\\
			\begin{overpic}[height=\fig_height]{./pic/attack_effect/\model/misd/Clean/False/\figa}
			\end{overpic}&
			\begin{overpic}[height=\fig_height]{./pic/attack_effect/\model/casia1/Clean/False/\figb}
			\end{overpic}&
			\begin{overpic}[height=\fig_height]{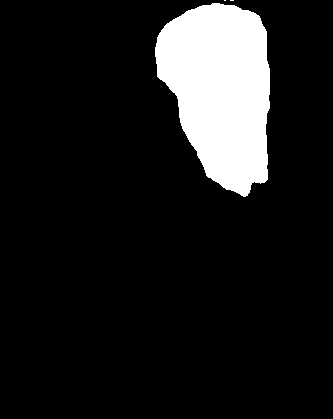}
			\end{overpic}&
			\vspace{0.05cm}
			\\
			\begin{overpic}[height=\fig_height]{./pic/attack_effect/\model/misd/FGSM/False/\figa}
			\end{overpic}&
			\begin{overpic}[height=\fig_height]{./pic/attack_effect/\model/casia1/FGSM/False/\figb}
			\end{overpic}&
			\begin{overpic}[height=\fig_height]{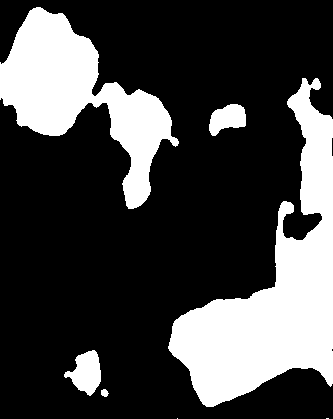}
			\end{overpic}&
			\vspace{0.05cm}
			\\
			\begin{overpic}[height=\fig_height]{./pic/attack_effect/\model/misd/CW/False/\figa}
			\end{overpic}&
			\begin{overpic}[height=\fig_height]{./pic/attack_effect/\model/casia1/CW/False/\figb}
			\end{overpic}&
			\begin{overpic}[height=\fig_height]{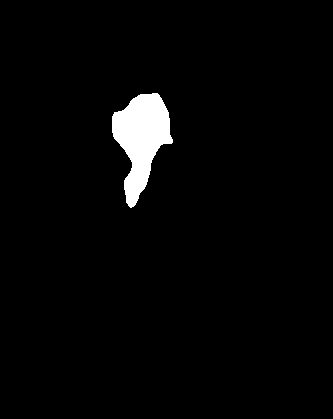}
			\end{overpic}&
			\vspace{0.05cm}
			\\
			\begin{overpic}[height=\fig_height]{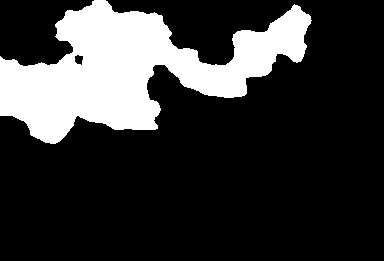}
			\end{overpic}&
			\begin{overpic}[height=\fig_height]{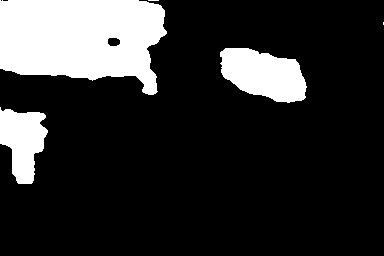}
			\end{overpic}&
			\begin{overpic}[height=\fig_height]{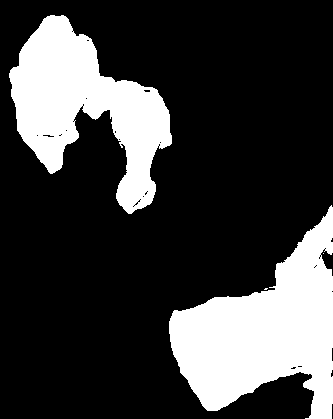}
			\end{overpic}&
			\vspace{0.05cm}
			\\
			\begin{overpic}[height=\fig_height]{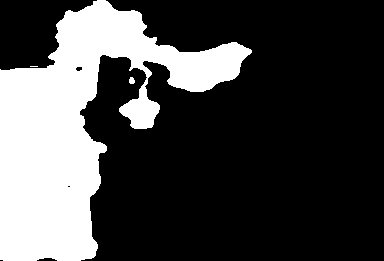}
			\end{overpic}&
			\begin{overpic}[height=\fig_height]{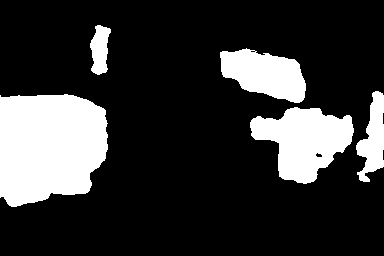}
			\end{overpic}&
			\begin{overpic}[height=\fig_height]{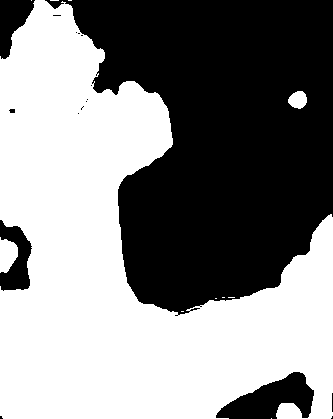}
			\end{overpic}&
			\vspace{0.05cm}
			\\
			\begin{overpic}[height=\fig_height]{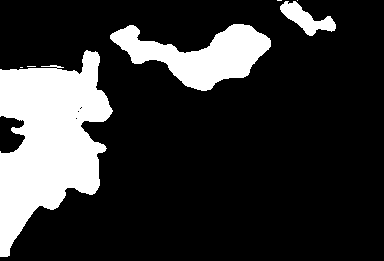}
			\end{overpic}&
			\begin{overpic}[height=\fig_height]{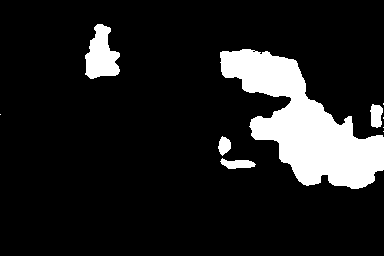}
			\end{overpic}&
			\begin{overpic}[height=\fig_height]{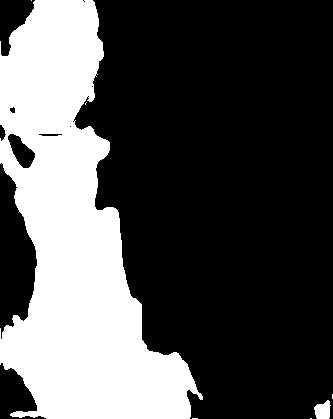}
			\end{overpic}&
			\vspace{0.05cm}
			\\
			\begin{overpic}[height=\fig_height]{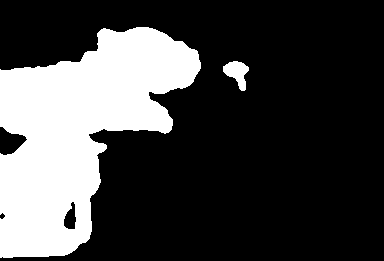}
			\end{overpic}&
			\begin{overpic}[height=\fig_height]{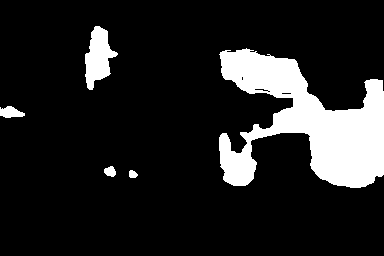}
			\end{overpic}&
			\begin{overpic}[height=\fig_height]{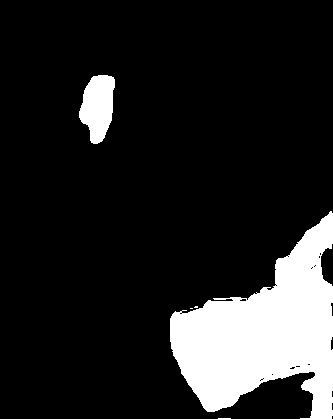}
			\end{overpic}&
			\\\\
			\cdashline{1-3}\\
			\begin{overpic}[height=\fig_height]{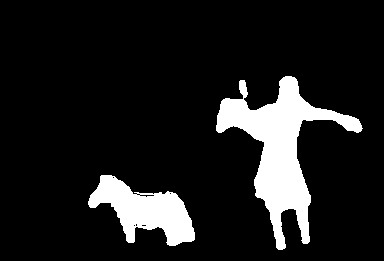}
			\end{overpic}&
			\begin{overpic}[height=\fig_height]{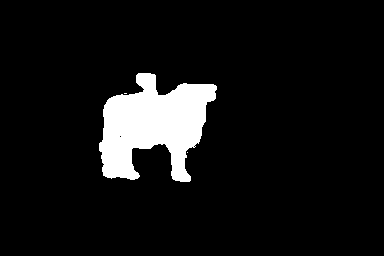}
			\end{overpic}&
			\begin{overpic}[height=\fig_height]{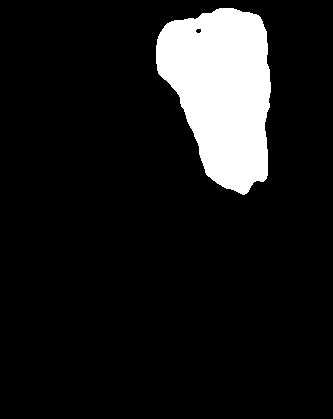}
			\end{overpic}&
			\vspace{0.05cm}
			\\
			\begin{overpic}[height=\fig_height]{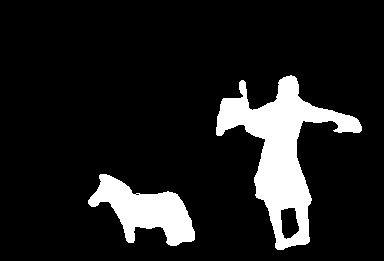}
			\end{overpic}&
			\begin{overpic}[height=\fig_height]{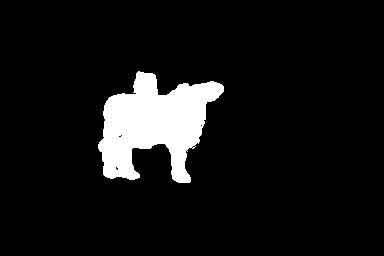}
			\end{overpic}&
			\begin{overpic}[height=\fig_height]{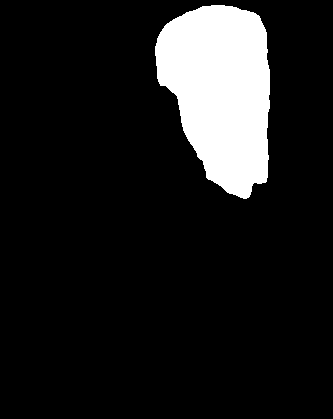}
			\end{overpic}&
			\vspace{0.05cm}
			\\
			\begin{overpic}[height=\fig_height]{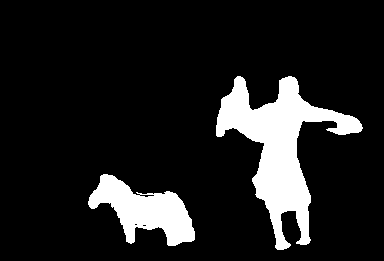}
			\end{overpic}&
			\begin{overpic}[height=\fig_height]{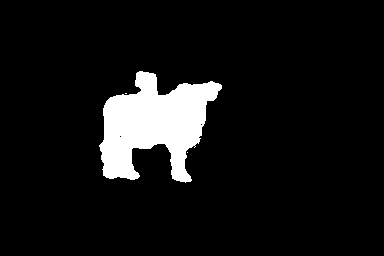}
			\end{overpic}&
			\begin{overpic}[height=\fig_height]{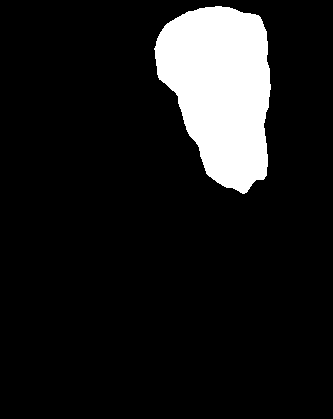}
			\end{overpic}&
			\vspace{0.05cm}
			\\
			\begin{overpic}[height=\fig_height]{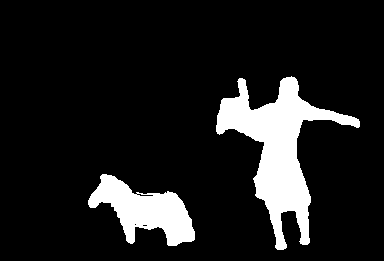}
			\end{overpic}&
			\begin{overpic}[height=\fig_height]{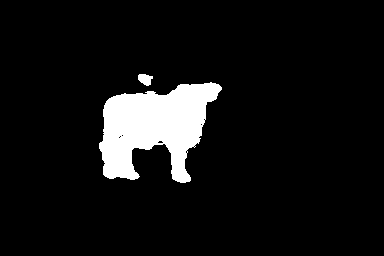}
			\end{overpic}&
			\begin{overpic}[height=\fig_height]{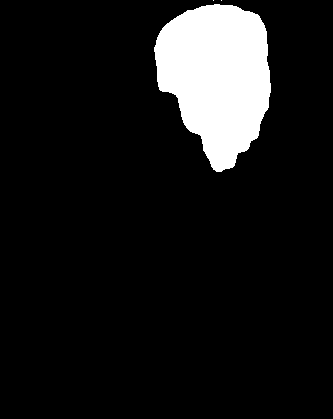}
			\end{overpic}&
			\vspace{0.05cm}
			\\
			\begin{overpic}[height=\fig_height]{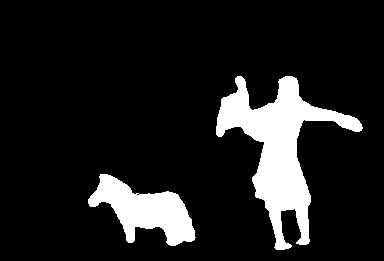}
			\end{overpic}&
			\begin{overpic}[height=\fig_height]{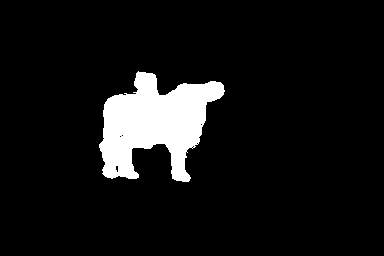}
			\end{overpic}&
			\begin{overpic}[height=\fig_height]{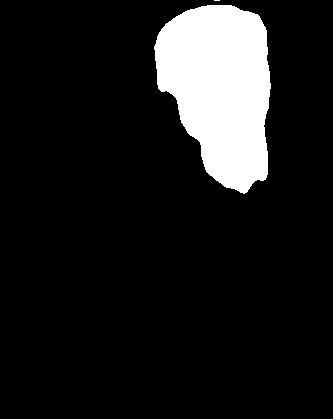}
			\end{overpic}&
			\vspace{0.05cm}
			\\
			\begin{overpic}[height=\fig_height]{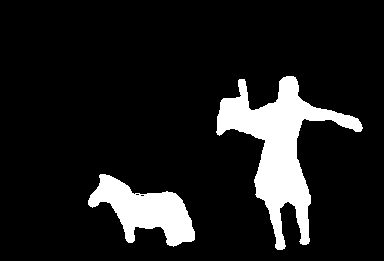}
			\end{overpic}&
			\begin{overpic}[height=\fig_height]{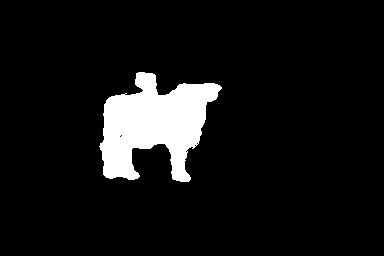}
			\end{overpic}&
			\begin{overpic}[height=\fig_height]{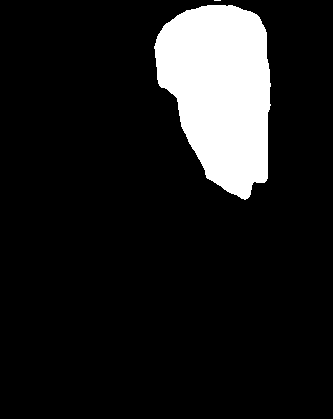}
			\end{overpic}&
			\vspace{0.05cm}
			\\
			\begin{overpic}[height=\fig_height]{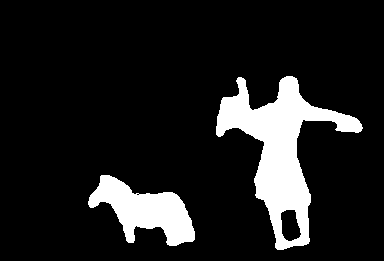}
			\end{overpic}&
			\begin{overpic}[height=\fig_height]{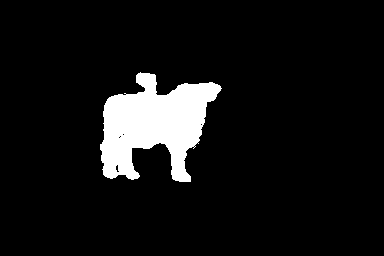}
			\end{overpic}&
			\begin{overpic}[height=\fig_height]{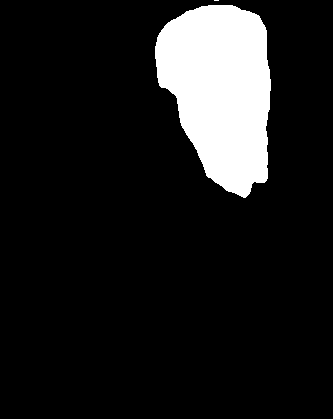}
			\end{overpic}&
		\end{tabular}
	}
	\caption{Visual comparisons under various adversarial attacks. The forged area is denoted by white color and pristine area is denoted by black color. The upper right corner of $\bm{M_a}$ and $\bm{M_{pa}}$ indicates the adversarial attack method.}
	\label{fig:visual_comparison} 
\end{figure*}
Table~\ref{table:defense_effect} summarizes the quantitative evaluation results of the proposed ANSM's defense effect against adversarial attacks. We report the average performance across multiple noise intensities for each adversarial attack algorithm. 

Attribute to the two-stage training strategy: FFA and MgR, the optimized ANSM substantially restores the forgery localization performance, achieving an overall RP $\geq$ 90\%.
Despite being trained exclusively on the CASIAv2 dataset using the FGSM algorithm, the ANSM demonstrates outstanding generalization capabilities across different adversarial algorithms. Remarkably, under six adversarial attack algorithms FGSM, C\&W, BIM, PGD, MI-FGSM and PGN, CoDE 's average RP recover from 24.3\%, 8.9\%, 3.3\%, 3.5\%, 2.6\%, 4.2\% to 99.7\%, 94.9\%, 95.9\%, 94.7\%, 96.9\%, 95.0\%, respectively. Importantly, when original forged images are pre-processed by the ANSM, IF-OSN, MVSS-Net, HDF-Net, and CoDE achieve average RPs of 94.3\%, 99.3\%, 98.6\%, and 94.8\%, respectively. This clearly demonstrates that ANSM has strong adaptability to original forged images and preserves the original forgery localization performance to a large extent.

\subsection{Visual Analysis}
Fig.~\ref{fig:visual_comparison} illustrates the forgery localization results obtained with and without the ANSM. The visual comparison clearly indicates that adversarial attacks substantially disrupt forgery localization, resulting in inaccurate identification of forgery region. When the ANSM is applied, the forgery localization performance is significantly restored, with $\bm{M_{pa}}$ closely matching $\bm{M_{o}}$. Furthermore, ANSM has strong adaptability. Even if original forged image is pre-processed by ANSM, the predicted forgery mask $\bm{M_{po}}$ maintains highly consistent with $\bm{M_{o}}$.
\begin{figure*}[t]
	\centering
	\def\fig_height{2.2cm}
	\setlength\tabcolsep{0.0mm}
	\subfloat[FGSM]{
		\label{fig:distribution_feature_fgsm}
		\hspace{-0.1cm}
		\begin{tabular}{c}
			\begin{overpic}[height=\fig_height]{./pic/feature_visualization/fgsm_feature.png}
				\put(295, 95){Before ANSM}
				\put(303, -24){After ANSM}
			\end{overpic}
			\vspace{0.5cm}
			\\
			\begin{overpic}[height=\fig_height]{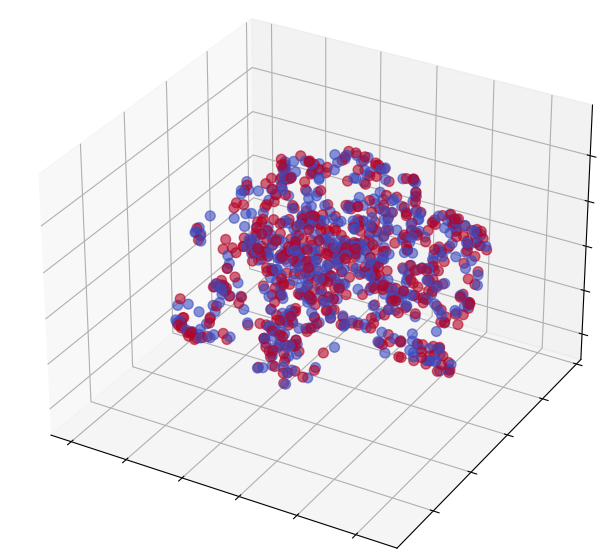}
			\end{overpic}
		\end{tabular}
	}	
	\subfloat[C$\&$W]{
		\label{fig:distribution_feature_cw}
		\hspace{-0.1cm}
		\begin{tabular}{c}
			\begin{overpic}[height=\fig_height]{./pic/feature_visualization/cw_feature.png}
			\end{overpic}
			\vspace{0.5cm}
			\\
			\begin{overpic}[height=\fig_height]{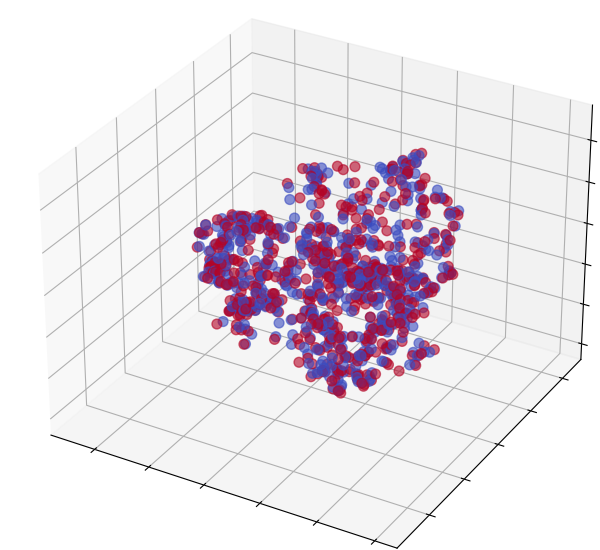}
			\end{overpic}
		\end{tabular}
	}
	\subfloat[BIM]{
		\label{fig:distribution_feature_bim}
		\hspace{-0.1cm}
		\begin{tabular}{c}
			\begin{overpic}[height=\fig_height]{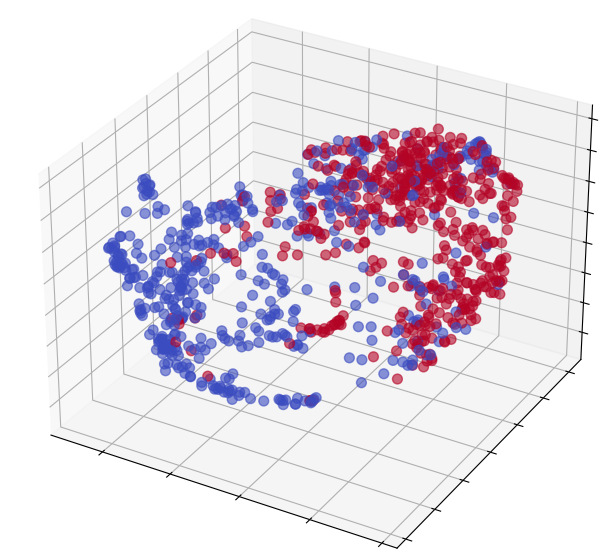}
			\end{overpic}
			\vspace{0.5cm}
			\\
			\begin{overpic}[height=\fig_height]{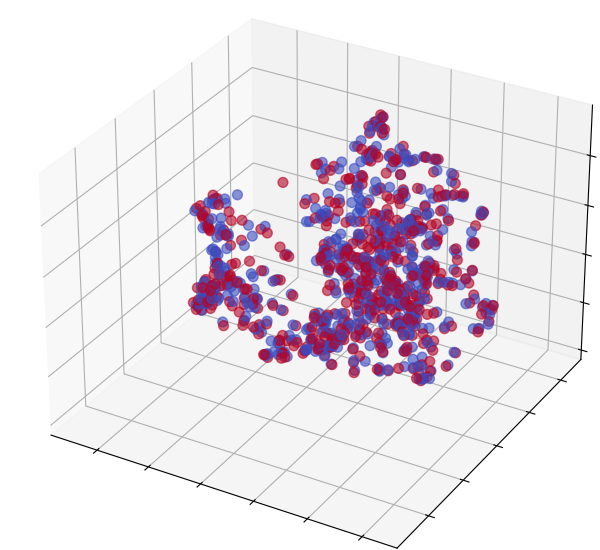}
			\end{overpic}
		\end{tabular}
	}
	\subfloat[PGD]{
		\label{fig:distribution_feature_pgd}
		\hspace{-0.1cm}
		\begin{tabular}{c}
			\begin{overpic}[height=\fig_height]{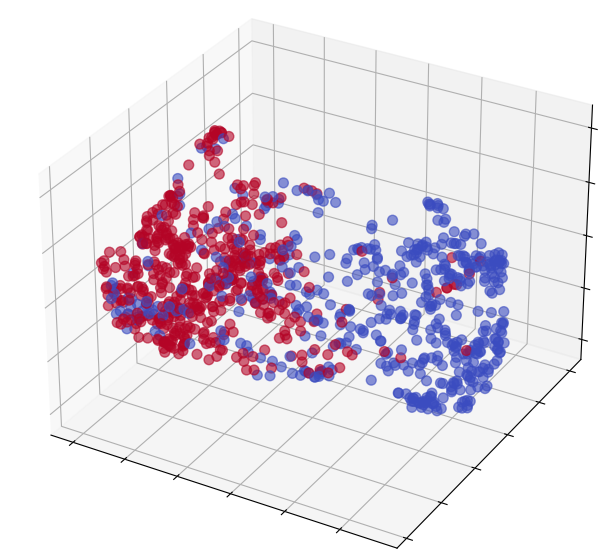}
			\end{overpic}
			\vspace{0.5cm}
			\\
			\begin{overpic}[height=\fig_height]{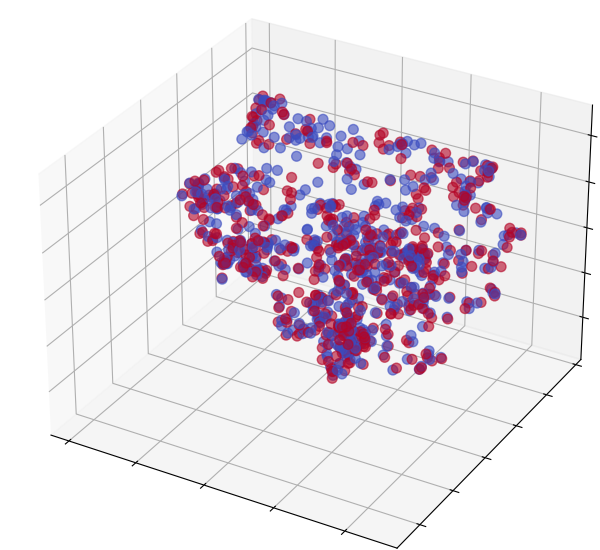}
			\end{overpic}
		\end{tabular}
	}
	\subfloat[MI-FGSM]{
		\label{fig:distribution_feature_mifgsm}
		\hspace{-0.1cm}
		\begin{tabular}{c}
			\begin{overpic}[height=\fig_height]{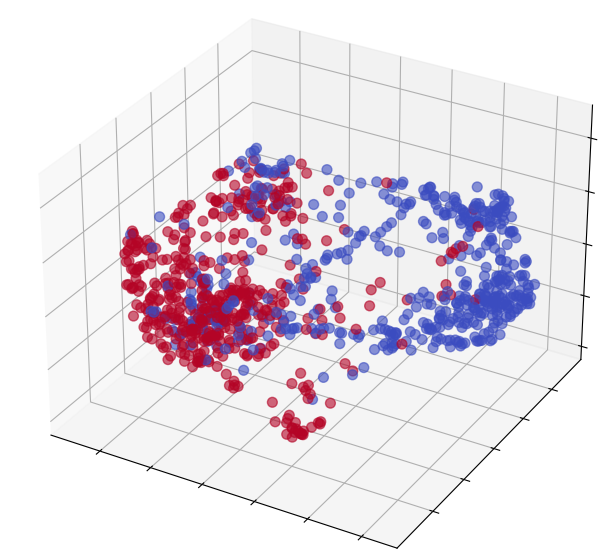}
			\end{overpic}
			\vspace{0.5cm}
			\\
			\begin{overpic}[height=\fig_height]{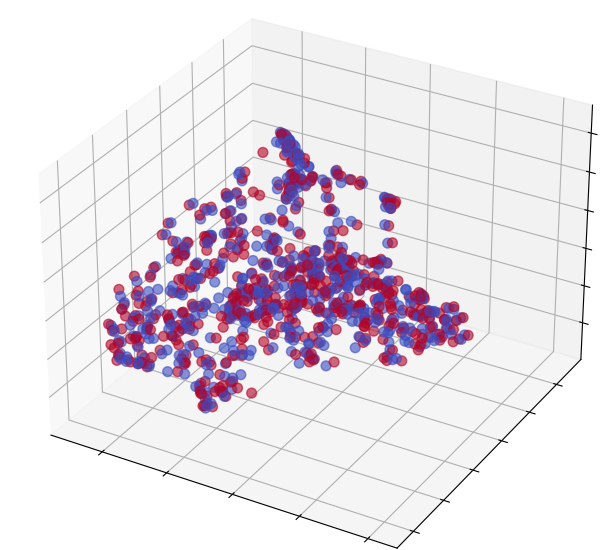}
			\end{overpic}
		\end{tabular}
	}
	\subfloat[PGN]{
		\label{fig:distribution_feature_pgn}
		\hspace{-0.1cm}
		\begin{tabular}{c}
			\begin{overpic}[height=\fig_height]{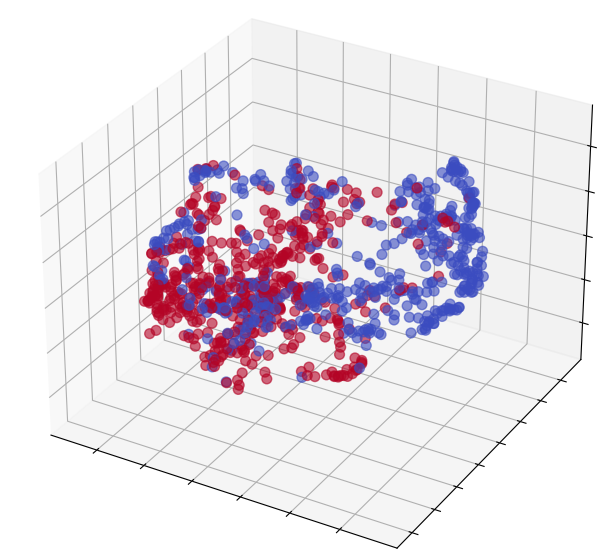}
			\end{overpic}
			\vspace{0.5cm}
			\\
			\begin{overpic}[height=\fig_height]{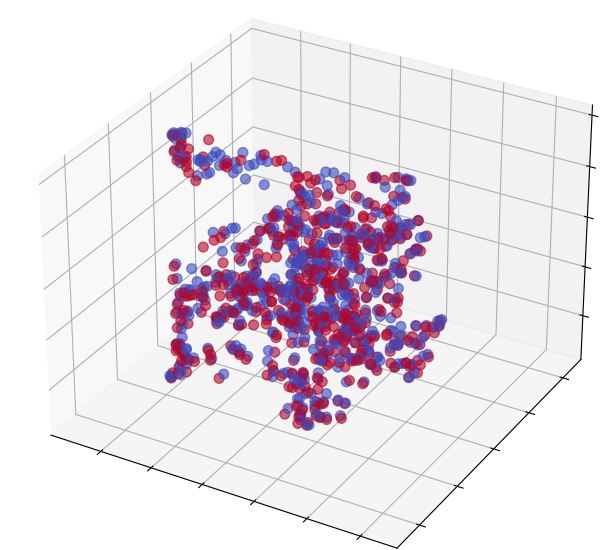}
			\end{overpic}
		\end{tabular}
	}
	\caption{Visualization of forgery-relevant features using the dimensionality reduction algorithm UMAP. The first row presents the results between $\bm{X_{o}}$ (red) and $\bm{X_{a}}$ (blue), while the second row shows the results between $\bm{X_{o}}$ (red) and $\bm{X_{pa}}$ (blue).}
	\label{fig:distribution_alignment_analysis}
\end{figure*}

To further examine the impact of feature alignment, we employed the UMAP~\cite{mcinnes2018umap} algorithm to project forgery-relevant features into a three-dimensional latent space, as depicted in Fig.~\ref{fig:distribution_alignment_analysis}. Initially, adversarial attacks lead to notable deviations in the feature distributions, distinctly separating adversarial forged images from original forged images into two separate clusters. After applying the ANSM, the defensive perturbed images demonstrate substantial realignment with the original forged images in the latent space, effectively merging into overlapping cluster structures. This alignment clearly highlights the ANSM's capability to restore intrinsic feature distributions disrupted by adversarial noise.

\subsection{Ablation Study}
In this section, we show the effect of different factors in FFA and MgR. All results reported in this section are obtained with CASIAv1, IMD20, MISD as the test set.

\begin{table}[t]\footnotesize
	\centering
	\renewcommand{\arraystretch}{1.1}
	\tabcolsep=0.12cm
	\caption{The effect of different depths of features during FFA.}
	\label{table:aligned_feature_layers}
	\begin{tabular}{c | c | c c | c c}	
		\Xhline{2pt}
		\multirow{2}*{Model} & \multirow{2}*{\makecell[c]{Depth}} & \multicolumn{2}{c|}{Original} & \multicolumn{2}{c}{FGSM} \\
		& & F1 & RP(\%) & F1 & RP(\%)
		\\
		\hline
		\multirow{3}*{IF-OSN} 
		& Shallow 
		& \textbf{0.461} & \textbf{90.9} 
		& 0.251 & 49.4 
		\\
		& Topmost
		& 0.433 & 85.4 
		& 0.430 & 84.7 
		\\
		& Middle
		& \underline{0.444} & \underline{87.5} 
		& \textbf{0.454} & \textbf{89.4} 
		\\
		\hline
		\multirow{3}*{CoDE} 
		& Shallow
		& \textbf{0.717} & \textbf{97.5} 
		& 0.307 & 41.7 
		\\
		& Topmost
		& 0.621 & 84.4 
		& \textbf{0.685} & \textbf{93.0} 
		\\
		& Middle
		& \underline{0.689} & \underline{93.6} 
		& \underline{0.670} & \underline{91.1} 
		\\
		\Xhline{2pt}
	\end{tabular}
\end{table}
\subsubsection{Aligned Feature Layers}
\label{section:aligned_feature_layers}
Table~\ref{table:aligned_feature_layers} summarizes the effect of aligning different depths of feature layers. The results clearly demonstrate that aligning shallow-layer features provides the weakest defense against FGSM attack, with RP of only 49.4\% and 41.7\% for IF-OSN and CoDE, respectively. In contrast, aligning the topmost-layer features significantly enhances defense effect, achieving RP of 84.7\% and 93.0\%, respectively. While aligning middle-layer features also yields strong defense effect, with RP of 89.4\% and 91.1\%, respectively. Although both middle-layer and topmost-layer features alignment offer comparable defense effect, aligning middle-layer features has a crucial advantage—it achieves less performance degradation on original forged images. Specifically, under topmost-layer features alignment, IF-OSN and CoDE achieve RP of 85.4\% and 84.4\% on original forged images. While under middle-layer features alignment, their RP improve to 87.5\% and 93.6\%, respectively. This balance between defending against adversarial forgeries and preserving performance on original forged images is particularly important, as it demonstrates that ANSM has strong adaptability to handle scenarios both with and without adversarial noise.

\begin{table}[t!]\footnotesize
	\centering
	\renewcommand{\arraystretch}{1.1}
	\tabcolsep=0.12cm
	\caption{The effect of ANSM optimized with PwR and FFA.}
	\label{table:pr_ffa}
	\begin{tabular}{c | c | c c | c c}	
		\Xhline{2pt}
		\multirow{2}*{Model} &  \multirow{2}*{\makecell[c]{Strategy}} & \multicolumn{2}{c|}{Original} & \multicolumn{2}{c}{FGSM} \\
		& & F1 & RP(\%)& F1 & RP(\%)
		\\
		\hline
		\multirow{2}*{IF-OSN} 
		& PwR
		& 0.441 & 87.0 
		& 0.250 & 49.3 
		\\
		& FFA
		& \textbf{0.444} & \textbf{87.5} 
		& \textbf{0.454} & \textbf{89.4} 
		\\
		\hline
		\multirow{2}*{CoDE} 
		& PwR
		& \textbf{0.700} & \textbf{95.1} 
		& 0.372 & 50.6 
		\\
		& FFA
		& 0.689 & 93.6 
		& \textbf{0.670} & \textbf{91.1} 
		\\
		\Xhline{2pt}
	\end{tabular}
\end{table}
\subsubsection{Pixel-wise Reconstruction versus Forgery-relevant Features Alignment}
\label{section:Pixel Reconstruction vs Forgery-relevant Feature Alignment}
Table~\ref{table:pr_ffa} provides a comparative analysis of the defensive effectiveness of Pixel-wise Reconstruction (PwR) and FFA in optimizing the ANSM. For PwR, we aim to minimize the L1 loss between the defensive perturbed image $\bm{X_{pa}}$ and the corresponding original forged image $\bm{X_{o}}$. 

While both PwR and FFA achieve comparable performance on original forged images, FFA significantly outperforms PwR under adversarial attacks. Specifically, under the FGSM attack, FFA achieves RP scores of 89.4\% and 91.1\% for IF-OSN and CoDE, markedly exceeding the 49.3\% and 50.6\% obtained by PwR.
This comparison underscores a crucial insight: PwR emphasizes superficial differences, neglecting the deeper, structural disruptions induced by adversarial noise. In contrast, FFA directly addresses these underlying issues by aligning forgery-relevant features, thereby offering superior resistance against adversarial noise.

\begin{table*}[t!]\footnotesize
	\centering
	\renewcommand{\arraystretch}{1.1}
	\tabcolsep=0.12cm
	\caption{The effect of ANSM optimized with different combination of FFA and MgR.}
	\label{table:Two-stage progressive optimization}
	\begin{tabular}{c | c c c | c c | c c | c c | c c | c c | c c | c c}
		\hline\hline
		\multirow{3}*{Model} & \multirow{3}*{FFA} & \multicolumn{2}{c|}{MgR} & \multicolumn{2}{c|}{\multirow{2}*{Original}} & \multicolumn{2}{c|}{\multirow{2}*{FGSM}} & \multicolumn{2}{c|}{\multirow{2}*{C\&W}} & \multicolumn{2}{c|}{\multirow{2}*{BIM}} & \multicolumn{2}{c|}{\multirow{2}*{PGD}} & \multicolumn{2}{c|}{\multirow{2}*{MI-FGSM}} & \multicolumn{2}{c}{\multirow{2}*{PGN}}
		\\
		\cmidrule(lr){3-4}
		& & $\bm{M_{gt}}$ & $\bm{M_{o}}$ & F1 & RP(\%) & F1 & RP(\%) & F1 & RP(\%) & F1 & RP(\%) & F1 & RP(\%) & F1 & RP(\%) & F1 & RP(\%)
		\\
		\hline
		\multirow{5}*{IF-OSN} 
		& \checkmark  & & 
		& 0.444 & 87.5 
		& 0.454 & 89.4 
		& 0.446 & 87.8 
		& 0.425 & 83.8 
		& 0.430 & 84.7 
		& 0.428 & 84.3 
		& 0.428 & 84.3 
		\\
		& & \checkmark &  
		& 0.442 & 87.1 
		& 0.259 & 51.0 
		& 0.361 & 71.2 
		& 0.294 & 57.9 
		& 0.307 & 60.5 
		& 0.251 & 49.4 
		& 0.244 & 48.2 
		\\
		& & & \checkmark 
		& 0.432 & 85.0 
		& 0.216 & 42.6 
		& 0.327 & 64.4 
		& 0.177 & 34.9 
		& 0.191 & 37.6 
		& 0.144 & 28.5 
		& 0.142 & 27.9 
		\\
		& \checkmark & \checkmark &  
		& 0.423 & 83.3 
		& 0.487 & 96.0 
		& 0.428 & 84.4 
		& 0.437 & 86.2 
		& 0.424 & 83.6 
		& 0.443 & 87.3 
		& 0.418 & 82.3 
		\\
		& \checkmark & & \checkmark  
		& \textbf{0.477} & \textbf{94.0} 
		& \textbf{0.503} & \textbf{99.2} 
		& \textbf{0.484} & \textbf{95.5} 
		& \textbf{0.476} & \textbf{93.8} 
		& \textbf{0.470} & \textbf{92.7} 
		& \textbf{0.473} & \textbf{93.3} 
		& \textbf{0.464} & \textbf{91.5} 
		\\
		\hline
		\multirow{5}*{CoDE} 
		& \checkmark & & 
		& 0.689 & 93.6 
		& 0.670 & 91.1 
		& 0.698 & 94.9 
		& 0.690 & 93.8 
		& 0.692 & 94.1 
		& 0.692 & 94.1 
		& 0.685 & 93.1 
		\\
		& & \checkmark &  
		& 0.553 & 75.2 
		& 0.414 & 56.2 
		& 0.537 & 73.0 
		& 0.479 & 65.1 
		& 0.492 & 66.8 
		& 0.427 & 58.1 
		& 0.462 & 62.8 
		\\
		& & & \checkmark 
		& 0.569 & 77.4 
		& 0.413 & 56.1 
		& 0.556 & 75.6 
		& 0.486 & 66.0 
		& 0.499 & 67.8 
		& 0.424 & 57.6 
		& 0.463 & 63.0 
		\\ 
		& \checkmark & \checkmark &  
		& 0.675 & 91.8 
		& \textbf{0.759} & \textbf{103.2} 
		& 0.685	& 93.2 
		& 0.706 & 95.9 
		& \textbf{0.699} & \textbf{95.0} 
		& \textbf{0.725} & \textbf{98.5} 
		& 0.698 & 94.9 
		\\
		& \checkmark & & \checkmark 
		& \textbf{0.697} & \textbf{94.7} 
		& 0.734 & 99.7 
		& \textbf{0.698} & \textbf{94.9} 
		& \textbf{0.707} & \textbf{96.0} 
		& 0.697 & 94.7 
		& 0.713 & 96.9 
		& \textbf{0.699} & \textbf{95.0} 
		\\		
		\hline\hline
	\end{tabular}
\end{table*}

\subsubsection{Two-stage Optimization Strategy} 
\label{section:Two-stage progressive optimization}
Table~\ref{table:Two-stage progressive optimization} presents the individual and combined effects of FFA and MgR. 

When utilized independently, FFA achieves strong performance in adversarial defense, consistently delivering RP exceeding 80\% for IF-OSN and 90\% for CoDE across various adversarial attack algorithms. In contrast, MgR alone performs poorly against adversarial attacks, regardless of whether $\bm{M_{gt}}$ or $\bm{M_{o}}$ is used as the supervised forgery mask, leading to considerably lower RP. 

However, when FFA and MgR are integrated into a two-stage framework, their complementary strengths lead to substantial improvements, particularly when MgR utilizes $\bm{M_{o}}$ as the supervised forgery mask. Compared to FFA alone, the inclusion of MgR significantly improves IF-OSN's RP by 9.8\%, 7.7\%, 10.0\%, and 8.0\% under FGSM, C\&W, BIM, and PGD attacks, respectively. CoDE demonstrates a relatively modest improvement, largely due to its already strong baseline performance with FFA alone. But the important thing is, since MgR concurrently performs the optimization for original forged images, IF-OSN and CoDE's RP increase 6.5\% and 1.1\%, respectively, further reduce the impact of the ANSM on original forged images.

\section{Conclusion}
\label{sect:Conclusion}
In this paper, we propose an ANSM that generates counteracting noise to be added to the input image, effectively safeguarding the forgery localization model against potential adversarial noise. We propose a two-stage optimization strategy to train ANSM. The first stage aligns the forgery-relevant features of adversarial forged images with those of original forged images, addressing feature shifting caused by adversarial noise. The second stage introduces mask-guided refinement, which leverages mask-level supervision to further enhance the precision of forgery localization. Experimental results demonstrate that the proposed method not only significantly restores the forgery localization model's performance on adversarial forged images but also have little performance degradation on original forged images. Furthermore, our proposed method exhibits superior generalization across multiple adversarial attack algorithms and datasets.

Future work could explore more adaptive purification mechanisms, such as dynamic noise modulation based on attack intensity estimation, to further reduce unintended modifications to original forged images. Additionally, unsupervised or semi-supervised approaches could be investigated, focusing on extracting generalized representations directly from unlabeled data to achieve universal defense under novel adversarial attack.
\bibliographystyle{IEEEtran}
\bibliography{references}

\end{document}